\documentclass[runningheads]{llncs}

\usepackage{xcolor}
\usepackage[utf8]{inputenc} 
\usepackage[T1]{fontenc}    
\usepackage{hyperref}       
\usepackage{url}            
\usepackage{booktabs}       
\usepackage{amsfonts}       
\usepackage{nicefrac}       
\usepackage{microtype}      
\usepackage{graphicx}
\usepackage{siunitx}

\usepackage[normalem]{ulem} 
\usepackage{multirow}
\usepackage{caption}
\usepackage{subcaption}
\usepackage{amsmath}
\usepackage{tikz}

\usepackage{float}

\title{SincVAE: A new semi-supervised approach to improve anomaly detection on EEG data using SincNet and variational autoencoder
\thanks{This paper was published in its final version on \textit{Computer Methods and Programs in Biomedicine Update} journal with DOI: \texttt{https://doi.org/10.1016/j.cmpbup.2025.100213} in Open Access mode. Please consider it as final and peer-reviewed version.}}

\author{
    Andrea Pollastro \and
    Francesco Isgrò \and
    Roberto Prevete
}

\institute{Department of Electrical Engineering and Information Technology, University of Naples Federico II, Naples, Italy}

\authorrunning{Pollastro, A., Isgrò, F., \& Prevete, R.}
\titlerunning{Please refer to: \texttt{https://doi.org/10.1016/j.cmpbup.2025.100213}}

\begin{document}

\maketitle

\begin{abstract}
Over the past few decades, electroencephalography monitoring has become a pivotal tool for diagnosing neurological disorders, particularly for detecting seizures. 
Epilepsy, one of the most prevalent neurological diseases worldwide, affects approximately \SI{1}{\percent} of the population. These patients face significant risks, underscoring the need for reliable, continuous seizure monitoring in daily life. 
Most of the techniques discussed in the literature rely on supervised machine learning methods. However, the challenge of accurately labeling variations in epileptic electroencephalography waveforms complicates the use of these approaches. Additionally, the rarity of ictal events introduces a high imbalance within the data, which could lead to poor prediction performance in supervised learning approaches.
Instead, a semi-supervised approach allows training the model only on data that does not contain seizures, thus avoiding the issues related to the data imbalance.
This work introduces a semi-supervised approach for detecting epileptic seizures from electroencephalography data based on a novel deep learning-based method called SincVAE. This method integrates SincNet, designed to learn an ad-hoc array of bandpass filters, as the first layer of a variational autoencoder, potentially eliminating the preprocessing stage where informative frequency bands are identified and isolated.
Experimental evaluations on the Bonn and CHB-MIT datasets indicate that SincVAE improves seizure detection in electroencephalography data, with the capability to identify early seizures during the preictal stage and monitor patients throughout the postictal stage. 
\end{abstract}

\section{Introduction}
\label{sec:introduction}
In recent years, the growth of information technology has had a significant impact on our daily life. 
In particular, its impact is evident by observing the quantity of data produced every day. 
These vast datasets provide a snapshot of the entities under observation, offering valuable insights for companies and organizations.
These insights not only enhance understanding but also furnish competitive advantages.
Consequently, it is crucial that these datasets are meticulously processed \cite{sagiroglu2013big}.
\textit{Anomalies} (also known as \textit{outliers}, \textit{deviants} or \textit{rare events} in some context \cite{aggarwal2017introduction}) represent unique behaviors within observed phenomena that can significantly influence the data generation process \cite{aggarwal2017introduction,chalapathy2019deep}.
The presence of anomalies in data during analysis can be dangerous, as they may lead to erroneous conclusions during data interpretation.
Consequently, it is crucial for analysts to meticulously identify and properly address these anomalies both before and during the analysis process.
The increasing interest in identifying and analyzing anomalies led scientists to isolate this problem into the active and dedicated research field of \textit{anomaly detection}.

In most instances, phenomena are monitored over time using \textit{time series} data. 
Within this framework, not every outlier is pertinent to an analysis.
For instance, some outliers could be attributed to sensor transmission errors or other sources of noise, while others might represent unusual phenomena, such as those observed in fraud detection scenarios \cite{aggarwal2017introduction}. 
In the former case, outliers can be eliminated or corrected to enhance data quality. Conversely, in the latter case, these outliers become anomalies of interest, often providing significant and crucial insights across diverse application domains \cite{aggarwal2017introduction}.
For instance, anomalies in credit card transaction records may indicate potential fraud or identity theft \cite{aleskerov1997cardwatch}. 
Similarly, an unusual pattern in network traffic could suggest that a compromised computer is transmitting confidential information to an unauthorized destination \cite{kumar2005parallel}. 
Additionally, anomalies detected in sensor data from civil infrastructure might signal structural damages \cite{pollastro2023semi,de2024dynamic}.

Significantly, healthcare represents a domain where anomaly detection can have a profound impact \cite{schlegl2017unsupervised,salem2014online,kavitha2021machine,naidoo2020unsupervised,you2022semi}. 
For example, the ability to detect abnormal physiological data through these techniques can expedite emergency responses and provide new insights into the progression of medical conditions, greatly influencing everyday life.
In \cite{vsabic2021healthcare}, the authors presented a methodology for identifying anomalies in heart rate data, leveraging its value as a noninvasive indicator of health concerns and physical activity. 
Meanwhile, the study in \cite{ukil2016iot} explores the potential of anomaly detection within healthcare analytics, specifically through IoT systems.

Numerous studies have explored detecting epileptic seizures as abnormal brain activity using electroencephalography (EEG) data, which are applicable in various settings, including brain-computer interfaces (BCI) \cite{shih2012brain,angrisani2022instrumentation}. 
In such contexts, accurately identifying seizures through EEG can trigger hardware interventions designed to assist patients and improve their quality of life \cite{hosseini2017optimized,liang2010closed,vaid2015eeg}.
Epilepsy is a chronic central nervous system disorder characterized by recurrent seizures, affecting approximately \SI{1}{\percent} of the global population \cite{shahbazi2018generalizable,litt2002prediction}. 
Seizures manifest as temporary disruptions in brain electrical activity, leading to symptoms like attention lapses, memory gaps, sensory hallucinations, or full-body convulsions. 
Despite treatment with various anti-epileptic drugs, about one-third of affected individuals frequently experience seizures that are challenging to control. 
These seizures significantly increase the risk of injury, limit personal independence and mobility, and can result in social and economic difficulties \cite{shoeb2009application,shoeb2010application}. 
The brain activity of individuals with epilepsy can be categorized into four states: regular brain activity (interictal), brain activity preceding the seizure (preictal), brain activity during the seizure (ictal), and brain activity immediately following a seizure (postictal) \cite{georgis2023supervised}.

Among various methodologies \cite{ahmed2016survey,bhuyan2013network}, machine learning (ML) techniques provide systematic approaches for extracting insights from a vast amount of data. 
These techniques enable researchers to explore the anomaly detection landscape, and propose solutions to diverse scenarios. 
In particular, deep learning (DL) methods have consistently outperformed traditional ML approaches in the last decades. 
This success is largely due to their ability to extract intricate patterns within complex, high-dimensional datasets \cite{bishop2006pattern,goodfellow2016deep}. 
As a result, DL has played a leading role in various fields that leverage data-driven strategies, especially in the development of anomaly detection methodologies \cite{omar2013machine,pang2021deep,wang2020deep}.
A significant portion of DL approaches for anomaly detection relies on autoencoder (AE) architectures \cite{li2019video,fan2020robust,zhou2017anomaly}. 
AEs are neural networks comprising two main components: an encoder, which compresses the input into a latent representation, and a decoder, which reconstructs that representation back to the input space.
In anomaly detection, a well-established strategy involves training AEs specifically to minimize reconstruction errors for normal data instances, effectively framing it as a semi-supervised problem \cite{pollastro2023semi}. 
This technique results in higher reconstruction errors when processing anomalous data, making these errors a useful anomaly score. When combined with a user-defined decision rule, this score becomes an effective tool for classifying data as normal or anomalous \cite{pollastro2023semi}.
Using this methodology, AEs effectively handle variations in normal data while being less affected by the specific characteristics of rare and diverse anomalies. Moreover, their semi-supervised approach eliminates the need for labeled anomaly data during training, making them well-suited for real-world scenarios where such labels are frequently limited or unavailable.
AEs architectures can be realized in various forms, including variational autoencoder (VAE), which is a generative model where the encoder and the decoder do not represent a functional mapping as in standard AEs \cite{kingma2013auto}.
Due to their promising performance, there is a growing interest in using generative models to identify anomalies \cite{challu2022deep,sheynin2021hierarchical,schlegl2017unsupervised,choi2018generative} that, within the context of AEs, has been manifested through the use of VAEs \cite{an2015variational,luchnikov2019variational,zimmerer2018context,ren2020data}.

VAEs can be implemented using a variety of processing strategies documented in the literature, including multilayer perceptrons (MLP) \cite{noriega2005multilayer} and recurrent neural networks (RNN) \cite{medsker2001recurrent}. Convolutional neural networks (CNNs), widely utilized for processing time series data \cite{zhao2017convolutional,cui2016multi,borovykh2017conditional,koprinska2018convolutional,zheng2014time}, operate through a series of trainable filters. 
These filters are inspired by the biological mechanisms of visual perception, enabling the recognition of informative patterns from the input data \cite{lindsay2021convolutional}. In the field of speaker recognition, Ravanelli and Bengio introduced \textit{SincNet} \cite{ravanelli2018speaker}, a CNN whose filters are structured as a learnable array of parametrized sinc functions that are designed to operate as bandpass filters. 

Numerous ML methods have been applied to detect epileptic seizures, with the aim of not only mitigating risks for patients but also enhancing their ability to seek timely assistance and reduce the likelihood of injury \cite{georgis2023supervised,humairani2022wavelet,ahmad2017mallat,boonyakitanont2020review}. Although the physiological activity involved in epilepsy is inherently multi-class, according to the four brain activity states introduced above, many studies, such as \cite{shoeb2009application,shoeb2010application}, approach seizure detection as a binary supervised classification problem, where the two classes to be identified are \textit{seizure activity} (ictal) and \textit{non-seizure activity} (interictal). This reduction to two classes is due to the challenges and impracticalities associated with an expert's ability to identify and label transitional states between ictal and interictal states, whereas an expert can categorize brain electrical activity as seizure or non-seizure, adhering to current standard clinical protocols \cite{shoeb2010application}. 
Furthermore, epilepsy detection could significantly benefit from a semi-supervised learning approach, especially given the scarcity of seizure data in contrast to the abundance of non-seizure data. For a patient known to suffer from epilepsy, data collected during non-seizure states can be easily obtained and can provide a substantial foundation for building a ML model capable of accurately detecting seizures when they occur.
\newline
\newline
This work proposes SincVAE, a DL architecture that introduces SincNet in the VAE framework to process EEG data for seizure detection in a semi-supervised setting.
Bandpass filters are crucial for isolating meaningful frequency bands from input signals, especially in BCI applications \cite{ang2012filter,shajil2020multiclass,apicella2022eeg,arpaia2023sinc}. 
However, the application of bandpass filters typically involves two stages: (i) selecting well-known informative band frequencies pertinent to the specific context, and (ii) conducting an analysis where the analyst identifies and isolates relevant frequency bands from the data. 
SincNet offers an efficient method in this process, providing a compact and precise solution to develop custom bandpass filters optimized for specific applications \textcolor{black}{through the training stage}.
Applications of seizure detection could benefit from this improvement since it could allow to enhance (or eventually, eliminate), preliminary phases dedicated to the study and extraction of frequency bands from the data and, thus, achieve a more refined and precise extraction of frequency bands, leading to an acceleration and improvement of the whole processing pipeline development.
\textcolor{black}{Furthermore, SincNet filters act as bandpass filters, and after the training stage, they exhibit inherent interpretability \cite{ravanelli2018speaker}, offering analysts valuable insights into the frequency bands relevant to the seizure detection on the training dataset. The challenge of interpreting DL models is widely acknowledged in the literature, as their over-parameterized, black-box nature often obscures the rationale behind their predictions \cite{li2022interpretable}. Consequently, numerous algorithms have been proposed to enhance the interpretability of DL models by analyzing their components and understanding their behavior during inference.
\textcolor{black}{In particular, various strategies have been proposed to address these issues, including post-hoc explanation methods (e.g., saliency maps, SHAP, LIME), model simplification, and the design of inherently interpretable architectures. In most conventional deep learning models, the learned configurations, such as convolutional filters or dense layer weights, lack clear structure and are not directly interpretable, making it difficult to understand what the model has actually learned \cite{chakraborty2017interpretability}}.
In this context, SincNet allows to have a notable advantage, as its filters design inherently provides interpretability without requiring the application of additional algorithms. This built-in transparency enables a more direct understanding of the model's processing, offering valuable insights into the specific frequency bands involved, which is particularly important for clinical applications.}
Additionally, the signal decomposition performed by SincNet enhances the VAE's capabilities in detecting and extracting meaningful features that are representative of the non-anomalous condition during the training phase. This improvement allows for more accurate representation of the non-anomalous state, thereby emphasizing the distinction between non-anomalous and anomalous patterns in the data.
A semi-supervised approach \textcolor{black}{\cite{tan2025seda,hu2023,hu2024,pollastro2023semi,de2024dynamic}} obtained through the application of a VAE network is particularly advantageous as it allows to fit a model on training sets that contain only non-anomalous instances \cite{aggarwal2017introduction,ruff2019deep,akcay2019ganomaly,pollastro2023semi}, in this context represented by interictal brain activity. 
This choice is motivated by the flexibility this approach offers, especially in scenarios with imbalanced datasets, where anomalous data points are scarce or even absent. 
\textcolor{black}{Indeed, data imbalancing is a well-known challenge in healthcare problems, and various proposals in the literature have been made to address it \cite{zhao2024digan,yu2014improved,roy2024learning}.}
This is particularly relevant in healthcare, where the prevalence of non-anomalous data contrasts with the rarity of abnormal data.
\newline
\newline
This paper is organized as follows: Section 2 provides an overview of existing works on AEs and VAEs for seizure detection from EEG data; Section 3 introduces the proposed method, beginning with an overview of the SincNet and VAE frameworks; Section 4 details the experimental assessment; Section 5 presents the results of the experimental evaluation; and Section 6 concludes the paper, discussing potential future directions.
\section{Related Works}
\label{sec:related_works}
This section reviews existing research that employs AEs and VAEs for detecting seizures from EEG data.

Khan et al. in \cite{khan2023shallow} show a novel seizure detection method by integrating AEs with traditional classifiers in an hybrid model. Specifically, the AE is used to extract features from the input data through its encoder. These latent representations are then fed into a classifier, such as a support vector machine (SVM) \cite{hastie2009elements} or k-nearest neighbors (k-NN) \cite{hastie2009elements}, to perform a supervised classification.
Yuan et al. in \cite{yuan2018multi} proposed a novel approach that employs an AE model to extract multi-view features from multi-channel EEG data. Then, the extracted features are then used for supervised seizure detection.
Abdelhameed et al. in \cite{abdelhameed2019semi} proposed a methodology based on convolutional VAE to extract features from EEG input data with the goal of eliminating the need for an engineered feature extraction phase previous to the model fitting. Then, the extracted features are fed in a supervised classifier to detect seizures.
The same authors in \cite{abdelhameed2021deep} improved the feature extraction phase by using a two-dimensional deep convolutional AE (2D-DCAE). The extracted features are then used to train a neural network-based classifier for seizure detection in a supervised manner.
In \cite{abdelhameed2021efficient} instead, the same authors proposed a methodology based on a convolutional VAE trained in a supervised manner to perform simultaneously automatic feature learning and classification on the data latent representations.
Daoud et al. in \cite{daoud2019deep} compared two methodologies, where the main goal was automatic feature extraction. The first method utilizes a deep convolutional AE, where features extracted from the encoder are classified using a MLP classifier. The second method consists in an unsupervised pipeline integrating a deep convolutional VAE with the k-means \cite{hastie2009elements} clustering algorithm fitted on the latent representations of data.
Wang et al. in \cite{wang2022epileptic} introduced the residual convolution VAE (RCVAE) method to extract features from EEG recordings. The extracted features are used to train a supervised neural network classifier for seizure detection. The same research group in \cite{he2023unsupervised} proposed an improved version called residual convolution VAE with randomly translation strategy (RTS-RCVAE) to solve issues related to the introduction of data augmentation strategies.
Wen et al. in \cite{wen2018deep} proposed the AE-CDNN method to extract the features prior to a supervised classification stage performed using MLPs.
Similarly, Shoeibi et al. in \cite{shoeibi2022detection} employed AEs for dimensionality reduction. Latent representations of data were used to fit several methods to classify seizures including the adaptive neuro-fuzzy inference system (ANFIS) and its variants.
Huang et al. in \cite{huang2022novel} used AEs for feature extraction in epilepsy detection, comparing their performance against traditional principal component analysis (PCA) \cite{hastie2009elements}. Then, the authors employed three metrics, namely original-to-reconstructed signal ratio (ORSR), mean squared error (MSE) and cosine similarity (CS), to evaluate the signal reconstruction and identify these metrics as sensitive indicators for epilepsy. Also, the authors utilize permutation importance and Shapley additive explanations (SHAP) \cite{lundberg2017unified} for model interpretability, confirming the better efficacy and rationale of the AE-based feature extraction compared to the PCA one.

Most of the works on seizure detection rely on AEs or VAEs primarily for feature extraction, which are then utilized in conjunction with supervised methods for classifying EEG data. 
There exists a relatively small subset of methodologies that rely exclusively on AEs and VAEs for the entire process of seizure detection, using only the reconstruction error metrics for the classification stage. 
The authors in \cite{yildiz2022unsupervised} relied on a three-layered convolutional VAEs trained exclusively on non-seizure EEG recordings to detect seizures. Reconstruction error was used to identify seizure activity. In particular, they adopted the median reconstruction error as a metric to distinguish between seizure and non-seizure events.
You et al. in \cite{you2022semi} used a VAE to model the latent representations of non-seizure EEG signals. They then used deviations from these baseline representations, along with reconstruction loss, to devise a personalized anomaly score for each patient.
De Sousa et al. in \cite{de2024detection} used AEs and VAEs to detect interictal epileptiform discharges (IEDs) by treating these events as anomalies within EEG data. The comparative analysis in their study revealed that VAEs outperformed traditional AEs, likely due to their enhanced ability to model the distribution of EEG data and handle anomalies more effectively. 
Potter et al. in \cite{potter2022unsupervised} proposed an architecture based on AE with a transformer encoder to reconstruct EEG recordings of non-seizure activity. Then, reconstruction error served as anomaly score to detect EEG recordings containing seizures.

\section{Proposed Method}
\label{sec:proposed_method}
This section introduces the SincNet model, followed by an overview of the VAE framework. Finally, it presents SincVAE, the proposed method developed in this work.

\subsection{SincNet}
Ravanelli and Bengio in \cite{ravanelli2018speaker} introduced SincNet in the context of speaker recognition.
The convolution operation in a standard CNN layer is represented as follows \cite{ravanelli2018speaker}:
\begin{equation}
y[n] = x[n] * h[n] = \sum_{l=0}^{L-1}x[l] \cdot h[n-l],
\end{equation}
where $x[n]$ is the input signal segment, $h[n]$ is the filter of size $L$, $*$ denotes the convolution operation, and $y[n]$ is the output. Typically, each element of the filter $h[n]$ is learned during the training phase. SincNet modifies this process by using a pre-defined function $g$ that depends on a limited number of learnable parameters $\psi$:
\begin{equation}
y[n] = x[n] * g[n,\psi].
\end{equation}
The function $g$ is designed to implement a filterbank consisting of rectangular bandpass filters.
The magnitude of a generic bandpass filter, in the frequency domain, can be expressed as the difference between two low-pass filters as follows \cite{ravanelli2018speaker}:
\begin{equation}
G[f,f_1,f_2] = \text{rect}\Bigl(\frac{f}{2f_2}\Bigr) - \text{rect}\Bigl(\frac{f}{2f_1}\Bigr),
\end{equation}
with $f_1$ and $f_2$ representing the learned low and high cutoff frequencies, respectively. 
The $\text{rect}(\cdot)$ function denotes the rectangular function in the frequency domain. In the time domain, the function $g$ is defined as:
\begin{equation}
g[n,f_1,f_2] = 2f_2\text{sinc}(2\pi f_2n) - 2f_1\text{sinc}(2\pi f_1n),
\end{equation}
where $\text{sinc}(x) = \sin(x)/x$.

The cutoff frequencies are randomly initialized within the range $[0, f_s/2]$, where $f_s$ is the sampling frequency of the input signal. To ensure that $f_1 \ge 0$ and $f_2 \ge f_1$, the parameters are adjusted as:
\begin{align}
f_1^{\text{abs}} &= |f_1|,\\
f_2^{\text{abs}} &= f_1 + |f_2 - f_1|.
\end{align}
The authors in \cite{ravanelli2018speaker} point out that the training process naturally keeps $f_2$ below the Nyquist frequency \cite{grenander1959nyquist}, eliminating the need for explicit constraints.

Additionally, SincNet applies a windowing function \cite{rabiner2010theory} to smooth the discontinuities at the boundaries of $g$. \textcolor{black}{This is achieved by multiplying $g$ with a window function $w$:
\begin{equation}
g[n,f_1,f_2] = g[n,f_1,f_2] \cdot w[n],
\end{equation}
where the windowing function $w$ is the Hamming window \cite{mitra2001digital}, defined as
\begin{equation}
w[n] = 0.54 - 0.46 \cdot \cos\Bigl(\frac{2\pi n}{L}\Bigr).
\end{equation}
}

All operations within SincNet are differentiable, allowing the optimization of cutoff frequencies alongside other neural network parameters through gradient-based methods.

\subsection{Variational Autoencoder}

A variational autoencoder (VAE) \cite{kingma2013auto} is a generative model comprising a probabilistic decoder and encoder. 
The decoder $p_{\theta}(x|z)$, with parameters $\theta$, generates new data $x$ given a latent variable $z$, while the encoder $q_{\phi}(z|x)$, with parameters $\phi$, approximates the true posterior $p_{\theta}(z|x)$, which is generally intractable.
A prior $p(z)$, typically the isotropic unit Gaussian $\mathcal{N}(0, I)$, is placed over the latent variables.

To train a VAE, the parameters $\theta$ and $\phi$ are optimized by maximizing the evidence lower bound (ELBO) of the marginal log-likelihood $\log p(x)$:
\begin{equation}
\log p(x) \ge \mathcal{L}(\theta,\phi;x) = \mathbb{E}_{q_\phi(z|x)}[\log{p_\theta(x|z)}] - D_{KL}(q_\phi(z|x)||p(z)),
\end{equation}
where $D_{KL}(\cdot||\cdot)$ is the Kullback-Leibler divergence. The first term represents the expected reconstruction accuracy, and the second term regularizes the latent space by ensuring $q_\phi(z|x)$ is close to the prior $p(z)$.
To make the optimization tractable, the reparameterization trick is employed. 
Assuming $q_\phi(z|x)$ is a Gaussian with diagonal covariance, each latent variable $z_i \sim q_\phi(z_i|x) = \mathcal{N}(\mu_i, \sigma_i^2)$ is reparameterized as a deterministic transformation of a noise variable $\epsilon_i \sim \mathcal{N}(0,1)$ \cite{burgess2018understanding}:
\begin{equation}
z_i = \mu_i + \sigma_i\epsilon_i.
\end{equation}
This reparameterization enables backpropagation through stochastic sampling, making the ELBO differentiable with respect to the model parameters, and allowing it to be optimized using gradient-based methods. Under this framework, the ELBO can be maximized, offering significant flexibility in the design of the encoder and decoder architectures.

\subsection{SincVAE}
VAEs represent a versatile framework where both the encoder and decoder can be implemented in various ways. This flexibility allows the method to be adapted to different types of data. 
In particular, several research works have employed VAEs implemented with convolutional layers to process time series data (see, for example \cite{sadouk2019cnn,choi2022multivariate}).
During the training stage, filters of the convolutional layers are optimized to extract the features necessary to solve a given task. 
While these filters may be meaningful to the neural network, they are not intuitive for humans and, on time series, they can often take noisy and incongruous multi-band shapes \cite{ravanelli2018speaker}.
Also, a crucial component of current waveform-based CNNs is the first convolutional layer, since it handles high-dimensional inputs and is particularly susceptible to vanishing gradient issues, especially in very deep architectures \cite{ravanelli2018speaker}, and this aspect might affect the performance of the whole network.
Thus, to simplify the feature extraction stage of the neural network, one approach could involve preprocessing the time series to reduce the complexity of the input data using various techniques, including signal decomposition via fast Fourier transform (FFT) \cite{brigham1988fast}, which is commonly applied in the BCI domain (see, for example \cite{ravi2019user,nakayama2007brain}).
However, this approach is restricted to predefined frequency bands well-established in the literature or requires a prior analysis phase where experts examine the signal to identify points of interest. This process can be time-consuming and prone to errors, potentially affecting the overall accuracy in solving a given task.
As stated by the authors in \cite{ravanelli2018speaker}, SincNet has the potential to address these issues. Acting as band-pass filters with a minimal number of trainable parameters, SincNet filters in the first layer decompose the signal into frequency bands during the optimization process. This approach provides filters that can be interpreted by humans, eliminates the need for manual preprocessing to simplify the input data and enhances the feature extraction performed by subsequent convolutional layers.

In the context of anomaly detection using VAEs in a semi-supervised setting, the model is trained exclusively on non-anomalous data. The goal is to optimize the feature extraction and reconstruction processes - handled by the encoder and decoder networks, respectively - specifically for normal data instances \cite{an2015variational}. As a result, once training is complete, the VAE's encoder is expected to encounter difficulties in extracting meaningful features from anomalous data, leading to poor data reconstructions. In this framework, the more effectively the VAE performs feature extraction on normal data, the clearer the distinction between anomalous and non-anomalous inputs will be.

\begin{figure}[ht!]
    \centering
    \scalebox{1}{
        \includegraphics[width=\textwidth]{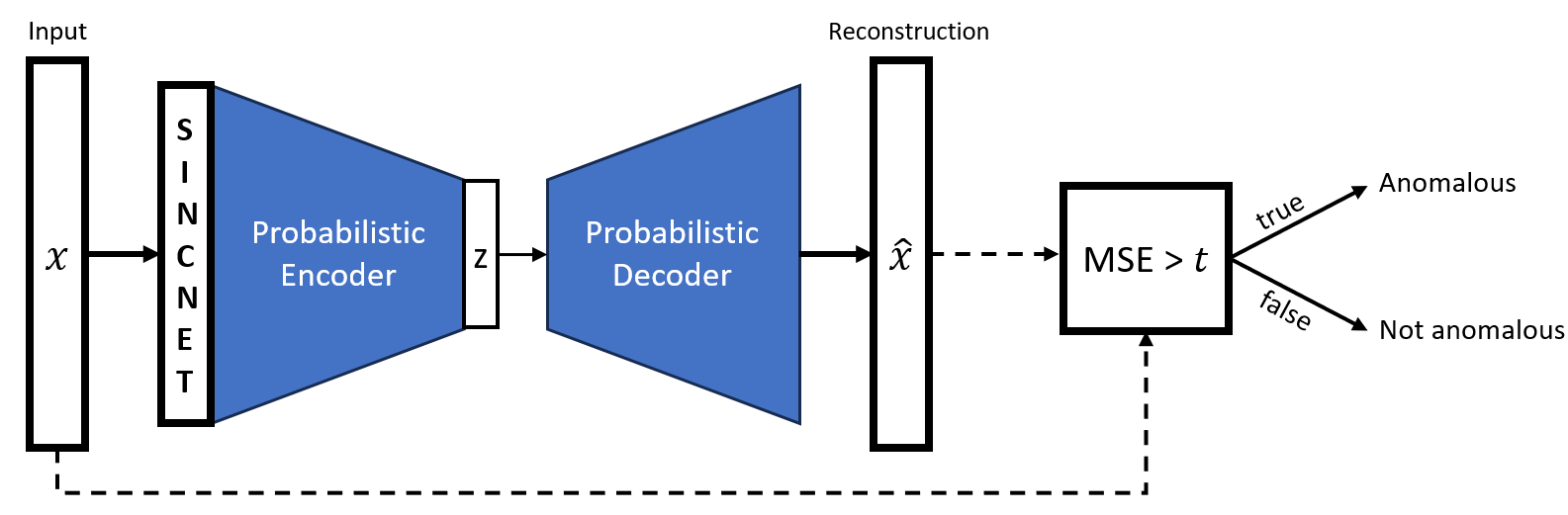}
    }
    \caption[The SincVAE architecture]{Graphical representation of the SincVAE architecture. An input time series $x$ is given as input to the VAE that generates its reconstruction $\hat{x}$. The VAE's probabilistic encoder is preceded by the SincNet layer, that filters the data by using an ad-hoc array of bandpass filters learned during the training stage. Then, both the input time series and its reconstruction are used to compute the reconstruction error (shown in terms of MSE in this picture). Finally, a threshold $t$ is involved to classify the input time series as anomalous or not.}
    \label{fig:proposedpipeline}
\end{figure}

The proposed method, referred to as \textit{SincVAE}, incorporates SincNet as a convolutional layer preceding the VAE's probabilistic encoder. Specifically designed to improve anomaly detection on time series data, SincVAE leverages SincNet's ability to learn a customized array of bandpass filters while providing a set of human-interpretable filters. 
This allows the decomposition of input time series, thereby enhancing the VAE's feature extraction and improving the overall effectiveness of time series processing. In particular, the introduction of the SincNet layer is expected to improve the VAE's functioning in extracting features more representative of a non-anomalous state during the training stage. Consequently, this should result in producing poorer reconstructions for anomalous data, thereby emphasizing the distinction between non-anomalous and anomalous data, and improving classification accuracy. Within this framework, the reconstruction error, computed in terms of MSE between the input data and its reconstruction, is involved in the classification process given a threshold $t$. Specifically, an MSE whose value is higher than $t$, will indicate that the input data is anomalous; conversely, it will indicate that the input data is non anomalous. A graphical representation of the proposed pipeline is shown in Figure \ref{fig:proposedpipeline}.

This study investigates the effectiveness of SincVAE in addressing anomaly detection challenges in the healthcare domain, with a specific focus on seizure detection from EEG data.
It is expected that the integration of SincNet within the VAE framework in the seizure detection problem can enhance the learning of non-seizure patterns within EEG data, thus leading to a more robust identification of seizure patterns. 
The SincVAE's ability to learn custom bandpass filters could not only boost the overall efficacy of the seizure detection process, but also potentially reduce (or, even, eliminate) the preprocessing often required to identify band frequencies where informative content is more present, which is frequently done in the BCI context.

\section{Experimental Assessment}
\label{sec:experimental_setup}
In this work, the effectiveness of SincVAE for seizure detection was analyzed through experimental results obtained from two datasets: the Bonn dataset \cite{andrzejak2001indications} and the CHB-MIT dataset \cite{physiobank2000physionet}. These datasets have been widely utilized in numerous studies, including \cite{prasanna2021automated,choi2019novel,zhou2018epileptic,tasci2023epilepsy,park2018epileptic,kaziha2020convolutional,hassan2022epileptic,shoeibi2022detection,qiu2018denoising}. 
The seizure detection problem was tackled with a semi-supervised approach, therefore, models were trained exclusively on non-seizure data. The classification of data into seizure or non-seizure categories is determined by analyzing the reconstruction error obtained from the input data and their reconstructions generated by the model.

For a robust comparative analysis, the experiments employed a fixed base VAE architecture in two distinct configurations: (i) the VAE with a SincNet preceding the encoder network, referred to as SincVAE, and (ii) the VAE without the SincNet layer, referred to as VAE. 
Specifically, the AE-CDNN architecture detailed in \cite{wen2018deep} was adopted, with modifications applied to the specific needs of VAE operations, particularly in the latent space and training configurations.
This methodological choice facilitates a clear examination of the SincNet layer's contribution to the VAE's performance.
In accordance with \cite{ravanelli2018speaker}, the SincNet layer was followed by a Layer Normalization and an activation function. 

The model selection stage, performed separately on each dataset, was aimed at optimizing the hyperparameters of the SincNet layer to enhance its effectiveness in seizure detection. 
Simultaneously, the latent space was tuned and consistently applied on both the VAE and SincVAE models. 
The best hyperparameter set was chosen to strike an optimal balance between model complexity and inference speed, i.e. by reducing the number of parameters while maintaining robust performance.

Grid search \cite{liashchynskyi2019grid} was used as the method for automatic hyperparameter tuning, employing specific search spaces outlined in the respective dataset sections.
The Adam optimizer was selected for weights optimization, with a learning rate of $0.0005$. Data were processed in random batches of $128$ samples each. The training stage was limited to a maximum of 1000 epochs, and early stopping \cite{yao2007early} was used as convergence criterion, with a patience of 20 epochs. These hyperparameters were established through manual preliminary assessment and fixed through all the experiments.

\subsection{The Bonn Dataset}
\paragraph{Dataset description}
The Bonn dataset \cite{andrzejak2001indications}, acquired by the Bonn University in Germany, consists of five distinct collections of EEG signals. Each collection includes 100 single-channel EEG segments, each lasting 23.6 seconds, derived from continuous multi-channel EEG recordings. These segments were selected after a visual inspection to eliminate artifacts. Each EEG segment was captured using a 128-channel amplifier system paired with a 12-bit analog to digital converter at a sampling rate of \SI{173.61}{Hz}. 
The dataset is organized into five groups:
\begin{itemize}
    \item Sets A and B contain EEG recordings from healthy volunteers. In particular, Set A contains data acquired during eyes-opened condition, and Set B data acquired during eyes-closed conditions;
    \item Sets C and D consist of interictal EEG signals from patients post-successful epilepsy surgery. Signals in Set C were recorded from the hippocampal formation opposite the epileptogenic zone, and Set D from within the epileptogenic zone;
    \item Set E contains only ictal segments.
\end{itemize}
Since this work focuses on the seizure detection as a binary anomaly detection problem, only Sets A, B, and E were utilized for analysis, following the methodology outlined in Table 1 of \cite{hassan2022epileptic}. Specifically, the efficacy of the proposed method was tested through the following comparisons:
\begin{enumerate}
\item Set A vs Set E
\item Set B vs Set E
\end{enumerate}
It is important to remark that the proposed method exclusively utilizes non-seizure data (Sets A and B) during the training stages. 

\paragraph{Data preprocessing}
\textcolor{black}{First, the data were segmented into one-second frames, yielding 2300 samples of 173 data points each for Sets A, B, and E. In both cases (i.e., Set A vs Set E and Set B vs Set E), the final \SI{20}{\percent} of each training dataset (Sets A and B) was set aside to evaluate the models' ability to classify non-seizure data alongside the seizure data in Set E. This resulted in 1840 training samples and 2760 testing samples.}
Subsequently, the dataset underwent filtering with a 40 Hz low-pass filter following \cite{andrzejak2001indications}. Z-score normalization \cite{apicella2023effects} was then applied.
A summary of the Bonn dataset, including details on the preprocessing and data partitioning, is presented in Table \ref{tab:bonn_dataset_description}.
\begin{table}[t]
\caption[Bonn dataset description]{Details on the training and test sets sizes involved in the experimental session on the Bonn dataset.}
\begin{center}
\scalebox{.8}{
    \begin{tabular}{c|cc}
        \hline
        \textbf{Case} & \textbf{Training set} & 
        \textbf{Test set}\\
        \hline
        Set A vs Set E & 1840 & 2760 \\
        Set B vs Set E & 1840 & 2760 \\
        \hline
    \end{tabular}
}
\end{center}
\label{tab:bonn_dataset_description}
\end{table}

\paragraph{Model selection}
The model selection stage was performed through the search space defined in Table \ref{tab:bonn_search_space}. \textcolor{black}{As previously mentioned, the hyperparameter ranges were selected based on preliminary experiments. These initial trials allowed us to identify areas of the search space where training generally led to promising outcomes, which were subsequently used to define the grid for model selection.}
\begin{table}[ht!]
\caption[Search spaces for the grid search conducted on the Bonn dataset]{Search spaces for the grid search conducted during the model selection process on the Bonn dataset.}
\begin{center}
\scalebox{.8}{
    \begin{tabular}{c|c}
        \hline
        \textbf{Hyperparameter} & \textbf{Search Space}\\
        \hline
        Kernel Size & \{3, 5, 7\} $\bigcup$ \{11, 21, \dots, 131\} \\
        Filters & $\{2^n\}, 1 \le n \le 9$\\
        Activation Function & \{ ReLU, Tanh, Identity \} \\
        Latent Space Dimension & $\{2^n\}, 3 \le n \le 7$\\
        \hline
    \end{tabular}
}
\end{center}
\label{tab:bonn_search_space}
\end{table}
The total number of configuration explored was 2295.
10-fold cross validation \cite{hastie2009elements} was employed as the resampling method, with the mean MSE across the 10 folds used as the criterion for selecting the best architecture. In this setup, for each fold of the cross validation, \SI{20}{\percent} of the training data was sampled and designated as the validation set. This analysis was done separately on the Set A and Set B. For the sake of clarity, the methodology in this paragraph is explained through the results obtained on the Set A only.

\begin{figure}
    \centering
    \includegraphics[width=0.9\linewidth]{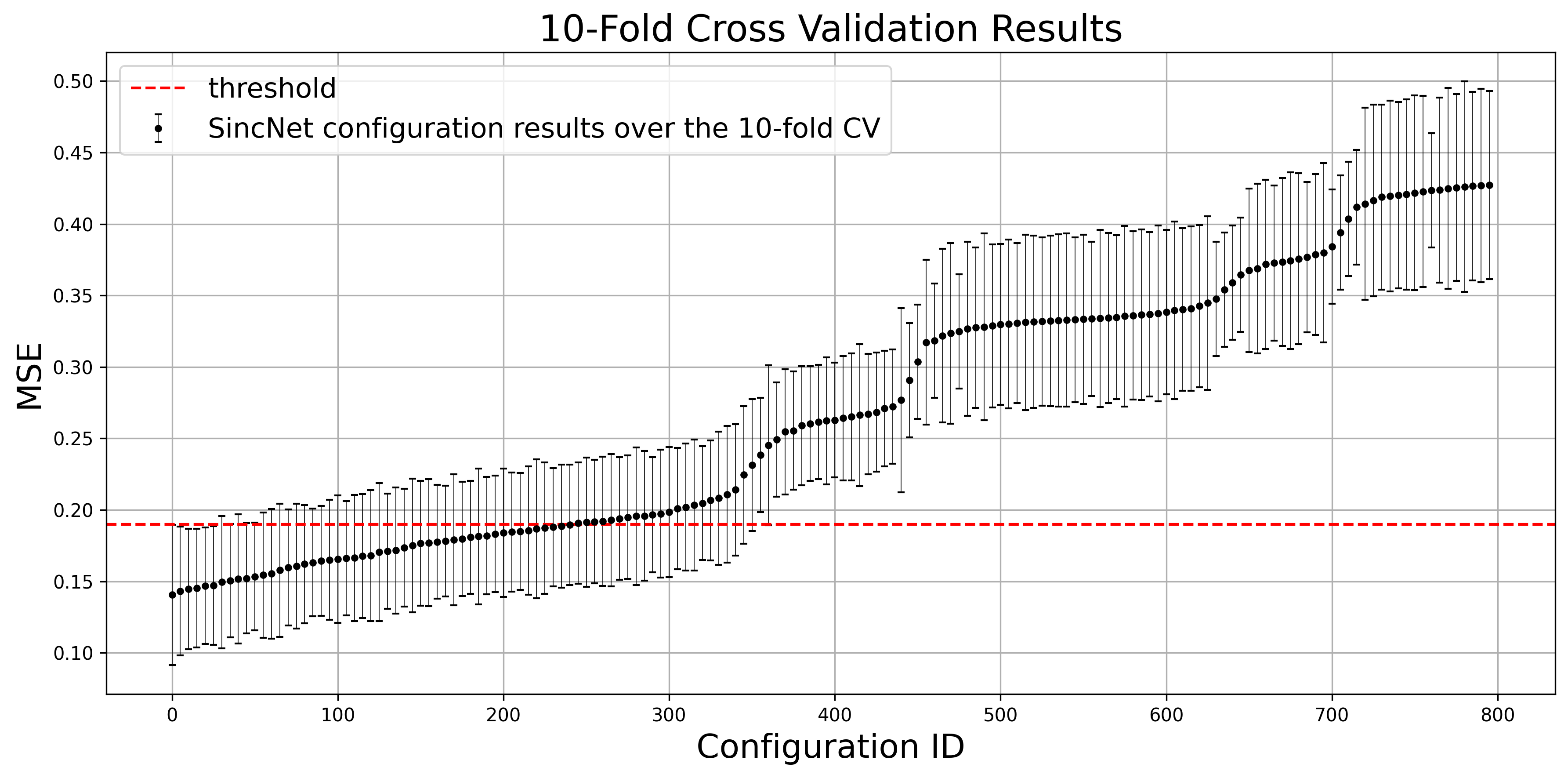}
    \caption{Graphical representation of the 10-fold cross validation applied to hyperparameter optimization on Set A, sorted by ascending mean MSE over the folds. For visualization purposes, (a) only the top 800 configurations are shown, and (b) one configuration every five is shown. The red dashed line indicates the threshold value used to filter configurations that fall within one standard deviation of the configuration with the lowest mean MSE.}
    \label{fig:bonn_CVA}
\end{figure}

From the cross validation analysis, as shown in Figure \ref{fig:bonn_CVA}, the configuration with the lowest mean MSE across the folds exhibited a result distribution that overlapped with several other configurations. This suggests that, while this configuration demonstrated strong performance in terms of MSE, it was not distinctly superior, highlighting the need for further analysis \textcolor{black}{through statistical hypothesis testing. To systematically assess performance differences among configurations, we followed a structured statistical testing procedure. First, the Shapiro-Wilk \cite{shapiro1965analysis} test was applied to evaluate the normality of each configuration's result distribution. This informed the choice between parametric (ANOVA \cite{kim2017understanding}) and non-parametric (Kruskal-Wallis \cite{ostertagova2014methodology}, Mann-Whitney U-test \cite{mcknight2010mann}) tests for subsequent comparisons. 
When statistical tests revealed significant differences, we prioritized configurations with superior performance as indicated by the test outcomes. Conversely, when no significant differences were found, configurations were considered statistically equivalent, and the final choice was guided by practical considerations such as model complexity or computational cost.}

Initially, configurations with a mean MSE that did not fall within one standard deviation of the lowest mean MSE were excluded from further analysis. This filtering process narrowed down the selection to the top 244 configurations.

Then, the normality of the results for each configuration was evaluated using the Shapiro-Wilk test. By setting the significance level at \(\alpha = 0.05\), it was found that 42 out of the 244 configurations yielded a p-value lower than \(\alpha\), indicating that their result distributions do not conform to a normal distribution.
Consequently, the Kruskal-Wallis test was applied to compare all configurations, using the same significance level $\alpha = 0.05$. This test produced a p-value lower than $\alpha$, confirming significant differences among the distributions of the configurations. 

\begin{figure}[ht!]
    \centering
    \scalebox{.9}{
        \includegraphics[width=\textwidth]{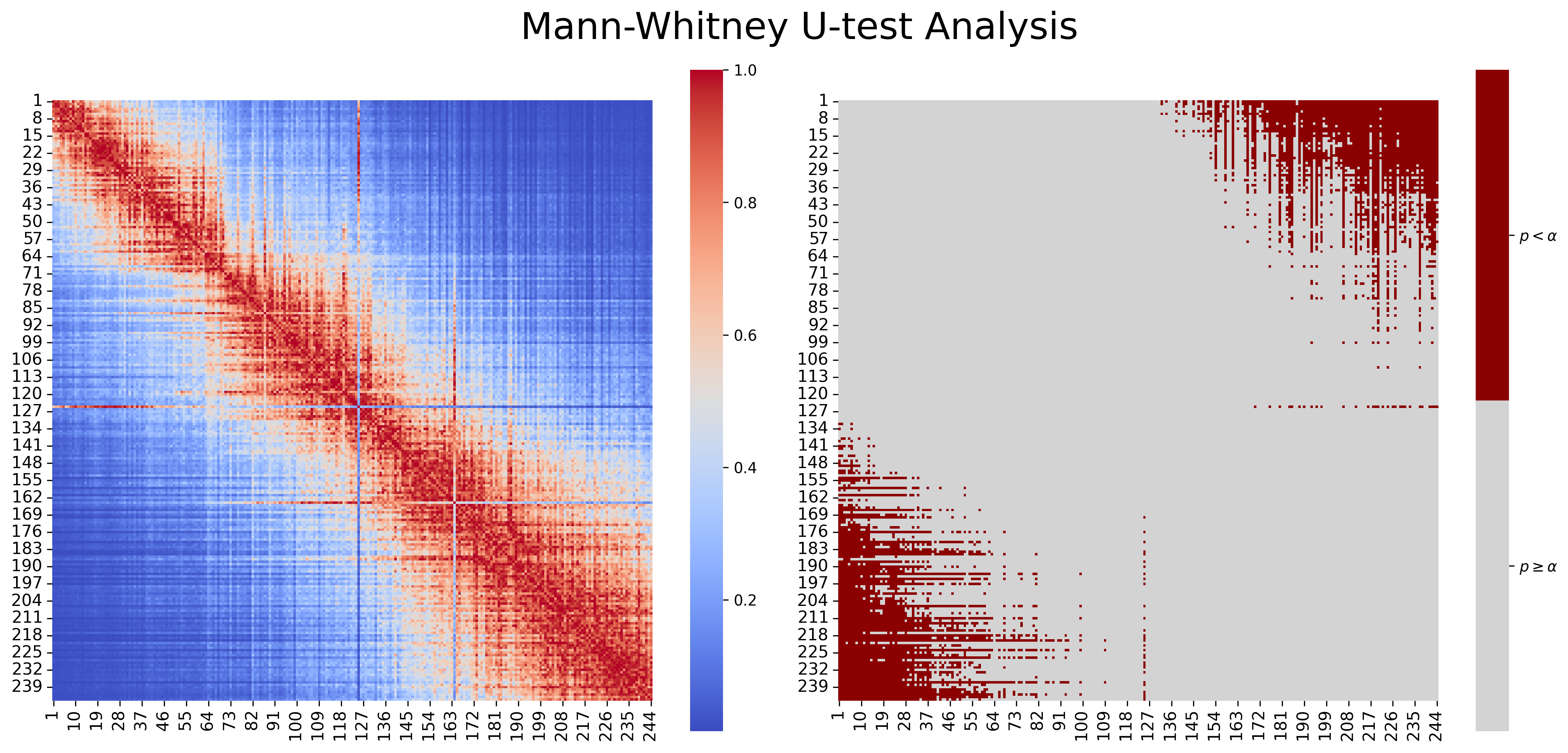}
    }
    \caption[Mann-Whitney U-test results on the Bonn Dataset, Set A]{Graphical representation of the Mann-Whitney U-test of multiple configurations over the best 244 configuration on Set A. The left heatmap details pairwise comparisons between configurations, with color intensity reflecting the p-value magnitude. The right heatmap displays these p-values thresholded by the significance level $\alpha = 0.05$; red denotes p-values below $\alpha$, highlighting statistically significant differences, whereas gray indicates p-values above $\alpha$, indicating non-significant differences between configurations.}
    \label{fig:bonn_utest_setA}
\end{figure}

Then, a pairwise comparison of the remaining configurations was conducted using the Mann-Whitney U-test. The results are shown in Figure \ref{fig:bonn_utest_setA}, where the 244 configurations are ordered in ascending order of mean MSE over the folds.
Assuming a significance level of $\alpha = 0.05$, this analysis was designed to identify configurations that do not significantly differ from the one exhibiting the lowest mean MSE across the 10 folds. Configurations yielding a p-value lower than $\alpha$ in these comparisons were excluded. This process effectively isolated the top 131 configurations that demonstrated no significant difference in MSE performance compared to the best-performing configuration.

As mentioned earlier, the best configuration was selected by prioritizing the lowest model complexity among the 131 configurations. From these, the configurations with the smallest latent space, specifically 32, were chosen. Then, from the remaining options, the configuration with the Identity function as the activation function for the SincNet layer was selected. Finally, among the remaining configurations, those with 16 filters were chosen, leading to the final selection of a kernel size of 41.

As mentioned before, the same procedure was applied to Set B. The analyses resulted in similar results and to the same configuration.

\subsection{The CHB-MIT Dataset}
\paragraph{Dataset description}
The CHB-MIT dataset \cite{physiobank2000physionet} was acquired by the Children's Hospital Boston (CHB) and the Massachusetts Institute of Technology (MIT). It features EEG recordings from pediatric patients diagnosed with intractable seizures. These patients were observed over multiple days following the discontinuation of anti-seizure medications, to assess their seizure activity and evaluate their suitability for surgical treatment. 
The dataset includes recordings from 24 subjects (see Table \ref{tab:chbmit_summary} for subjects' details).

\begin{table}[ht!]
\centering
\caption{Summary of CHB-MIT patients data, along with training and test sets sizes for each subject.}
\scalebox{.8}{
    \begin{tabular}{c|cccc|cc}
    \hline
    \textbf{Subject} & \textbf{Gender} & \textbf{Age} & \textbf{\# Tracks} & \textbf{\# Seizures Tracks} & \textbf{Training set} & 
        \textbf{Test set}\\
    \hline
    1 & F & 11 & 42 & 7 & 114230 & 1042\\
    2 & M & 11 & 36 & 3 & 110969 & 691\\
    3 & F & 14 & 38 & 7 & 103777 & 1002\\
    4 & M & 22 & 42 & 3 & 107686 & 760\\
    5 & F & 7 & 39 & 5 & 114578 & 1158\\
    6 & F & 1.5 & 18 & 7 & 156677 & 669\\
    7 & F & 14.5 & 19 & 3 & 181777 & 925\\
    8 & M & 3.5 & 20 & 5 & 46187 & 1519\\
    9 & F & 10 & 19 & 3 & 208250 & 726\\
    10 & M & 3 & 25 & 7 & 114604 & 1047\\
    11 & F & 12 & 35 & 3 & 107368 & 1406\\
    12 & F & 2 & 24 & 13 & 60628 & 664\\
    13 & F & 3 & 33 & 8 & 133796 & 815\\
    14 & F & 9 & 26 & 7 & 64182 & 727\\
    15 & M & 16 & 40 & 14 & 89375 & 1944\\
    16 & F & 7 & 19 & 7 & 38989 & 638\\
    17 & F & 12 & 21 & 3 & 35390 & 893\\
    18 & F & 18 & 36 & 6 & 96573 & 917\\
    19 & F & 19 & 30 & 3 & 85776 & 836\\
    20 & F & 6 & 29 & 6 & 78707 & 741\\
    21 & F & 13 & 33 & 4 & 96572 & 799\\
    22 & F & 9 & 31 & 3 & 92985 & 804\\
    23 & F & 6 & 9 & 3 & 68109 & 713\\
    24 & N/A & N/A & 22 & 12 & 38989 & 929\\
    \bottomrule
    \end{tabular}
}
\label{tab:chbmit_summary}
\end{table}

All EEG signals in the CHB-MIT dataset were recorded at a sampling rate of \SI{256}{Hz} with a 16-bit resolution. These recordings conform to the International 10-20 system for EEG electrode placement and naming conventions.
The number of channels recorded varied among subjects, with a minimum of 23 channels used. In the cases of subjects 4 and 9, the dataset includes additional channels for ECG and vagal nerve stimulus. In this work, the experiments were based only the data acquired from the 23 channels used by all the subjects.
Subjects' recordings were organized into tracks, which are labeled according to whether they contain seizure activity or not. For tracks with seizures, the specific time intervals of the seizure occurrences were meticulously documented by the dataset's authors\footnote{For further details on the CHB-MIT dataset, see https://physionet.org/content/chbmit/1.0.0/}.

\paragraph{Data preprocessing}
Data were segmented into frames of one second of length 256 with 23 channels. 
Then, the dataset was filtered using a bandpass filter of \SIrange{0.5}{25}{Hz}, following \cite{shoeb2010application}. Following this, Z-score normalization was applied.
Finally, for each subject, a random sample consisting of 1-second windows totaling 10 minutes was drawn from each non-seizure track to test the models' ability to detect non-seizure cases. The remaining data were used to train the models. 
Details on the CHB-MIT dataset are provided in Table \ref{tab:chbmit_summary}.

\paragraph{Model selection}
The model selection procedure for the CHB-MIT dataset follows the methodology applied to the Bonn dataset, with a critical adjustment to accommodate the specific structure of the CHB-MIT dataset. Each subject in the CHB-MIT dataset is associated with multiple tracks, each approximately one hour in length. Thus, the 10-fold cross validation was substituted with a leave-one-out strategy tailored for this context, that will be referred as \textit{leave-one-track-out}. In this approach, five randomly selected non-seizure tracks are used; four tracks are employed for training the model, and the remaining track serves as the test set. To maintain clarity and focus in the analysis, detailed results will be provided exclusively for subject 1.

The search space involved in this stage is reported in Table \ref{tab:chbmit_search_space}. \begin{table}[ht!]
\caption{Search spaces for the grid search conducted during the model selection process on the CHB-MIT dataset.}
\begin{center}
\scalebox{.8}{
    \begin{tabular}{cc}
        \hline
        \textbf{Hyperparameter} & \textbf{Search Space}\\
        \hline
        Kernel Size & \{71, 81, 111, 131, 151\} \\
        Filters & $\{2^n\}, 2 \le n \le 8$\\
        Activation Function & \{ ReLU, Identity \} \\
        Latent Space Dimension & $\{2^n\}, 5 \le n \le 7$\\
        \hline
    \end{tabular}
}
\end{center}
\label{tab:chbmit_search_space}
\end{table}
To manage the increased computational effort of the model selection stage, necessitated by the larger dataset size, the search space was limited based on insights gained from the Bonn dataset analysis. 
In particular, the total number of configuration explored was 210. 

Similar to the results obtained on the Bonn dataset, the configuration with the lowest mean MSE across the tracks showed notable overlap with several other configurations, \textcolor{black}{requiring additional analyses using statistical hypothesis testing}. Thus, only those configurations whose mean MSE falls within one standard deviation of the lowest mean MSE were retained for further analysis. After this filtering, the top 67 configurations were selected for further detailed analysis. 

By applying the Shapiro-Wilk test to the results of each configuration, and using a significance level of $\alpha = 0.05$, none of the configurations rejected the null hypothesis, suggesting that the data from all configurations can be considered normally distributed. Consequently, an ANOVA test was conducted using the same significance level. This test also did not reject the null hypothesis, indicating that there were no significant differences among the configurations. This result implies that the performance across different configurations is statistically comparable under the conditions tested.

Also for this dataset, the best configuration was identified from the 67 configurations selected by prioritizing a low model complexity. 
Thus, the latent space was fixed to 128, 
The Identity function was chosen as activation function, and 4 filters were used with a kernel size of 71 was chosen.

Interestingly, the analyses conducted on the first five subjects, under identical conditions, led to the identification of the same optimal configuration using the same selection criterion. As a result, this configuration was adopted for subsequent experiments across all subjects.

\subsection{\textcolor{black}{Sensitivity analysis of SincVAE's hyperparameters}}
\textcolor{black}{
This subsection examines the sensitivity of hyperparameters for the SincVAE architecture, building on the results from the model selection stages, which aimed to identify the optimal hyperparameters for reconstructing non-anomalous data. 
The focus is on the Bonn dataset, as similar findings apply to the CHB-MIT dataset. 
Consistent with the model selection criteria applied to both datasets, the average MSE over 10-fold cross-validation is used as the comparison metric. 
Only the top configurations, which showed no significant difference in MSE performance based on the hypothesis testing in the previous subsection, are considered. 
For reference, Table \ref{tab:bonn_search_space} provides the search space details for the Bonn dataset. 
The impact of each hyperparameter setting is discussed through figures, where the left panel shows the distribution of the average MSE across the 10 folds, and the right panel illustrates the frequency of each model configuration.}

\begin{figure}[ht!]
    \centering
    \includegraphics[width=0.9\linewidth]{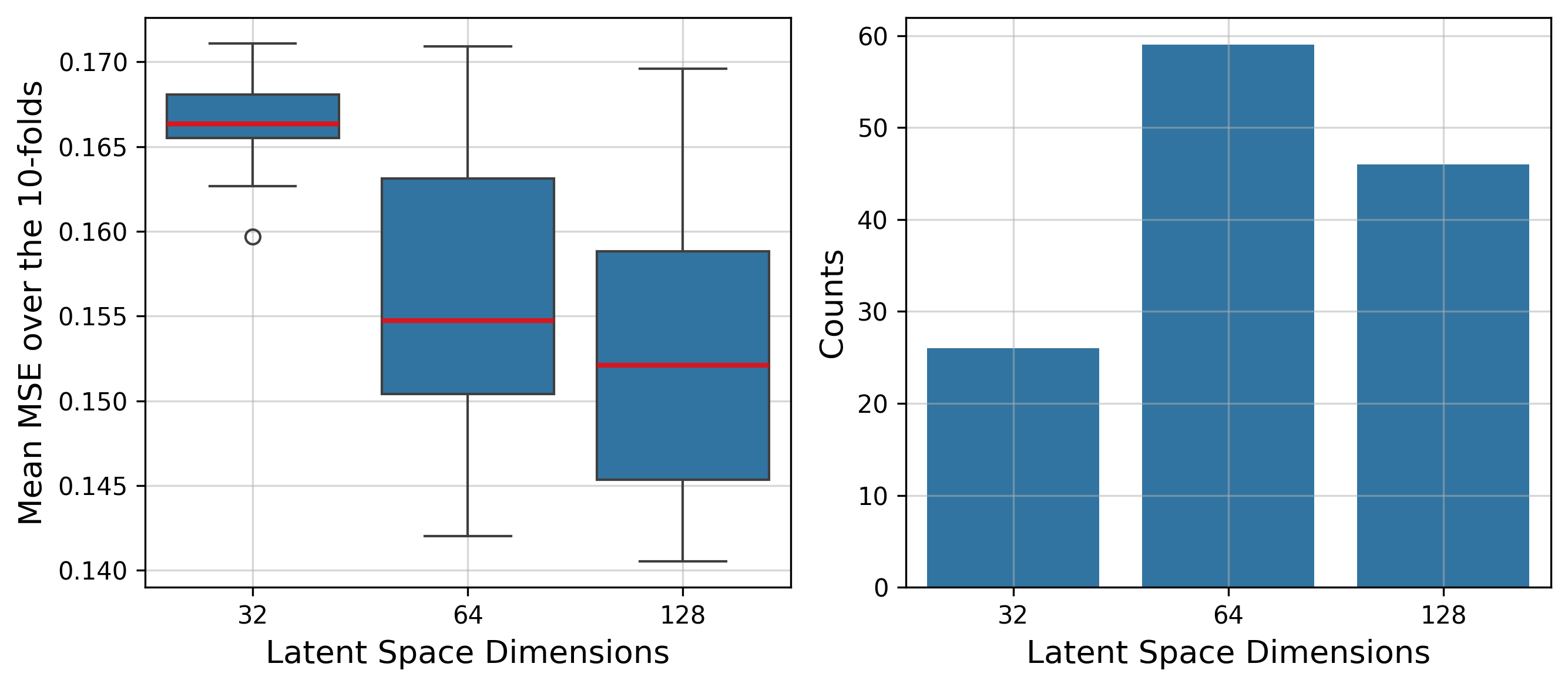}
    \caption{\textcolor{black}{Graphical representation of the mean MSE across 10 folds for the configurations in analysis, categorized by latent space dimensionality. The left panel shows a box plot illustrating the distribution of mean MSE values, while the right panel presents a bar plot describing the number of configurations having each latent dimension.}}
    \label{fig:bonn_latentdim}
\end{figure}

\textcolor{black}{\paragraph{Latent dimension} 
In the selected configurations, the dimensionality of the latent space is consistently set to 32, 64, or 128, as shown in Figure \ref{fig:bonn_latentdim}. 
Lower dimensionalities, such as 8 and 16, were found to significantly degrade network performance. 
This observation aligns with findings in the literature on AEs \cite{goodfellow2016deep}, suggesting that excessive compression prevents the VAE’s capacity to learn fundamental non-seizure patterns needed to achieve sufficient reconstruction accuracy for the input EEG recordings.
On the other hand, undercomplete AEs, where the latent dimension is smaller than the input dimension, are generally preferred as they force the encoder to extract the most salient features from the input data. However, this property diminishes as the latent dimension approaches the input dimension, leading to an overcomplete representation \cite{goodfellow2016deep}.
Since no significant difference in MSE performance was observed across the configurations with latent dimensions of 32, 64, or 128, a dimensionality of 32 is recommended for anomaly detection. This choice promotes more effective identification and representation of the critical features of non-seizure data, enhancing the model’s ability to distinguish it from anomalous seizure data.}

\begin{figure}[ht!]
    \centering
    \includegraphics[width=0.9\linewidth]{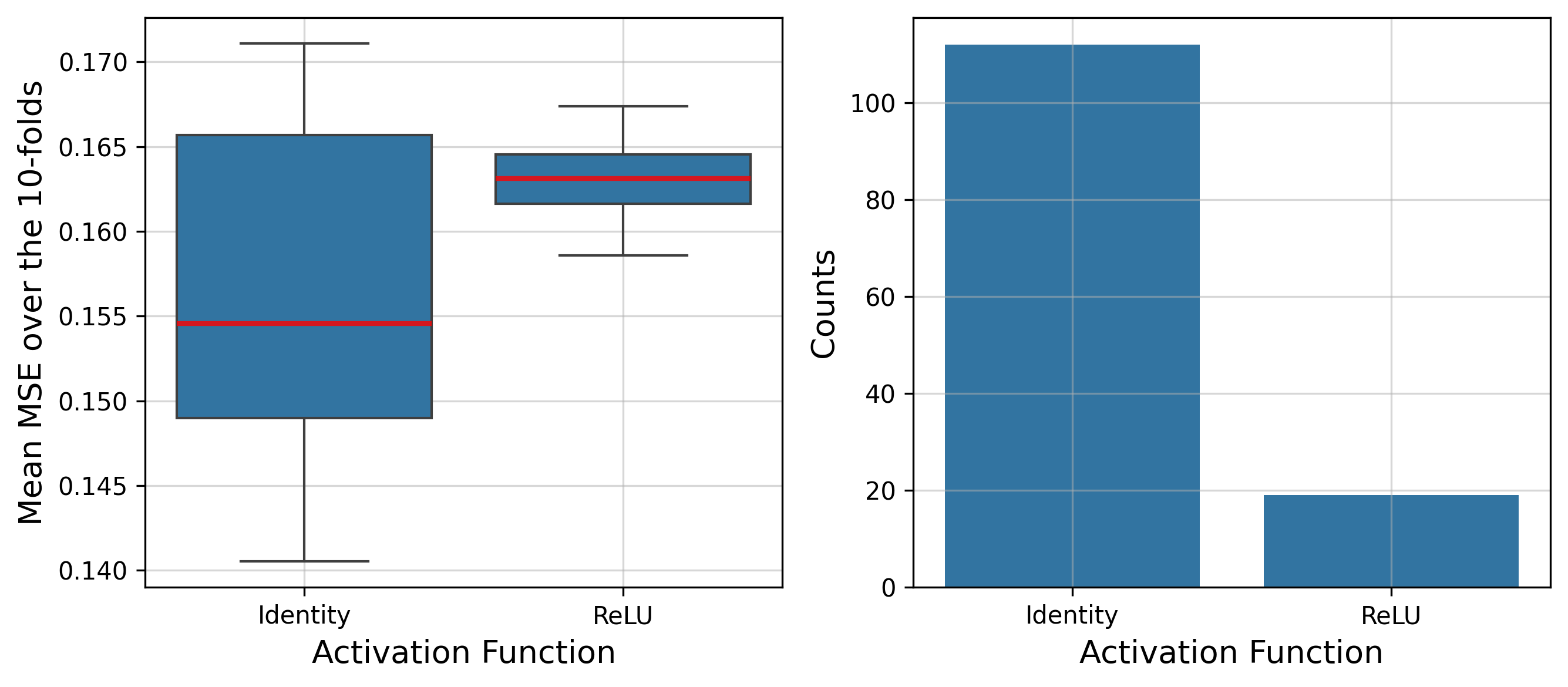}
    \caption{\textcolor{black}{Graphical representation of the mean MSE across 10 folds for the configurations in analysis, categorized by SincNet layer's activation function. The left panel shows a box plot illustrating the distribution of mean MSE values, while the right panel presents a bar plot describing the number of configurations having each activation function.}}
    \label{fig:bonn_actfn}
\end{figure}

\textcolor{black}{\paragraph{Activation function} 
None of the configurations use the Tanh activation function, relying exclusively on ReLU or Identity activations, as shown in Figure \ref{fig:bonn_actfn}. Notably, 112 out of 131 configurations incorporate the Identity function, consistent with the choice made by the SincNet authors in \cite{ravanelli2018speaker}. This strong preference suggests that preserving the original structure of the EEG recordings after the SincNet layer's decomposition may enhance the VAE's effectiveness more than applying non-linear transformations. In this scenario, the Identity function is also preferred since it introduces no additional complexity compared to non-linear transformations.}

\begin{figure}[ht!]
    \centering
    \includegraphics[width=0.9\linewidth]{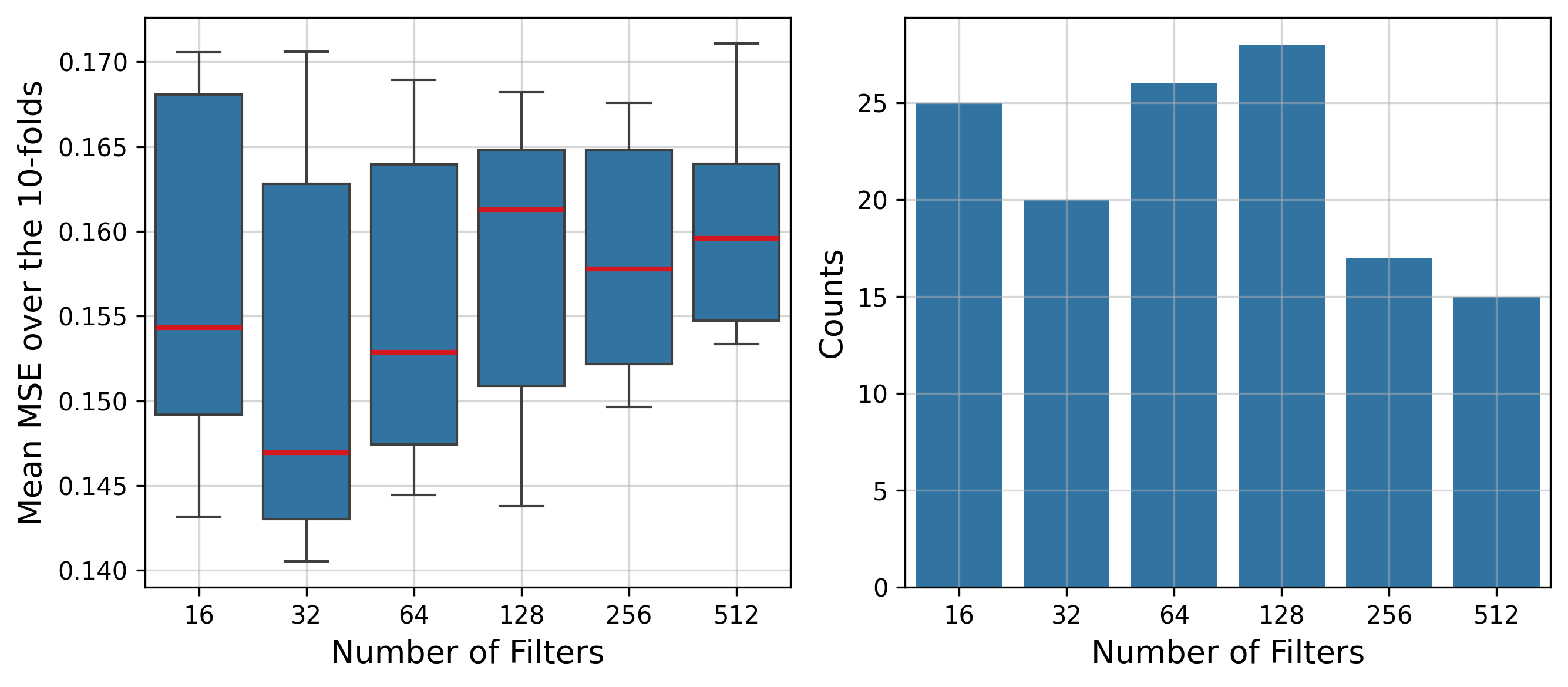}
    \caption{\textcolor{black}{Graphical representation of the mean MSE across 10 folds for the configurations in analysis, categorized by number of filters of the SincNet layer. The left panel shows a box plot illustrating the distribution of mean MSE values, while the right panel presents a bar plot describing the number of configurations having each number of filters in the SincNet layer.}}
    \label{fig:bonn_filters}
\end{figure}

\textcolor{black}{\paragraph{Filters} 
No configuration utilizes 2, 4 and 8 filters, as shown in Figure \ref{fig:bonn_filters}. This suggests that a minimal filter count is insufficient for effective feature extraction, which is crucial for the proper functioning of the VAE. Therefore, more detailed frequency band extraction is necessary to enhance performance. 
Among the different configurations, having 32 filters lead to slightly better performance.
However, it is important to remark that, based on the hypothesis tests, no significant performance differences were observed between the configurations.
Thus, the choice of the number of filters in the SincNet layer may be driven by factors such as inference speed, model complexity, or the need for a finer resolution in signal decomposition. In this work, as noted in the previous subsection, the primary goal was to maintain low model complexity, which is why 16 filters were selected.}

\begin{figure}[ht!]
    \centering
    \includegraphics[width=0.9\linewidth]{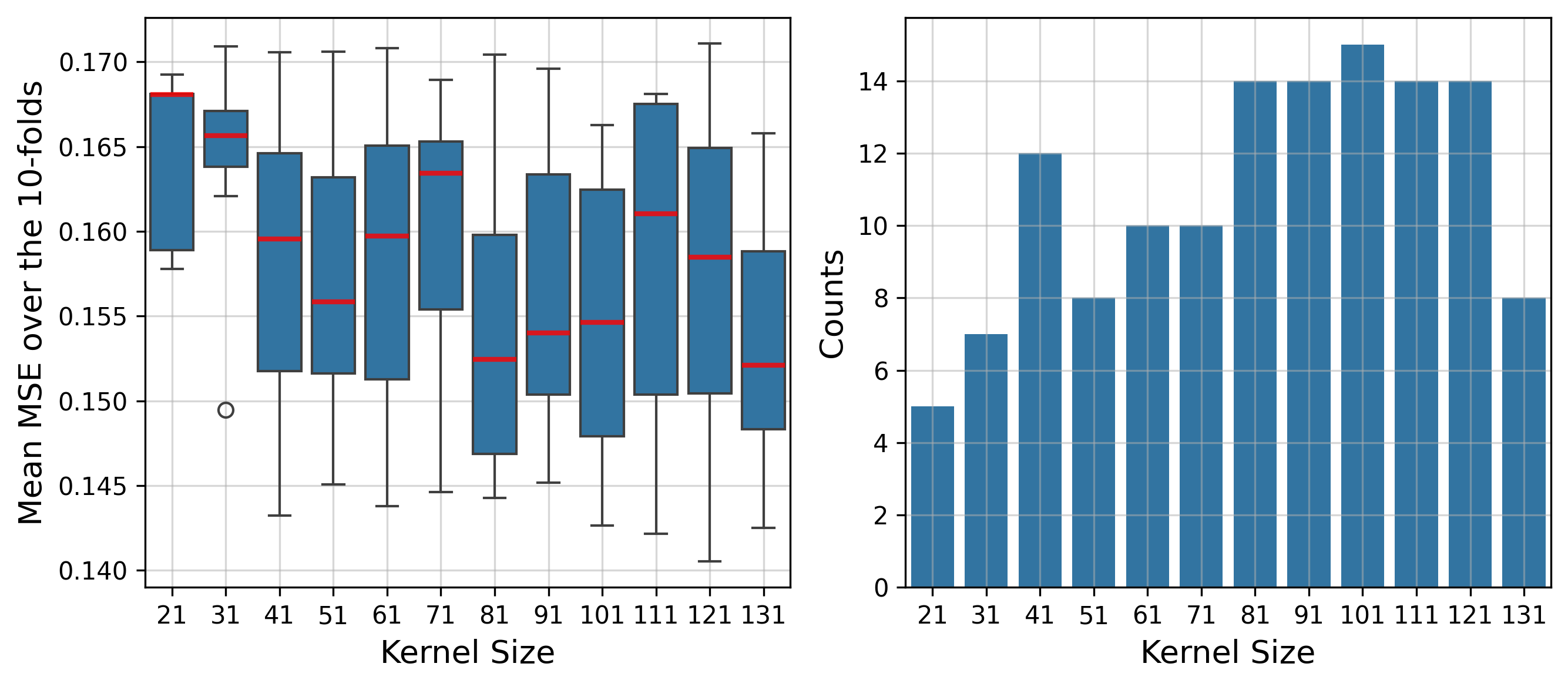}
    \caption{\textcolor{black}{Graphical representation of the mean MSE across 10 folds for the configurations in analysis, categorized by the kernel size. The left panel shows a box plot illustrating the distribution of mean MSE values, while the right panel presents a bar plot describing the number of configurations having each kernel size.}}
    \label{fig:bonn_kernels}
\end{figure}

\textcolor{black}{\paragraph{Kernel size} 
None of the configurations use kernel sizes of 3, 5, 7, or 11, as shown in Figure \ref{fig:bonn_kernels}. This suggests that these kernel sizes may lead to a low resolution in the Hamming windows applied by each filter during signal processing. Larger kernel sizes might offer better resolution, allowing for more detailed frequency band extraction. As for the number of filters, the choice of kernel size must ultimately be dictated by the specific needs of the task, balancing between resolution and computational efficiency.}

\section{Results}
\label{sec:results}
In this section, the results of the seizure detection experiments conducted on both the Bonn and CHB-MIT datasets are presented. For each dataset, the models were trained using the configurations identified during their respective model selection stages. During each training phase, \SI{20}{\percent} of the training data was randomly sampled and used as the validation set. 

It is worth noting that for each result, the model without the SincNet layer is referred to as VAE; whereas, the model incorporating a SincNet layer on top of the encoder network, which is the proposal of this work, is referred to as SincVAE.
It is crucial to mention that the primary focus of these experiments is to integrate SincNet into the VAE framework and evaluate its effectiveness in addressing the problem, rather than achieving new state-of-the-art results on the datasets.

\subsection{The Bonn Dataset}
The results of the seizure detection for both declared comparisons, i.e. Set A vs Set E and Set B vs Set E, will be discussed in this subsection. In both scenarios, the test set includes 2760 samples, which exhibits a significant imbalance with 460 samples labeled as non-seizure and 2300 labeled as seizure. This imbalance was taken in account during the evaluations.

\begin{figure}[ht!]
    \begin{minipage}{1\textwidth}
        \centering 
        \includegraphics[width=0.45\textwidth]{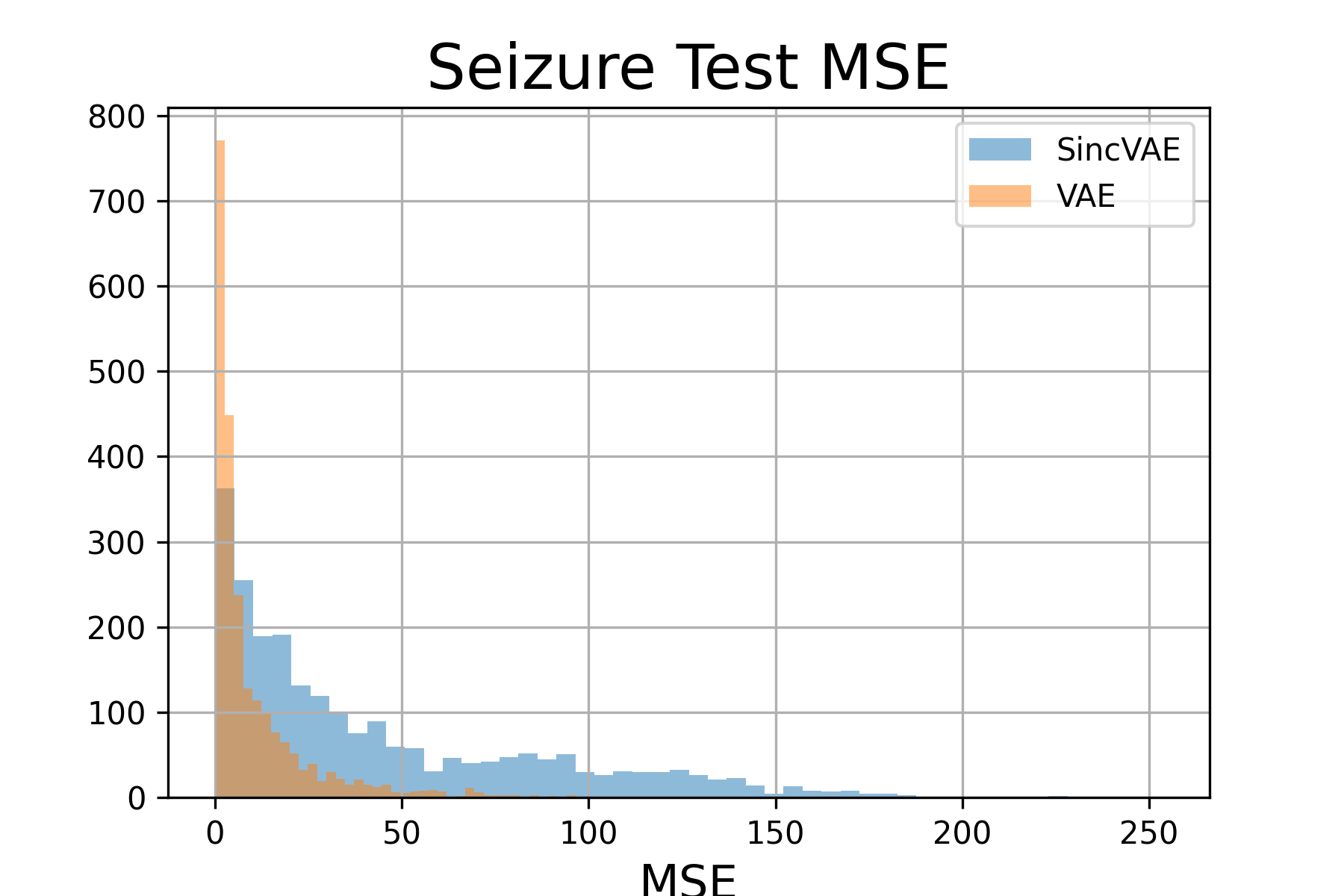} 
        \hfill 
        \includegraphics[width=0.45\textwidth]{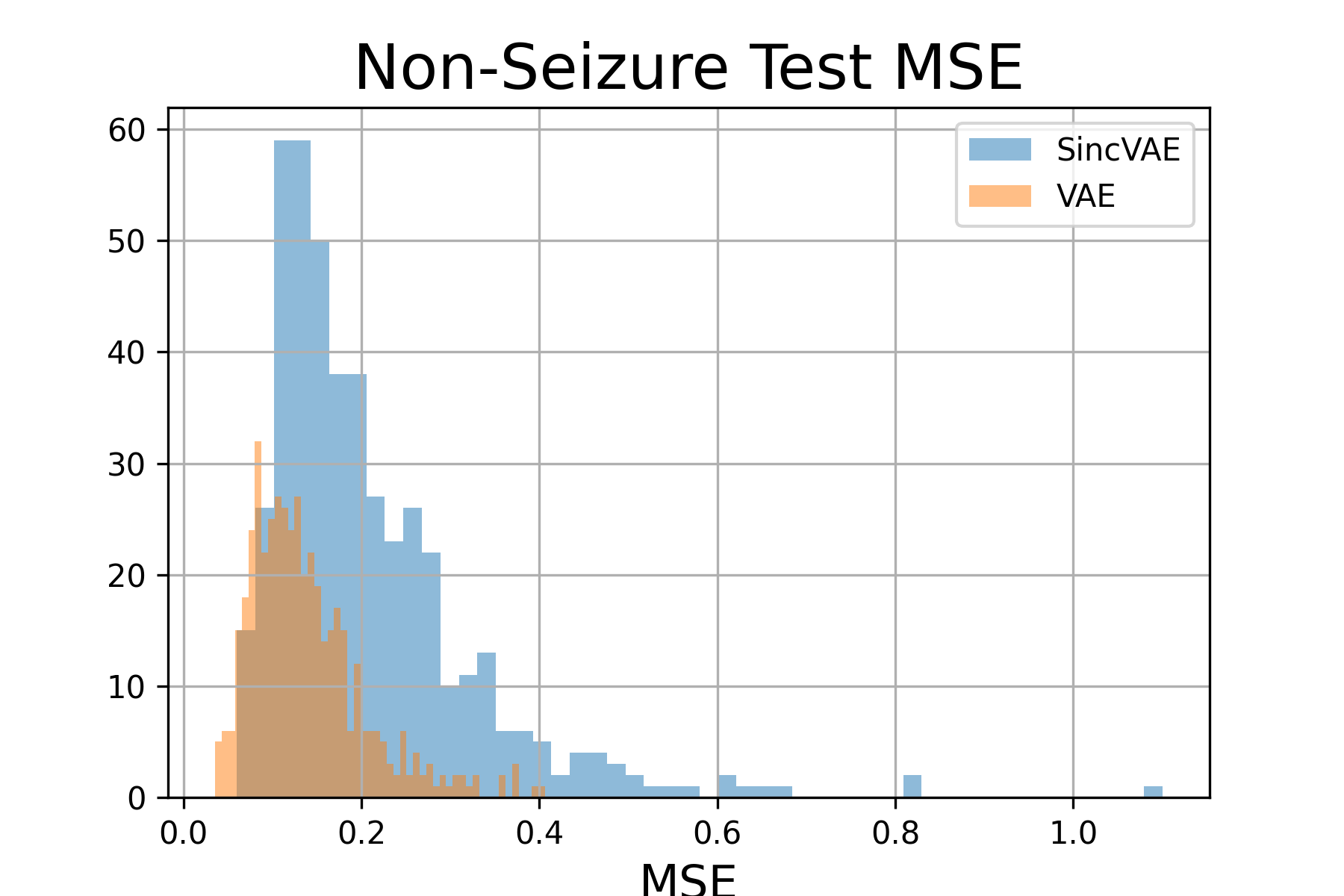} 
        \hfill 
        \includegraphics[width=0.45\textwidth]{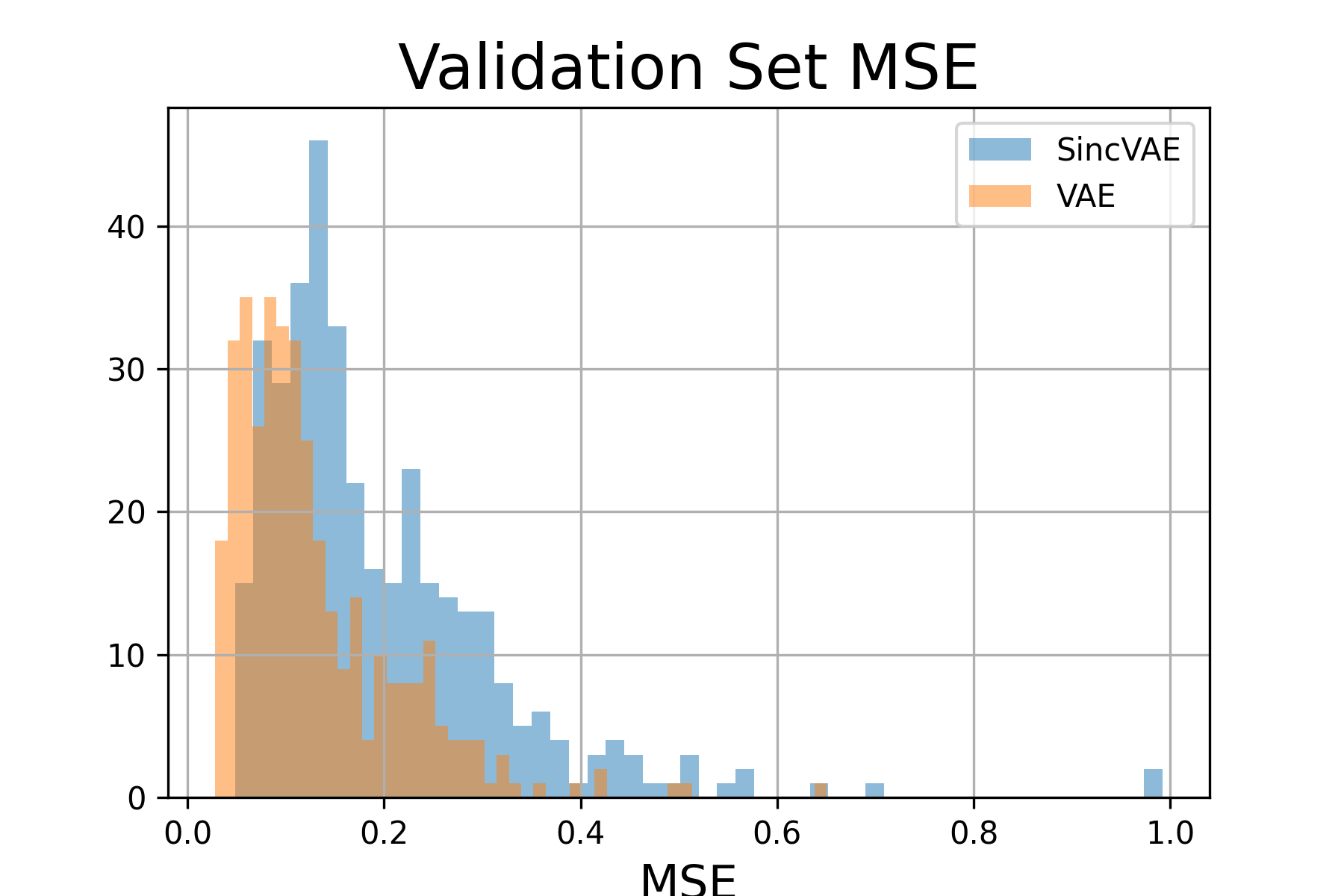} 
        \caption{Graphical representation of seizure (top-left) and non-seizure (top-right) test data, and validation (bottom) MSE distribution for VAE (orange) and SincVAE (blue). Non-seizure test data are drawn from Set A. 
        \textcolor{black}{Both models produce right-skewed distributions due to the focus on minimizing reconstruction error. While they show similar performance on non-seizure data, the VAE has a slightly lower mode. On seizure data, the VAE shows better reconstruction, whereas SincVAE exhibits higher variability with a pronounced right tail. The larger MSE gap in SincVAE suggests a clearer distinction between seizure and non-seizure conditions, potentially facilitating threshold-based classification.}
        }
        \label{fig:bonn_error_distribution_setA}
    \end{minipage}
\end{figure}

\begin{figure}[ht!]
    \vfill
    \begin{minipage}{1\textwidth}
        \centering 
        \includegraphics[width=0.45\textwidth]{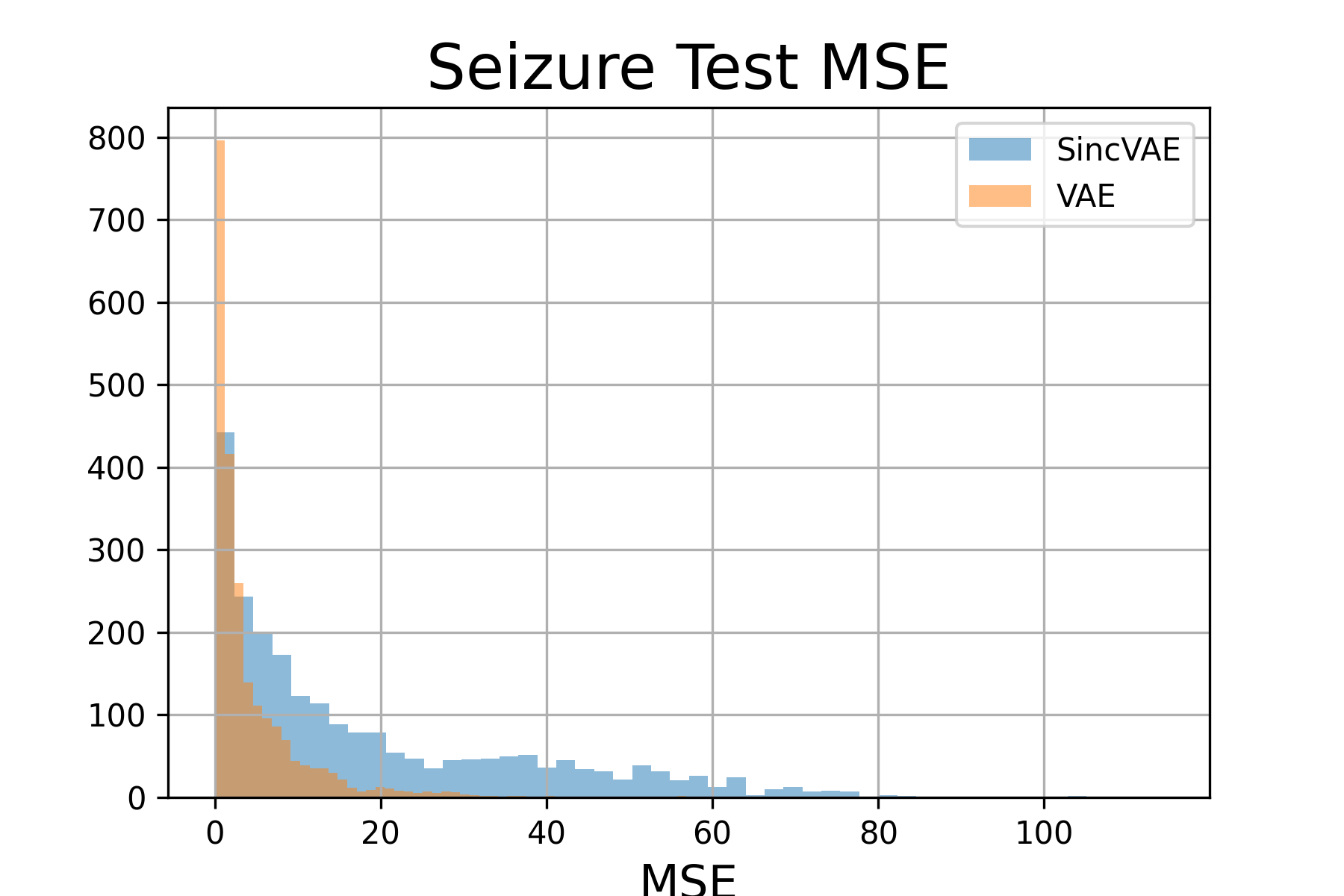} 
        \hfill 
        \includegraphics[width=0.45\textwidth]{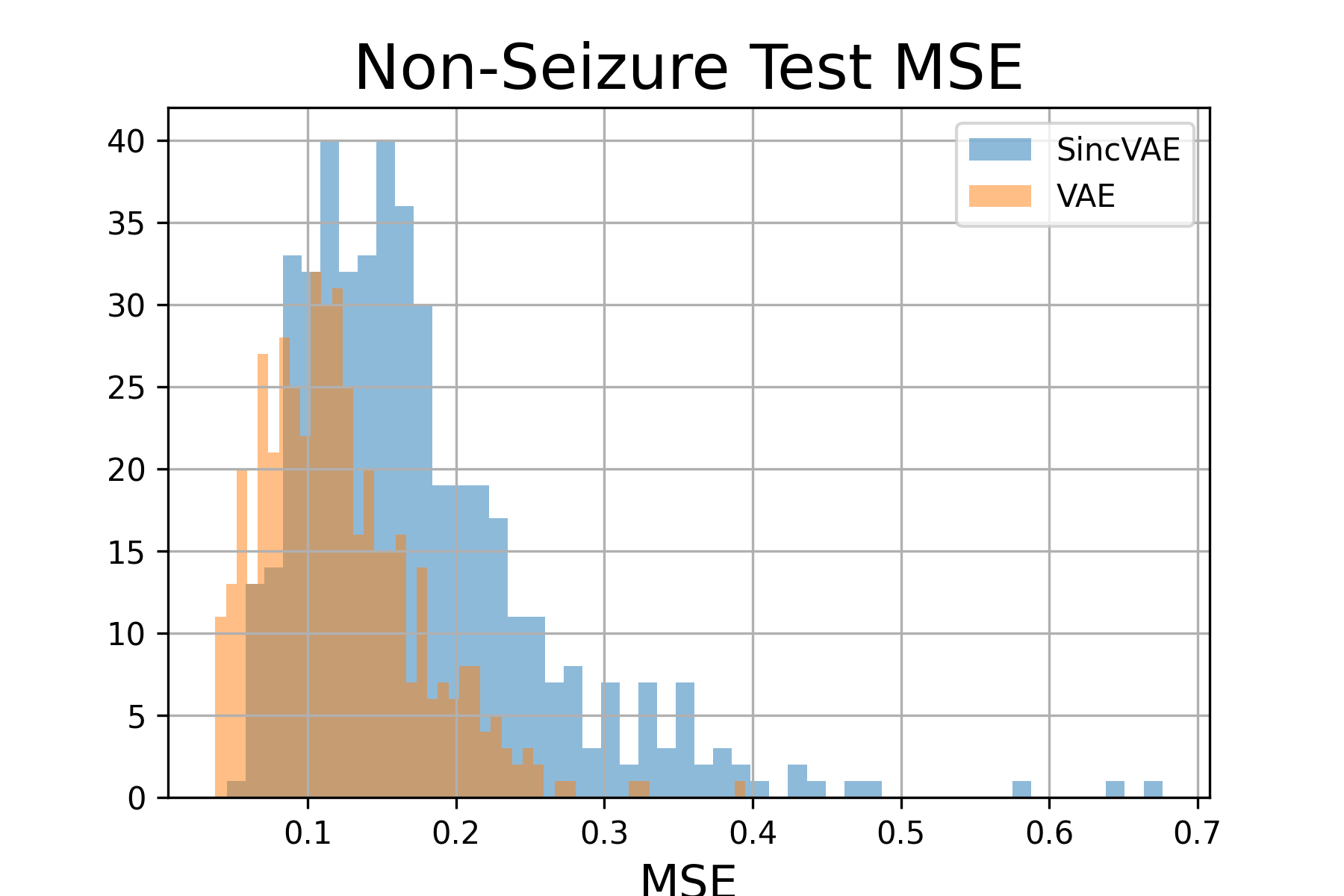} 
        \hfill 
        \includegraphics[width=0.45\textwidth]{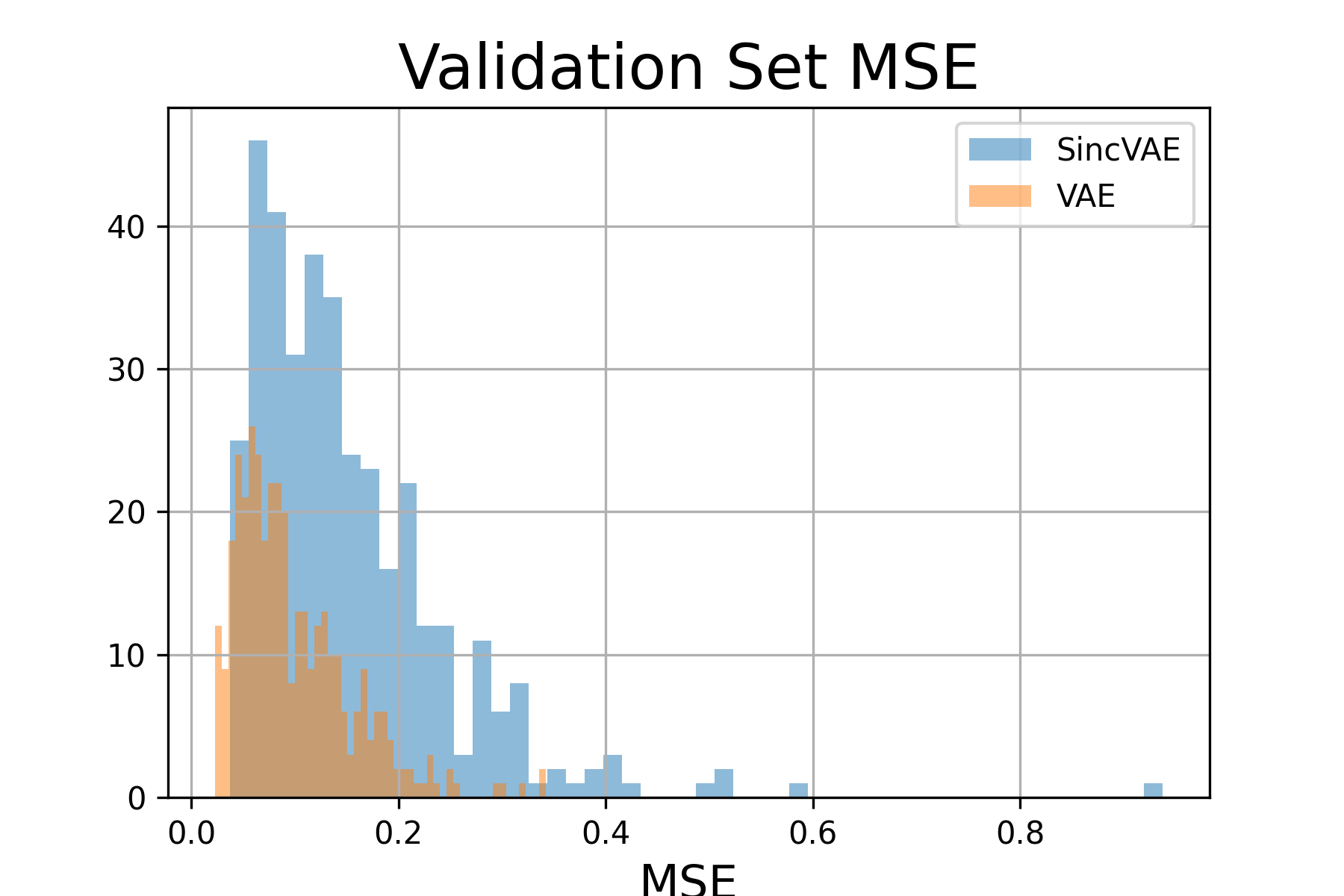} 
        \caption{Graphical representation of seizure (top-left) and non-seizure (top-right) test data, and validation (bottom) MSE distribution for VAE (orange) and SincVAE (blue). Non-seizure test data are drawn from Set B.
        \textcolor{black}{Similarly to Figure \ref{fig:bonn_error_distribution_setA}, both models produce right-skewed distributions due to the focus on minimizing reconstruction error. While they show similar performance on non-seizure data, the VAE has a slightly lower mode. On seizure data, the VAE shows better reconstruction, whereas SincVAE exhibits higher variability with a pronounced right tail. The larger MSE gap in SincVAE suggests a clearer distinction between seizure and non-seizure conditions, potentially facilitating threshold-based classification.}}
        \label{fig:bonn_error_distribution_setB}
    \end{minipage}
\end{figure}

Figure \ref{fig:bonn_error_distribution_setA} shows the MSE distributions for the seizure test data (top-left), non-seizure test data (top-right), and the validation data (bottom), comparing the performance of the VAE (orange) and SincVAE (blue) models. 
Both models produce right-skewed distributions, consistent with the optimization process focused on minimizing reconstruction error. For the non-seizure test data, the models exhibit similar performance, with the VAE showing a slightly lower mode. This behavior aligns with the distribution observed in the validation set, demonstrating the models' effectiveness in detecting non-seizure patterns.
However, on the seizure data, the VAE model demonstrates better reconstruction performance, while the SincVAE model exhibits higher variability, evident through its pronounced right tail and elevated reconstruction errors.
SincVAE's larger discrepancy between its seizure and non-seizure MSE values could indicate a better distinction between the seizure and non-seizure conditions, which could ease the threshold-based classification. In contrast, the narrower MSE range of VAE could make this differentiation more challenging. Similar conclusions can be drawn on Set B (see Figure \ref{fig:bonn_error_distribution_setB}).

To assess the classification performance of the models, a classification threshold must be determined. This threshold is typically based on the MSE values derived from the trained model on the validation or training set, as suggested by various studies \cite{givnan2022anomaly,tun2020network,elsayed2020detecting}, and tailored to meet specific user requirements. For this study, the classification threshold was selected using the following criteria:
\begin{itemize}
    \item The maximum MSE from the validation set, defined as $t_1$;
    \item The 95th percentile of validation MSE, defined as $t_2$;
\end{itemize}

\begin{figure}[!ht]
    \begin{minipage}{1\textwidth}
        \centering
\scalebox{.8}{
\begin{tabular}{c||ccc|ccc}
    \hline
    \multirow{2}{*}{Method} & \multicolumn{3}{c}{Threshold set to $t_1$ (\%)} & \multicolumn{3}{c}{Threshold set to $t_2$ (\%)}\\
    \cline{2-7}
    & F1 & Precision & Recall & F1 & Precision & Recall \\
    \hline
    

    SincVAE & \textbf{99.8} & 100 & \textbf{99.6} & \textbf{99.4} & 98.9 & \textbf{100} \\
    VAE & 95.6 & 100 & 91.7 & 99.2 & \textbf{99.2} & 99.2 \\
    
    \hline
\end{tabular}
}

    \end{minipage}
    \vfill
    \vspace{0.5cm}
    \begin{minipage}{1\textwidth}
        \centering
        \begin{tabular}{ccc}
            & Threshold set to $t_1$ & Threshold set to $t_2$ \\ 
            \multirow{2}{*}[5em]{SincVAE} & 
            \begin{subfigure}{0.35\textwidth}
                \includegraphics[width=\linewidth]{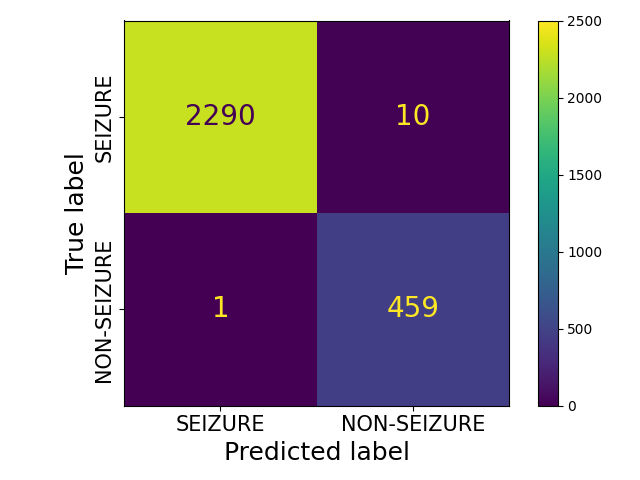}
            \end{subfigure} &
            \begin{subfigure}{0.35\textwidth}
                \includegraphics[width=\linewidth]{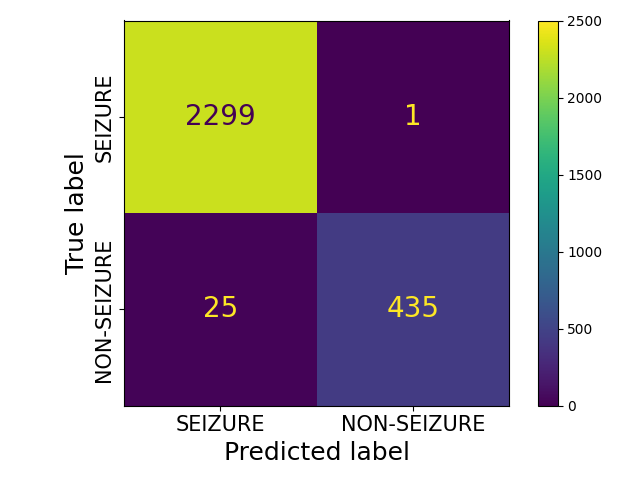}
            \end{subfigure} \\
            \multirow{2}{*}[5em]{VAE} & 
            \begin{subfigure}{0.35\textwidth}
                \includegraphics[width=\linewidth]{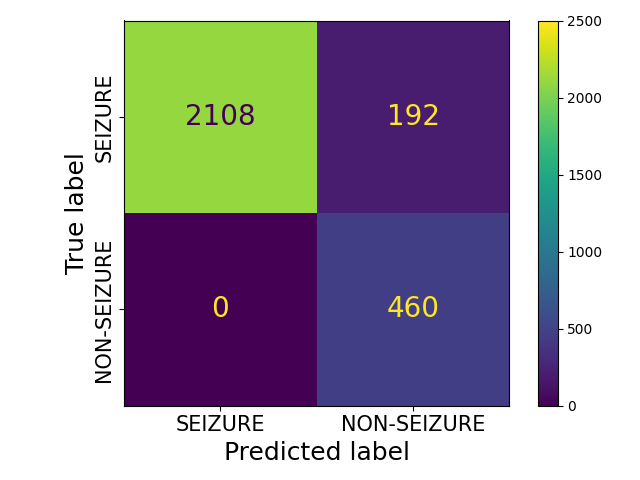}
            \end{subfigure} &
            \begin{subfigure}{0.35\textwidth}
                \includegraphics[width=\linewidth]{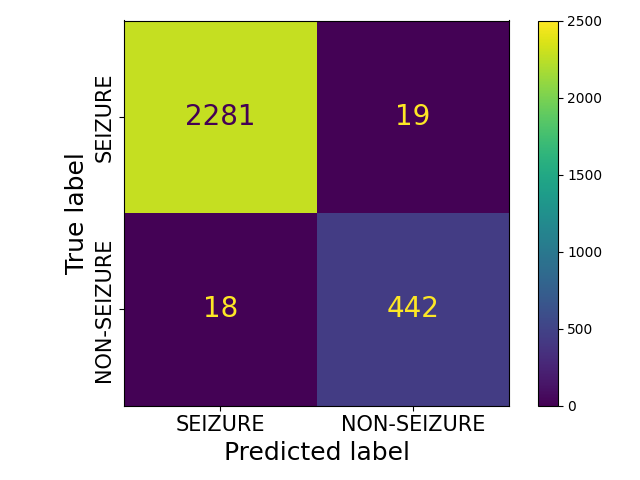}
            \end{subfigure} \\
        \end{tabular}
        \caption{Classification results shown as confusion matrices of SincVAE (first row) and VAE (second row) under the two selected thresholds $t_1$ (first column) and $t_2$ (second column) on the test data of the case Set A vs Set E. F1, Precision and Recall for each case are reported on top of the confusion matrices in percentages.}
        \label{fig:bonn_confmatrix_comparison_setA}
    \end{minipage}
\end{figure}

\textcolor{black}{Considering as true positive an anomalous sample classified as seizure class, and as true negative a non-anomalous sample classified as non-seizure class, the classification performance of the models over the test set was evaluated using the following metrics \cite{hastie2009elements}:
\begin{enumerate}
    \item Recall, defined as:
    \begin{equation}
        \textrm{Recall} = \frac{\mathrm{\# True\ Positives}}{\mathrm{\#True\ Positives} + \mathrm{\#False\ Negatives}},
    \end{equation}
    that measures the ability of the model to correctly identify all positive instances (e.g., seizure events);
    \item Precision, defined as:
    \begin{equation}
        \textrm{Precision} = \frac{\mathrm{\#True\ Positives}}{\mathrm{\#True\ Positives} + \mathrm{\#False\ Positives}},
    \end{equation}
    that measures the accuracy of the positive predictions made by the model;
    \item F1, defined as:
    \begin{equation}
        \textrm{F1} = 2 \times \frac{\mathrm{Precision} \times \mathrm{Recall}}{\mathrm{Precision} + \mathrm{Recall}},
    \end{equation}
    that provides a single metric that captures both the completeness and accuracy of the model’s predictions by balancing the trade-off between recall and precision.
\end{enumerate}
In this work, the F1 score was chosen as the primary metric because it offers a balanced evaluation of the model's performance by accounting for both recall and precision. This makes it especially appropriate for scenarios involving imbalanced data \cite{hastie2009elements}.
}
Figure \ref{fig:bonn_confmatrix_comparison_setA} shows the confusion matrices and F1, Precision and Recall scores for both SincVAE and VAE, evaluated under the two thresholds $t_1$ and $t_2$, on the test data for the Set A vs Set E case.
By focusing on the confusion matrices, both models demonstrate to correctly identify non-seizure instances, as reflected by high values of true negatives across both thresholds.
VAE tends to generate fewer false positives: using $t_1$ the results are comparable, indeed SincVAE detects the \SI{0.2}{\percent} of the non-seizure data as seizure against the \SI{0}{\percent} detected by VAE, while using $t_2$ SincVAE detects the \SI{5.4}{\percent} of the non-seizure data as seizure against the \SI{3.9}{\percent} detected by VAE.
Notably, SincVAE shows fewer false negatives compared to the VAE model: using $t_1$, SincVAE detects the \SI{0.4}{\percent} of the seizure data as non-seizure against the \SI{8.3}{\percent} detected by VAE, while using $t_2$, SincVAE detects the \SI{0}{\percent} of the seizure data as non-seizure against the \SI{0.8}{\percent} detected by VAE. This suggests that SincVAE may be more precise in predicting seizures, aligning with the observations made in Figure \ref{fig:bonn_error_distribution_setA} and Figure \ref{fig:bonn_error_distribution_setB}.
In terms of scores, the performance of SincVAE appears to be superior to that of VAE on this experimental case: SincVAE demonstrates higher F1 scores, indicating it is more reliable for seizure detection across the chosen thresholds. Specifically, SincVAE shows improvements of $+$\SI{4.2}{\percent} F1 on $t_1$ and $+$\SI{0.2}{\percent} F1 on $t_2$.

\begin{figure}[ht!]
    \begin{minipage}{1\textwidth}
        \centering
%
%

\scalebox{.8}{
\begin{tabular}{c||ccc|ccc}
    \hline
    \multirow{2}{*}{Method} & \multicolumn{3}{c}{Threshold set to $t_1$ (\%)} & \multicolumn{3}{c}{Threshold set to $t_2$ (\%)}\\
    \cline{2-7}
    & F1 & Precision & Recall & F1 & Precision & Recall \\
    \hline
    

    SincVAE & \textbf{96.7} & 100 & \textbf{93.6} & \textbf{98.9} & \textbf{98.5} & \textbf{99.4} \\
    VAE & 94.8 & 100 & 90.2 & 98.0 & 98.0 & 97.9 \\

    \hline
\end{tabular}
}

    \end{minipage}
    \vfill
    \vspace{0.5cm}
    \centering
    \begin{tabular}{ccc}
        & Threshold set to $t_1$ & Threshold set to $t_2$ \\ 
        \multirow{2}{*}[5em]{SincVAE} & 
        \begin{subfigure}{0.35\textwidth}
            \includegraphics[width=\linewidth]{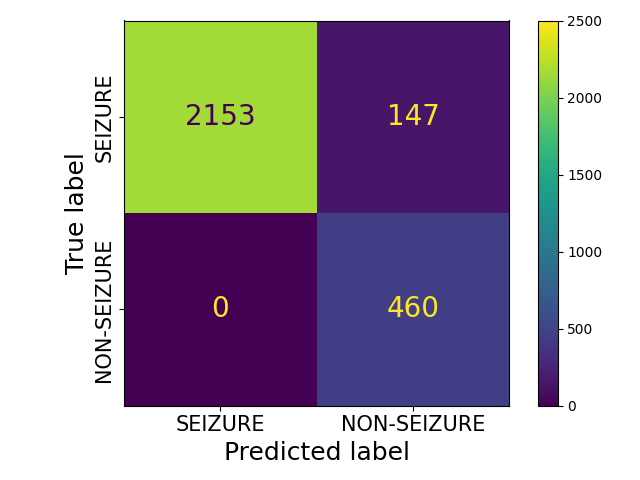}
        \end{subfigure} &
        \begin{subfigure}{0.35\textwidth}
            \includegraphics[width=\linewidth]{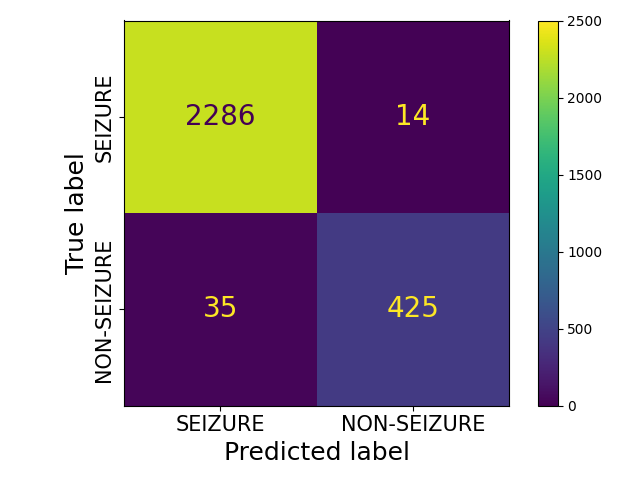}
        \end{subfigure} \\
        \multirow{2}{*}[5em]{VAE} & 
        \begin{subfigure}{0.35\textwidth}
            \includegraphics[width=\linewidth]{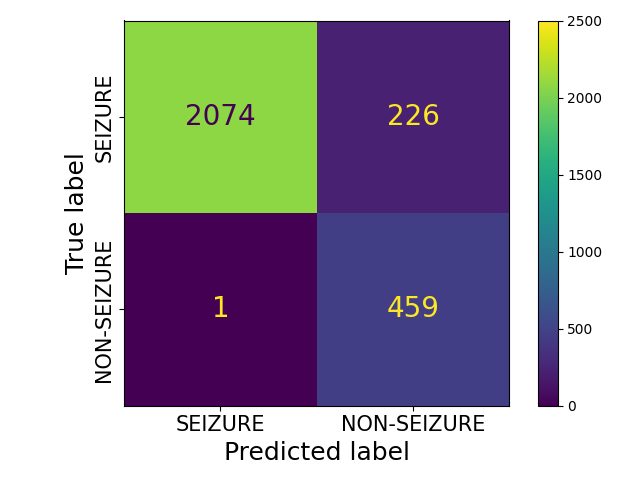}
        \end{subfigure} &
        \begin{subfigure}{0.35\textwidth}
            \includegraphics[width=\linewidth]{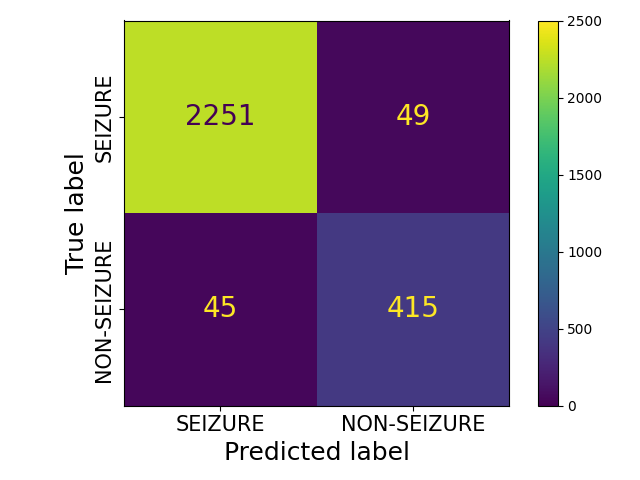}
        \end{subfigure} \\
    \end{tabular}
    \caption[Confusion matrices on the Bonn Dataset, Set B vs Set E case]{Confusion matrices obtained from the classification results of SincVAE (first row) and VAE (second row) under the two selected thresholds $t_1$ (first column) and $t_2$ (second column) on the case Set B vs Set E. F1, Precision and Recall for each case are reported on top of the confusion matrices in percentages.}
    \label{fig:bonn_confmatrix_setB}
\end{figure}

\begin{figure}[ht]
    \centering
    \begin{subfigure}{0.47\textwidth}
        \centering
        \includegraphics[width=\linewidth]{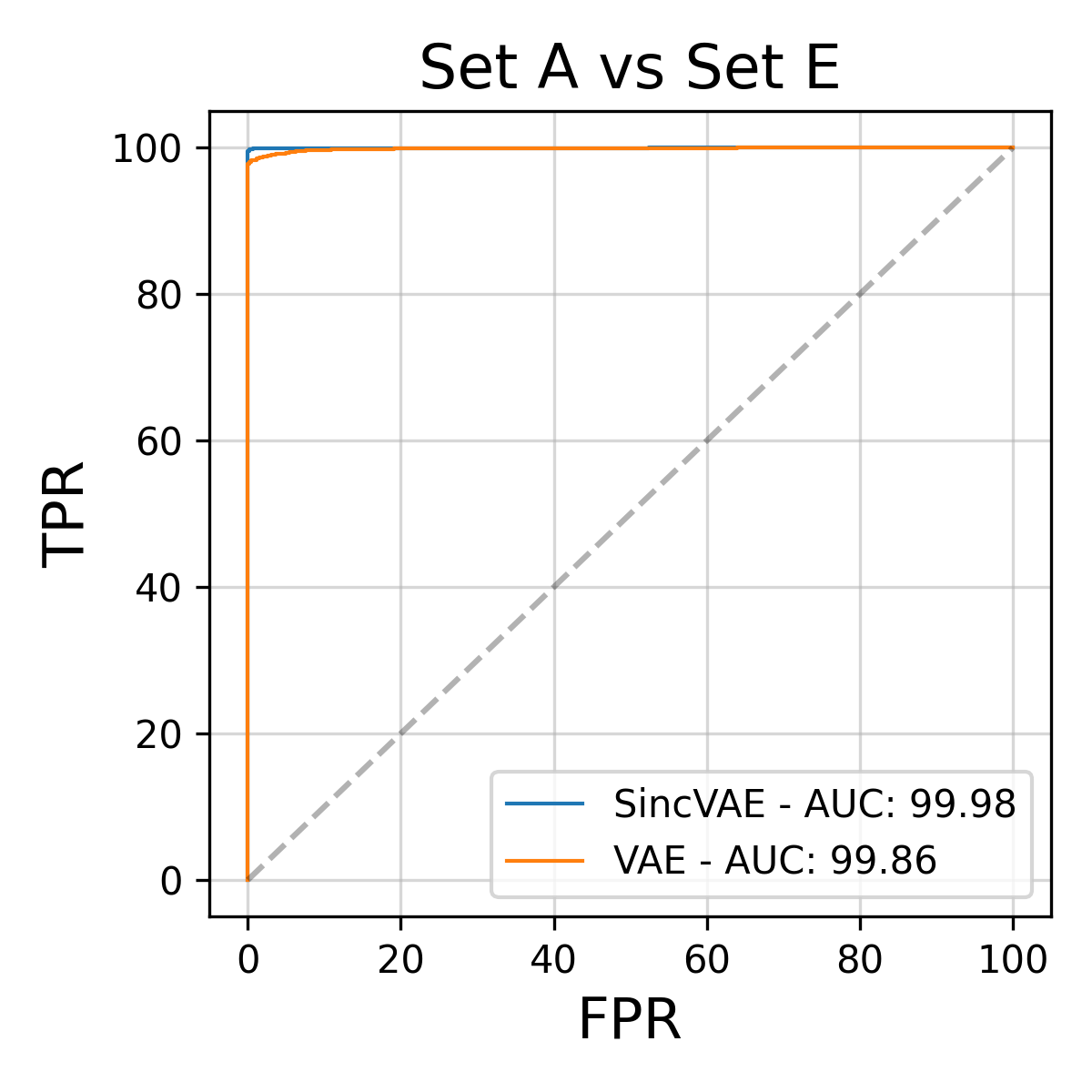}
    \end{subfigure}
    \hfill
    \begin{subfigure}{0.47\textwidth}
        \centering
        \includegraphics[width=\linewidth]{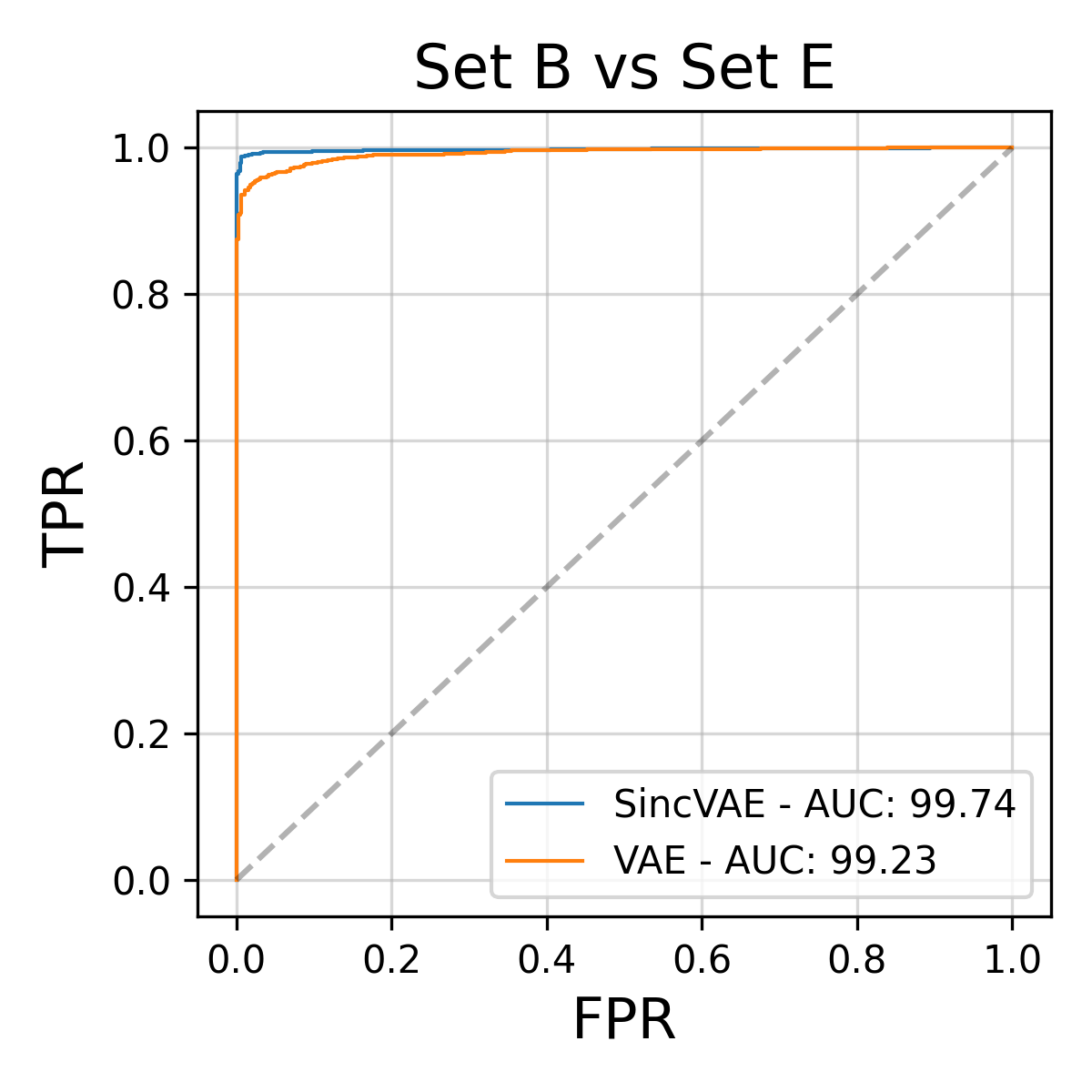}
    \end{subfigure}
    \caption{Comparison of ROC curves for SincVAE and VAE across the Set A vs Set E case (left) and the Set B vs Set E case (right). The AUC values are reported in the legend of each case, for each architecture, in percentage.}
    \label{fig:roc_comparison_bonn}
\end{figure}

Regarding the case Set B vs Set E, 
Figure \ref{fig:bonn_confmatrix_setB} shows the confusion matrices and the same classification scores.
By focusing on the confusion matrices, SincVAE consistently maintains high performance across both thresholds, demonstrating a lower number of both false positives and false negatives. 
In terms of false negatives, using $t_1$, SincVAE detects the \SI{6.4}{\percent} of the seizure data as non-seizure, against the \SI{9.8}{\percent} detected by VAE; using $t_2$, SincVAE detects the \SI{0.6}{\percent} of the seizure data as non-seizure, against the \SI{2.1}{\percent} detected by VAE. In terms of false positives instead, using $t_1$ the performance is comparable, indeed SincVAE detects the \SI{0}{\percent} of the non-seizure data as seizure against the \SI{0.2}{\percent} detected by VAE; using $t_2$, SincVAE detects the \SI{7.6}{\percent} of the non-seizure data as seizure, against the \SI{9.8}{\percent} detected by VAE.
This indicates a more robust ability of SincVAE to accurately differentiate between seizure and non-seizure events.
By focusing on the classification scores, the SincVAE model consistently outperforms in terms of F1 score and recall across both scenarios and thresholds, indicating robust predictive capabilities, particularly in minimizing missed seizures, which could be a critical aspect for clinical applications, due to its impact on enhancing patient safety and treatment efficacy. In particular, SincVAE shows improvements of $+$\SI{1.9}{\percent} F1 on $t_1$ and $+$\SI{0.9}{\percent} F1 on $t_2$.
\textcolor{black}{Figure \ref{fig:roc_comparison_bonn} shows the receiver operating characteristic (ROC) \cite{hastie2009elements} curve and the area under the curve (AUC) \cite{hastie2009elements} of SincVAE and VAE over the two cases in analysis. 
The ROC curve is a graphical representation of a classifier's performance across different threshold values. It plots the True Positive Rate (TPR) against the False Positive Rate (FPR), defined as:
\begin{align}
\mathrm{TPR,\ or\ Recall} &= \frac{\mathrm{\# True\ Positives}}{\mathrm{\# True\ Positives} + \mathrm{\# False\ Negatives}},\\ 
\mathrm{FPR} &= \frac{\mathrm{\# False\ Positives}}{\mathrm{\# False\ Positives} + \mathrm{\# True\ Negatives}}.
\end{align}
The AUC quantifies the overall performance of the model by measuring the area under the ROC curve. In percentage, an AUC value of \SI{100}{\percent} indicates a perfect classifier, while a value of \SI{50}{\percent} suggests no better performance than random chance. Since AUC is threshold-independent, it is particularly useful for comparing models and evaluating their discriminative ability across varying decision boundaries.
From Figure \ref{fig:roc_comparison_bonn}, it can be noticed that, in the case Set A vs Set E, SincVAE reaches an AUC of \SI{99.98}{\percent} while VAE an AUC of \SI{99.86}{\percent}, while in the case Set B vs Set E, SincVAE reaches an AUC of \SI{99.74}{\percent} while VAE an AUC of \SI{99.23}{\percent}. These results confirm that SincVAE consistently outperforms VAE in terms of AUC, highlighting its superior ability to discriminate between the classes in both scenarios.}

\begin{figure}[ht!]
    \centering
    \scalebox{0.85}{
        \includegraphics[width=\textwidth]{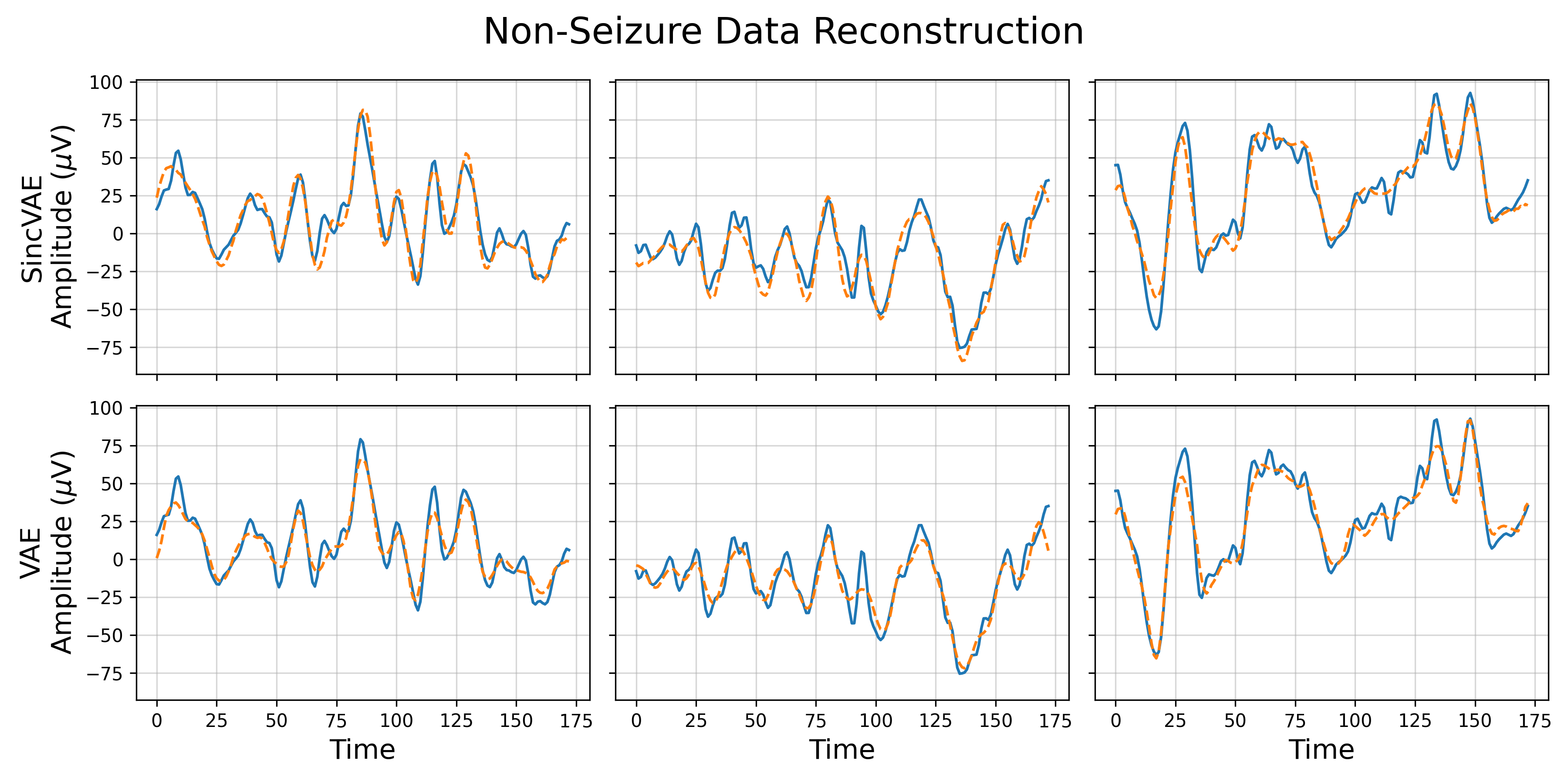}
    }
    \caption{Graphical representation of three randomly selected non-seizure samples drawn from Set A, reconstructed using SincVAE (first row) and VAE (second row). In each plot, the original input is represented by a solid blue line, while its reconstruction is represented by a dashed orange line.}
    \label{fig:bonn_nonseizurereconstruction}
\end{figure}

Figure \ref{fig:bonn_nonseizurereconstruction} displays the EEG recording reconstructions for three randomly selected non-seizure samples from Set A, using both the SincVAE and VAE models. The reconstructions of these non-seizure samples appear similarly accurate across both networks, suggesting that each model is effectively capturing and replicating EEG patterns associated with non-seizure states. 

\begin{figure}[ht!]
    \centering
    \scalebox{0.85}{
        \includegraphics[width=\textwidth]{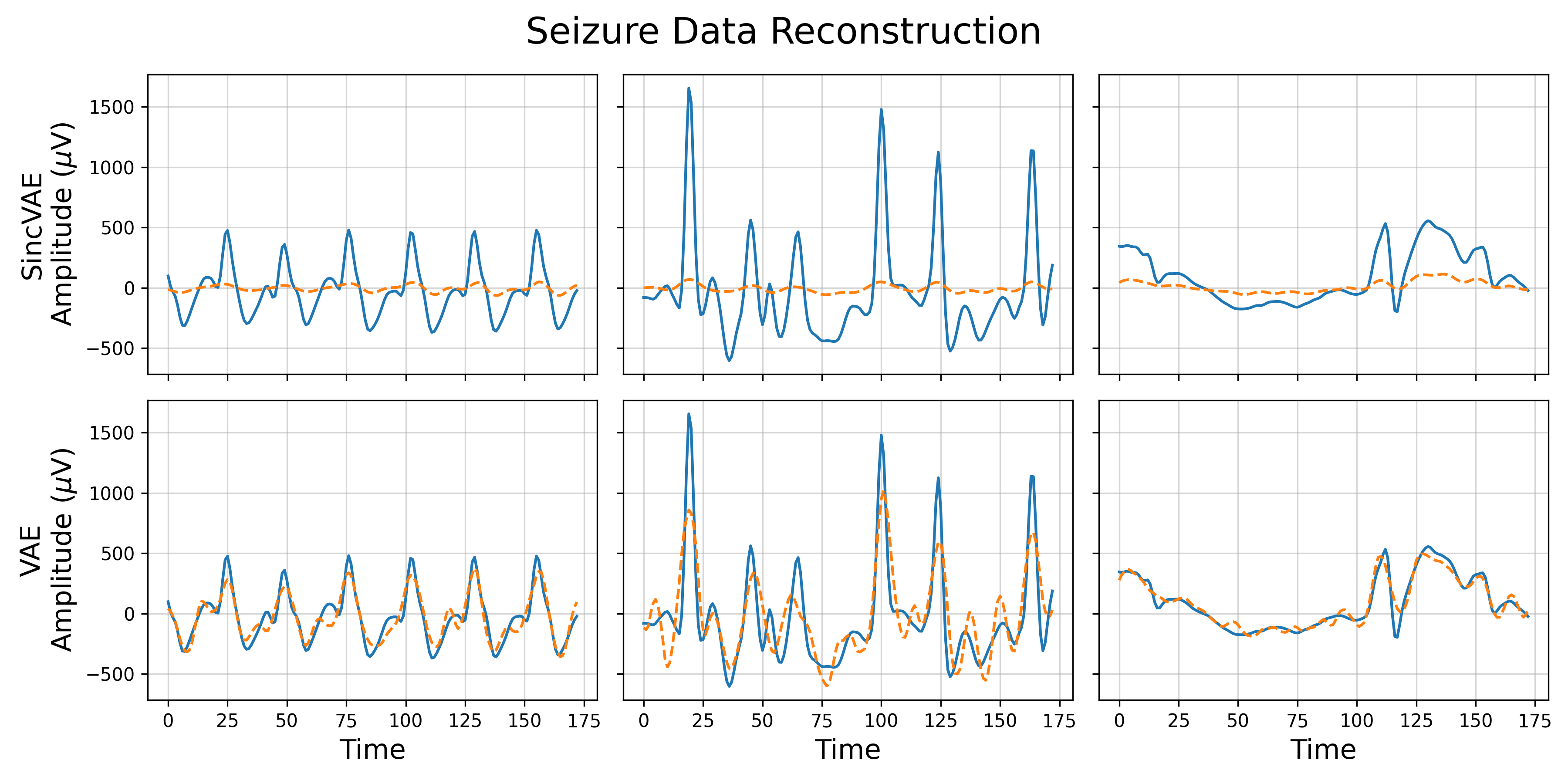}
    }
    \caption{Graphical representation of three randomly selected seizure samples drawn from Set E, reconstructed using SincVAE (first row) and VAE (second row). In each plot, the original input is represented by a solid blue line, while its reconstruction is represented by a dashed orange line.}
    \label{fig:bonn_seizurereconstruction}
\end{figure}

Figure \ref{fig:bonn_seizurereconstruction} shows instead the EEG signal reconstructions performed by both the VAE and SincVAE models on seizure samples from Set E. Notably, the VAE model exhibits superior reconstruction fidelity for these seizure signals compared to SincVAE, which results in a higher number of false negatives. Consistent with the classification results, this difference in reconstruction quality suggests varying performance between the two models in processing seizure-related EEG data. The SincVAE's difficulty in reconstructing seizure signals suggests its ability to distinguish between normal and seizure states, thereby establishing it as a more reliable solution for seizure detection.

\begin{figure}[ht!]
    \centering
    \begin{minipage}{0.52\textwidth}
        \centering
        \includegraphics[width=\textwidth]{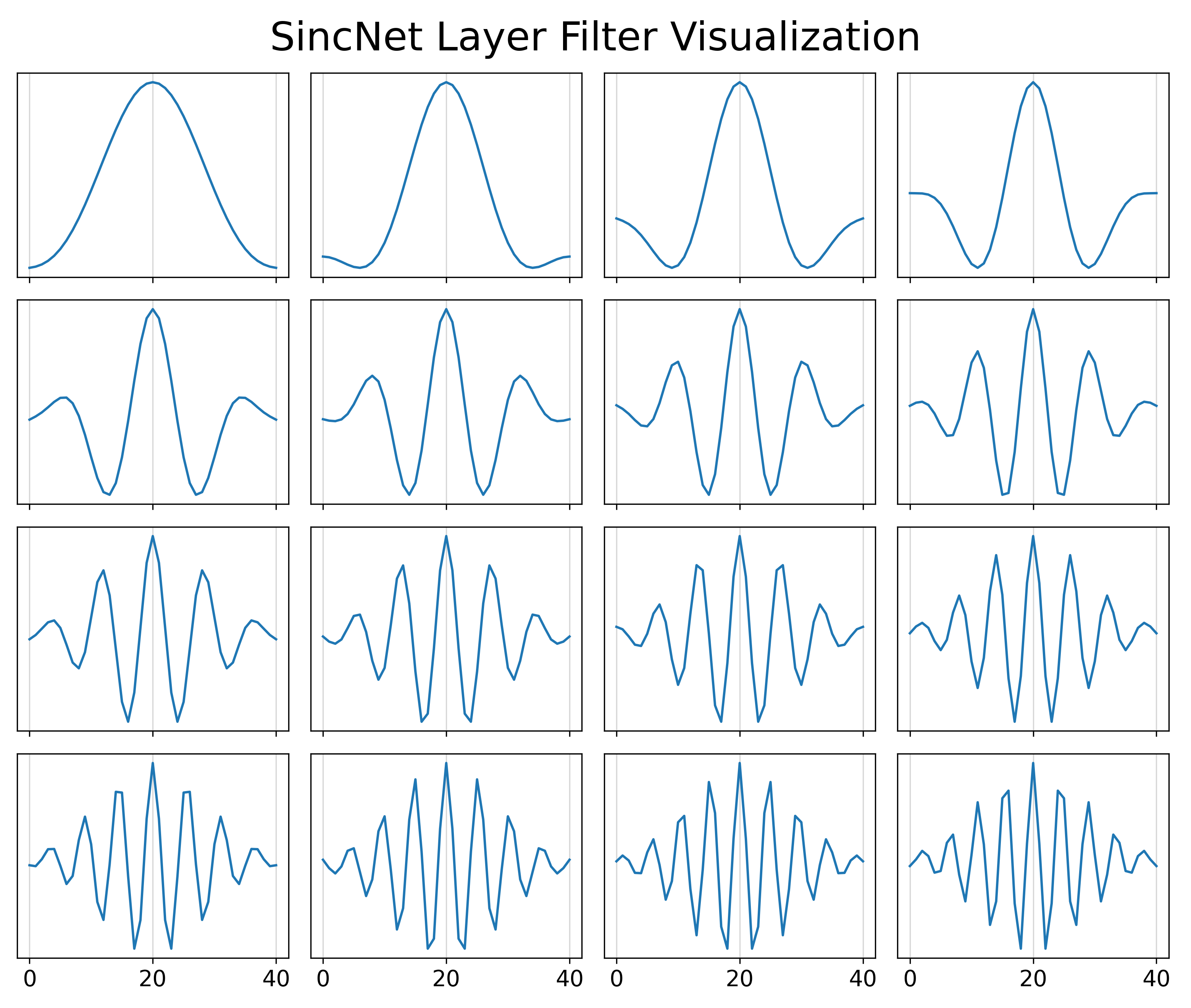}
        \caption{\textcolor{black}{Graphical representation of the filters learned by SincVAE in the SincNet layer in the time domain.}}
        \label{fig:sincvae_filters}
    \end{minipage}\hfill
    \begin{minipage}{0.45\textwidth}
        \centering
        \begin{table}[H] 
            \centering
            \scalebox{.62}{
                \begin{tabular}{ccc}
                \toprule
                LF (Hz) & HF (Hz) & Bandwidth (Hz) \\ 
                \midrule
                0.00 & 3.11 & 3.11 \\
                2.80 & 5.69 & 2.89 \\
                5.08 & 7.71 & 2.62 \\
                7.31 & 9.65 & 2.34 \\
                9.81 & 12.21 & 2.40 \\
                12.44 & 14.94 & 2.50 \\
                14.96 & 17.48 & 2.52 \\
                17.40 & 19.83 & 2.43 \\
                19.68 & 21.94 & 2.25 \\
                22.36 & 24.83 & 2.47 \\
                24.85 & 27.25 & 2.40 \\
                27.38 & 29.83 & 2.46 \\
                29.99 & 32.51 & 2.52 \\
                32.30 & 34.68 & 2.38 \\
                35.18 & 37.96 & 2.78 \\
                37.51 & 40.10 & 2.59 \\
                \bottomrule
                \end{tabular}
            }
            \caption{\textcolor{black}{Details of the 16 filters learned by the SincNet layer in the SincVAE architecture, including low-frequency (LF), high-frequency (HF), and bandwidth for each filter.}}
            \label{table:filter_details}
        \end{table}
    \end{minipage}
\end{figure}

\begin{figure}[ht!]
    \centering
    \scalebox{.9}{
        \includegraphics[width=\textwidth]{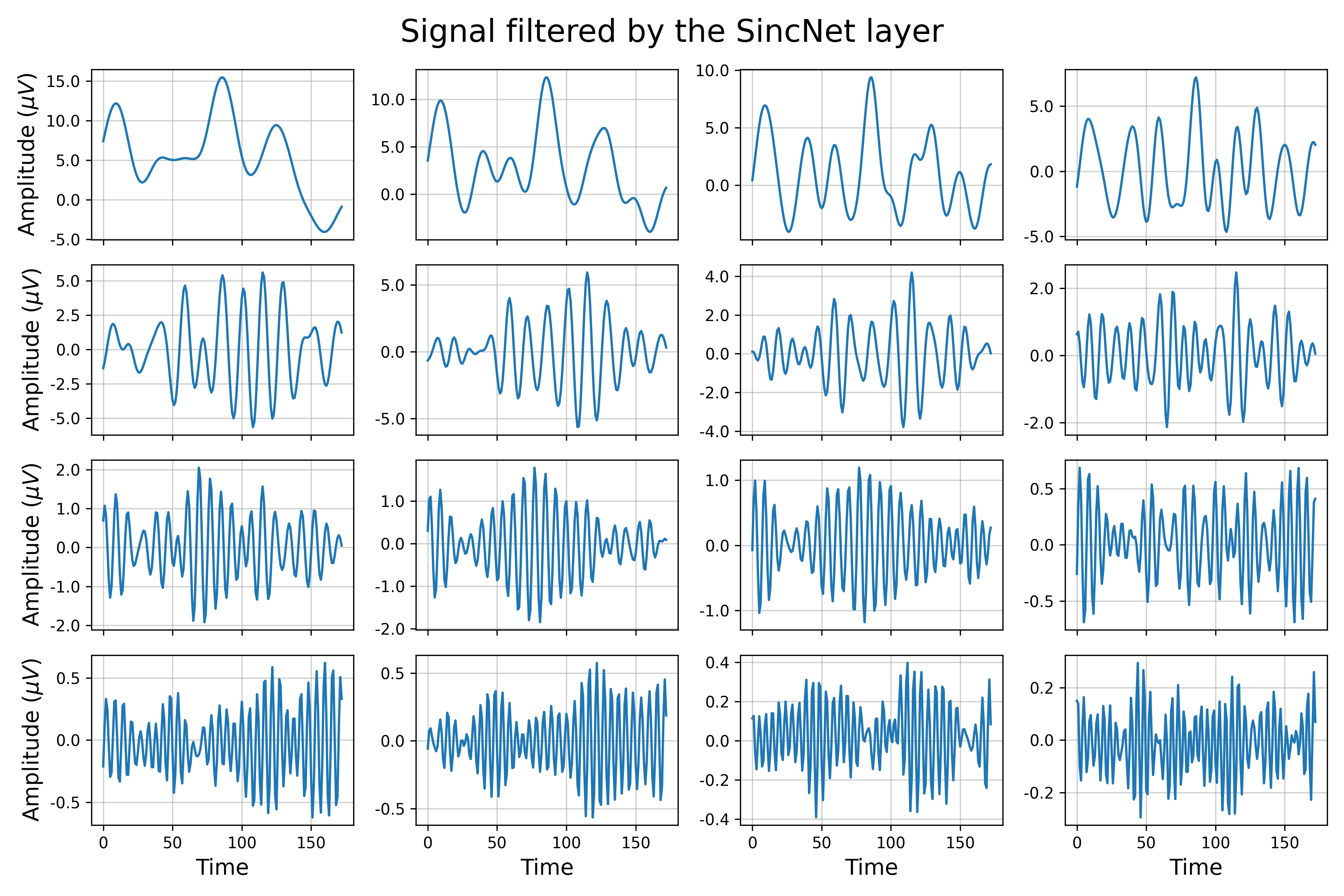}
    }
    \caption{\textcolor{black}{Graphical representation of the decomposition of the leftmost EEG recording shown in Figure \ref{fig:bonn_nonseizurereconstruction}. The signal decomposition was achieved using the SincNet layer after the training stage, which processed the recording with the learned filters.}}
    \label{fig:sincvae_decompos}
\end{figure}

\textcolor{black}{
Finally, Figure \ref{fig:sincvae_filters} shows the 16 filters learned by the SincNet layer in the SincVAE architecture after the training stage, in the time domain. These filters are visualized only for the network trained on Set A, as the network trained on Set B shows similar characteristics. Details regarding the learned filters, including the low-frequency, high-frequency, and bandwidth, are reported in Table \ref{table:filter_details}. Figure \ref{fig:sincvae_decompos} shows the leftmost non-seizure signal shown in Figure \ref{fig:bonn_nonseizurereconstruction} decomposed by the SincNet learned filters as an example of the filtering process.
}


\textcolor{black}{
\subsubsection{Comparison with Existing State-of-the-Art Anomaly Detection Methods}
To further assess the effectiveness of our proposed approach, we provide a comparison with several existing state-of-the-art anomaly detection methods. 
It is important to remark that our goal is not to establish new state-of-the-art results over the dataset under investigation. Instead, we are focused on offering a meaningful comparison with other widely recognized models, giving the reader a clearer idea of how our method performs in relation to existing techniques in the field.
Thus, we have chosen to evaluate the performance of several widely used anomaly detection methods to provide a robust comparison of our approach. Specifically, we include one-class support vector machine (OC-SVM) \cite{pollastro2023semi}, isolation forest (IF) \cite{liu2008isolation}, and local outlier factor (LOF) \cite{alghushairy2020review}. Additionally, we used PCA \cite{hastie2009elements}, where the classification was performed by thresholding on the MSE over the reconstruction, consistent with the approach used for SincVAE in our analysis. In particular, only the threshold $t_2$ was used in this case.
Each method was optimized through a model selection stage, following the same details applied on the SincVAE optimization. We employed random search \cite{goodfellow2016deep} to explore various hyperparameter configurations for each algorithm, conducting 15 trials per method to ensure robust evaluation. The results corresponding to the best model obtained from each algorithm are reported in Table \ref{tab:comparison_existing_methods}.
\begin{table}[ht!]
\centering
\caption{\textcolor{black}{Classification results for the comparison with the existing state-of-the-art anomaly detection methods. The table presents the performance for both experimental cases: Set A vs Set E and Set B vs Set E. For each case and metric, the model achieving the best results are highlighted in bold.}}
    \scalebox{.9}{
        \begin{tabular}{c||ccc|ccc}
            \hline
            \multirow{2}{*}{Method} & \multicolumn{3}{c}{Set A vs Set E} & \multicolumn{3}{c}{Set B vs Set E}\\
            \cline{2-7}
            & F1 & Precision & Recall & F1 & Precision & Recall \\
            \hline
            SincVAE & \textbf{99.4} & 98.9 & \textbf{100} & \textbf{98.9} & \textbf{98.5} & 99.4 \\
            VAE & 99.2 & \textbf{99.2} & 99.2 & 98.0 & 98.0 & 97.9 \\
            OC-SVM & 98.5 & 97.0 & \textbf{100} & 98.4 & 97.2 & 99.6 \\
            IF & 99.3 & \textbf{99.2} & 99.5 & 95.7 & 98.1 & 93.6 \\
            LOF & 99.2 & 98.5 & \textbf{100} & 98.7 & 97.6 & \textbf{99.8} \\
            PCA & 96.9 & \textbf{99.2} & 94.7 & 94.6 & 98.0 & 91.4 \\
            \hline
        \end{tabular}
    }
\label{tab:comparison_existing_methods}
\end{table}
The results show that SincVAE achieves the highest F1 score in both cases, i.e. \SI{99.4}{\percent} for Set A vs Set E and \SI{98.9}{\percent} for Set B vs Set E. This highlights the robustness of our approach in detecting anomalies while maintaining an optimal balance between Precision and Recall. Additionally, in terms of Recall, SincVAE achieves the best Recall on the Set A vs Set E case, that is \SI{100}{\percent}, while on the Set B vs Set E case its Recall is \SI{-0.4}{\percent} from the best, achieved by LOF; in terms of Precision, SincVAE reaches a value of \SI{-0.3}{\percent} from the best, achieved by VAE, IF and PCA, while on the Set B vs Set E it shows the best value, that is \SI{98.5}{\percent}.
Given the consistent superior performance of SincVAE, all subsequent experiments presented in this manuscript will be centered solely on the comparison between SincVAE and VAE. This focus allows for a more in-depth evaluation of the advantages offered by SincVAE in seizure detection, while maintaining a clear and direct comparison keeping only VAE results as baseline.
}

\subsection{The CHB-MIT Dataset}
The analyses conducted for the CHB-MIT dataset followed the binary semi-supervised classification approach used for the Bonn dataset. Specifically, the training dataset included only non-seizure data, and to evaluate the seizure class detection, only windows that occurred during the ictal periods, as denoted by the dataset's authors, were included in the test. Thus, preictal and postictal periods were excluded from the analyses.
\begin{table}[ht!]
  \centering
  \caption{The F1, precision, recall and ROC-AUC results for detecting seizure and non-seizure EEG recordings on the CHB-MIT dataset for each subject.}
  \scalebox{.8}{
    \begin{tabular}{c||cc|cc|cc||cc}
        \hline
        \multirow{2}{*}{Subject} & \multicolumn{2}{c|}{F1 (\%)} & \multicolumn{2}{c|}{Precision (\%)} & \multicolumn{2}{c||}{Recall (\%)} & \multicolumn{2}{c}{ROC-AUC (\%)} \\
        \cline{2-9}
         & SincVAE & VAE & SincVAE & VAE & SincVAE & VAE & SincVAE & VAE \\
        \hline
        1 & \textbf{91.9} & 91.8 & 90.0 & \textbf{90.9} & \textbf{93.9} & 92.8 & \textbf{97.5} & 97.2 \\
        2 & \textbf{75.6} & 70.8 & \textbf{65.1} & 62.7 & \textbf{90.1} & 81.3 & \textbf{97.2} & 95.8 \\
        3 & \textbf{95.0} & 94.3 & \textbf{92.7} & 92.2 & \textbf{97.5} & 96.5 & \textbf{98.5} & 97.9 \\
        4 & \textbf{71.8} & 64.5 & \textbf{64.4} & 61.7 & \textbf{81.2} & 67.5 & \textbf{91.7} & 88.9 \\
        5 & \textbf{95.6} & 93.4 & \textbf{94.6} & 94.2 & \textbf{96.6} & 92.7 & \textbf{99.2} & 98.4 \\
        6 & \textbf{10.8} & 6.9 & \textbf{20.8} & 16.7 & \textbf{7.2} & 4.3 & \textbf{79.7} & 79.3 \\
        7 & \textbf{96.2} & 93.1 & \textbf{98.1} & 98.0 & \textbf{94.5} & 88.6 & \textbf{99.7} & 99.6 \\
        8 & 76.8 & \textbf{79.9} & 80.8 & \textbf{84.7} & 73.1 & \textbf{75.6} & 80.4 & \textbf{84.5} \\
        9 & 84.9 & \textbf{92.3} & 74.7 & \textbf{89.6} & \textbf{98.4} & 95.2 & \textbf{99.1} & 97.6 \\
        10 & \textbf{95.9} & 87.8 & \textbf{99.8} & 99.4 & \textbf{92.4} & 78.5 & \textbf{99.8} & 99.5 \\
        11 & \textbf{97.5} & 96.8 & 98.6 & \textbf{98.7} & \textbf{96.4} & 94.9 & \textbf{99.3} & 99.1 \\
        12 & \textbf{17.5} & 16.1 & \textbf{43.8} & 30.4 & \textbf{10.9} & 10.9 & \textbf{76.1} & 74.5 \\
        13 & 15.7 & \textbf{16.5} & 70.4 & \textbf{71.4} & 8.8 & \textbf{9.3} & 66.9 & \textbf{72.0} \\
        14 & 30.8 & 30.8 & 50.9 & 50.9 & 22.0 & 22.0 & \textbf{55.9} & 55.3 \\
        15 & \textbf{66.0} & 60.6 & \textbf{92.6} & 92.2 & \textbf{51.3} & 45.2 & 88.0 & \textbf{88.8} \\
        16 & \textbf{5.5} & 5.3 & \textbf{4.2} & 4.0 & 7.9 & 7.9 & 69.5 & \textbf{70.7} \\
        17 & \textbf{20.7} & 9.6 & \textbf{94.4} & 83.3 & \textbf{11.6} & 5.1 & \textbf{54.1} & 53.9 \\
        18 & 85.3 & \textbf{85.4} & 94.6 & \textbf{95.0} & 77.6 & 77.6 & \textbf{95.4} & 95.3 \\
        19 & 85.9 & \textbf{88.1} & 77.9 & \textbf{82.8} & \textbf{95.8} & 94.1 & \textbf{98.4} & 98.2 \\
        20 & \textbf{69.8} & 67.0 & 87.2 & \textbf{88.4} & \textbf{58.2} & 53.9 & 83.4 & \textbf{83.6} \\
        21 & \textbf{26.6} & 9.1 & \textbf{54.7} & 50.0 & \textbf{17.6} & 5.0 & \textbf{77.7} & 70.9 \\
        22 & \textbf{92.7} & 90.2 & \textbf{95.3} & 95.1 & \textbf{90.2} & 85.8 & \textbf{99.2} & 97.9 \\
        23 & 81.7 & \textbf{85.4} & 89.5 & \textbf{94.6} & 75.2 & \textbf{77.9} & 98.6 & \textbf{99.0} \\
        24 & 67.0 & \textbf{68.3} & 70.8 & \textbf{75.2} & \textbf{63.5} & 62.6 & \textbf{69.8} & 69.6 \\
        \hline
    \end{tabular}
    }
  \label{table:chbmit_results_binary}
\end{table}
Based on the findings from the experiments with the Bonn dataset, the analyses on the CHB-MIT dataset will be presented in terms of F1, Precision, Recall and ROC-AUC performances of both the SincVAE and VAE models for each subject in the CHB-MIT dataset. 
The decision threshold was chosen to be unique and equal to the 95th percentile of the validation set's MSE for each subject. 
Results are shown in Table \ref{table:chbmit_results_binary}.

Observing the F1 scores, SincVAE performs comparably or slightly better than VAE for the majority of the subjects in the CHB-MIT dataset. Specifically, for subjects 1, 2, 3, 4, 5, 6, 7, 10, 11, 12, 15, 16, 17, 20, 21, and 22, SincVAE either matches or exceeds the performance of VAE.
Notably, there are cases where VAE has a higher F1 score than SincVAE, specifically for subjects 8, 9, 13, 18, 19, 23, and 24. These variations suggest that VAE might be better suited for certain cases.
\textcolor{black}{It is also important to note that, according to the ROC-AUC results, SincVAE outperforms VAE in 18 out of 24 cases, further validating its effectiveness over VAE.}

It is worth noting that the F1 scores for some subjects, such as 6 and 13, are low, which may be influenced by factors such as the inherent quality of the EEG recordings, the need for specific data preprocessing or the need for more sophisticated model architectures. However, the main aim of this study is to explore the effectiveness of integrating a SincNet layer within a VAE framework, rather than establishing new state-of-the-art results on the CHB-MIT dataset. Thus, discussions on specific architectural enhancements to improve EEG data quality are outside the scope of this work, which focuses instead on assessing the added value of the SincNet layer within the VAE architecture.

\textcolor{black}{
Finally, to assess whether SincVAE outperforms VAE from a statistical significance perspective, hypothesis testing was conducted, first considering F1 scores and then ROC-AUC values, with the significance level set at $\alpha = 0.05$.
For the F1 scores, the null hypothesis stated that the mean difference between the performance of SincVAE and VAE was zero, while the alternative hypothesis suggested that SincVAE outperforms VAE, i.e., the mean difference is greater than zero. 
To check for normality in the differences, the Shapiro-Wilk test was first applied. 
The p-value obtained was $p = 0.101$, indicating that the differences followed a normal distribution. Consequently, a one-sided paired t-test \cite{moore2009introduction} was performed to test whether the mean difference between SincVAE and VAE was greater than zero. The resulting p-value, $p = 0.023$, was below the significance level $\alpha$, leading to the rejection of the null hypothesis and the conclusion that SincVAE significantly outperforms VAE.
For the ROC-AUC values, a similar approach was followed. The Shapiro-Wilk test yielded a p-value of $p < 0.001$, indicating that the differences did not follow a normal distribution. Therefore, the one-sided Wilcoxon signed-rank test \cite{moore2009introduction} was applied. The p-value from the Wilcoxon test was $p = 0.042$, which is below the significance level $\alpha$. As a result, the null hypothesis was rejected, and it was concluded that SincVAE significantly outperforms VAE in terms of ROC-AUC values.
}

\begin{figure}[ht!]
    \centering
    \scalebox{.8}{
        \includegraphics[width=\textwidth]{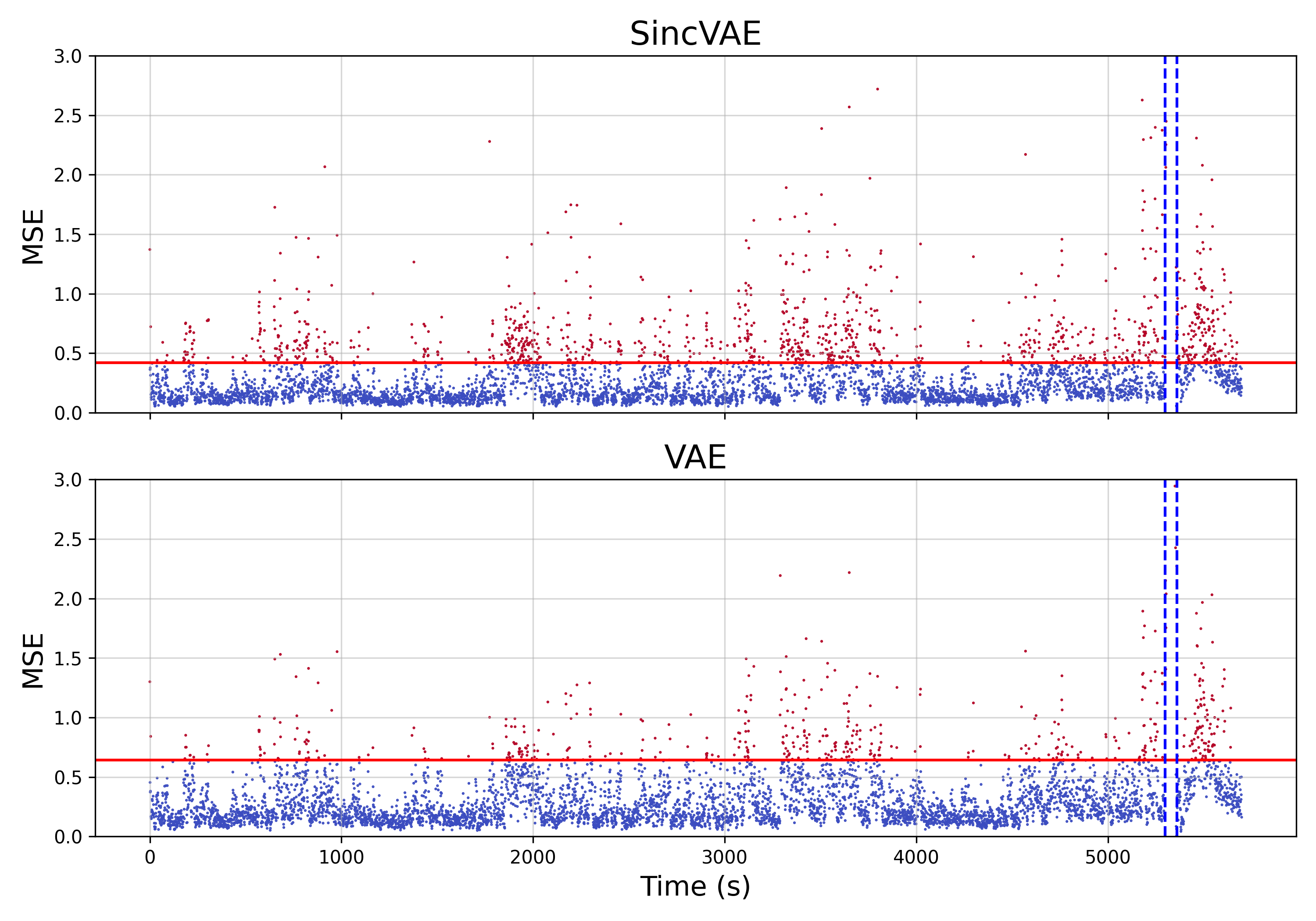}
    }
    \caption[MSE values obtained using SincVAE and VAE for each second on the track number 19 of the subject 9]{Graphical representation of MSE values obtained using SincVAE and VAE for each second on the track number 19 of the subject 9. The two vertical dashed blue lines indicate, from left to right, the ictal phase annotated by the dataset authors. In both of the plots, the horizontal red line indicates the decision threshold used to classify the EEG recordings. The horizontal red line represents the decision threshold, set at the 95th percentile of the MSE values from the validation sets. Blue dots indicate EEG recordings classified as non-seizure, while the red dots indicate EEG recordings classified as seizure.}
    \label{fig:bonn_mse_subj_19}
\end{figure}

As stated above, the seizure tracks in the CHB-MIT dataset include preictal, ictal, and postictal phases. Figure \ref{fig:bonn_mse_subj_19} shows the MSE values obtained for each second of track number 19 from subject 9, a track identified by the dataset's authors as including a seizure event from seconds 5299 to 5361. These specific seconds are visually demarcated with two dashed vertical blue lines. The horizontal red line on the plot indicates the decision threshold, set at the 95th percentile of the MSE values obtained from the validation set. Blue dots on the plot represent EEG recordings that have been classified as non-seizure, while red dots indicate those classified as seizure.

\begin{figure}[ht!]
    \centering
    \scalebox{.7}{
        \includegraphics[width=\textwidth]{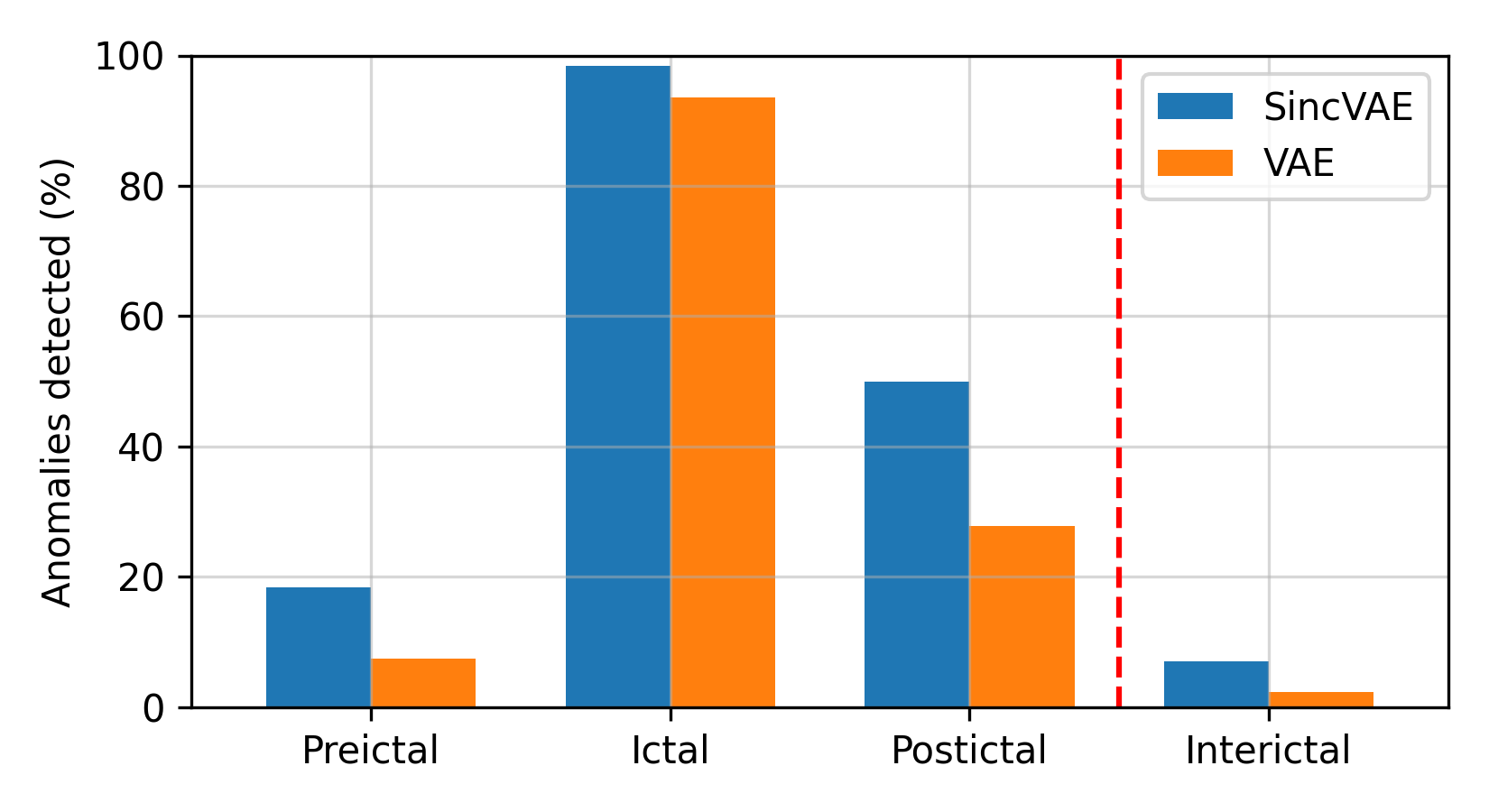}
    }
    \caption[Detection rates of anomalies detected by SincVAE and VAE on the track number 19 of the subject 9]{Graphical representation of the detection rates of anomalies by SincVAE and VAE on the track number 19 of the subject 9. The detection is shown across preictal, ictal, postictal, and interictal phases.}
    \label{fig:bonn_subj9_19_barplot}
\end{figure}

\begin{table}[ht!]
\centering
\caption{Average percentage of EEG readings identified as anomalous by SincVAE and VAE during the preictal and postictal phases across tracks containing seizures for each subject.}
\scalebox{.8}{
   \begin{tabular}{c||cc|cc}
   \hline
   \multirow{2}{*}{Subject} & \multicolumn{2}{c}{Preictal (\%)} & \multicolumn{2}{c}{Postictal (\%)} \\ 
   \cline{2-5}
    & SincVAE & VAE & SincVAE & VAE \\ 
    \hline
   1 & $1.91 \pm 1.48$ & $\mathbf{1.94 \pm 1.87}$ & $\mathbf{2.27 \pm 2.67}$ & $1.55 \pm 0.64$ \\
   2 & $\mathbf{6.39 \pm 0.23}$ & $5.1 \pm 1.25$ & $35.48 \pm 8.41$ & $\mathbf{35.99 \pm 9.71}$ \\
   3 & $3.41 \pm 3.77$ & $\mathbf{3.5 \pm 3.82}$ & $\mathbf{9.11 \pm 5.76}$ & $8.78 \pm 5.41$ \\
   4 & $54.07 \pm 40.03$ & $\mathbf{54.16 \pm 39.96}$ & $\mathbf{6.21 \pm 0.98}$ & $6.07 \pm 0.72$ \\
   5 & $5.14 \pm 4.64$ & $\mathbf{5.33 \pm 4.42}$ & $\mathbf{4.63 \pm 1.7}$ & $4.06 \pm 1.28$ \\
   6 & $\mathbf{3.89 \pm 4.38}$ & $3.38 \pm 4.35$ & $\mathbf{14.79 \pm 21.39}$ & $13.32 \pm 21.96$ \\
   7 & $\mathbf{4.34 \pm 1.16}$ & $3.36 \pm 1.11$ & $\mathbf{15.55 \pm 11.59}$ & $13.88 \pm 10.45$ \\
   8 & $\mathbf{21.62 \pm 9.76}$ & $17.71 \pm 9.01$ & $19.2 \pm 24.17$ & $\mathbf{19.94 \pm 24.28}$ \\
   9 & $\mathbf{18.18 \pm 0.14}$ & $9.96 \pm 2.46$ & $\mathbf{27.78 \pm 22.22}$ & $14.76 \pm 13.05$ \\
   10 & $\mathbf{10.82 \pm 14.71}$ & $9.72 \pm 13.76$ & $\mathbf{10.01 \pm 16.37}$ & $9.62 \pm 15.75$ \\
   11 & $\mathbf{21.09 \pm 10.45}$ & $18.96 \pm 9.55$ & $\mathbf{8.88 \pm 9.13}$ & $8.82 \pm 9.17$ \\
   12 & $\mathbf{2.38 \pm 1.77}$ & $2.21 \pm 1.75$ & $\mathbf{5.85 \pm 0.27}$ & $5.57 \pm 1.97$ \\
   13 & $\mathbf{2.09 \pm 1.65}$ & $1.95 \pm 1.82$ & $2.19 \pm 2.24$ & $\mathbf{2.53 \pm 2.5}$ \\
   14 & $\mathbf{6.14 \pm 2.81}$ & $6.25 \pm 3.19$ & $\mathbf{5.76 \pm 2.39}$ & $6.34 \pm 2.95$ \\
   15 & $\mathbf{14.14 \pm 19.81}$ & $12.7 \pm 17.86$ & $\mathbf{6.26 \pm 7.78}$ & $6.11 \pm 7.52$ \\
   16 & $7.25 \pm 4.53$ & $\mathbf{7.78 \pm 4.79}$ & $7.2 \pm 4.57$ & $\mathbf{7.53 \pm 4.81}$ \\
   17 & $\mathbf{2.72 \pm 1.41}$ & $2.47 \pm 1.45$ & $\mathbf{0.76 \pm 0.72}$ & $0.43 \pm 0.41$ \\
   18 & $\mathbf{6.61 \pm 5.14}$ & $6.36 \pm 5.16$ & $\mathbf{25.17 \pm 16.73}$ & $23.6 \pm 16.52$ \\
   19 & $\mathbf{6.58 \pm 2.77}$ & $4.01 \pm 2.37$ & $\mathbf{70.67 \pm 21.8}$ & $63.13 \pm 23.1$ \\
   20 & $\mathbf{0.97 \pm 1.37}$ & $0.77 \pm 1.07$ & $\mathbf{0.74 \pm 0.48}$ & $0.68 \pm 0.48$ \\
   21 & $6.44 \pm 2.9$ & $\mathbf{7.14 \pm 3.27}$ & $6.14 \pm 5.26$ & $\mathbf{6.77 \pm 5.5}$ \\
   22 & $4.63 \pm 0.48$ & $\mathbf{4.77 \pm 0.46}$ & $\mathbf{19.41 \pm 16.04}$ & $18.63 \pm 14.98$ \\
   23 & $2.95 \pm 0.0$ & $\mathbf{2.98 \pm 0.0}$ & $\mathbf{2.64 \pm 0.0}$ & $2.29 \pm 0.0$ \\
   24 & $\mathbf{3.75 \pm 8.83}$ & $3.35 \pm 7.71$ & $\mathbf{4.24 \pm 10.44}$ & $3.84 \pm 9.46$ \\
   \hline
   \end{tabular}
}
\label{table:chbmit_results_preictal}
\end{table}

It can be noted that several recordings are classified as anomalous by both of the models either in the preictal or in the postictal phase.
Figure \ref{fig:bonn_subj9_19_barplot} shows the percentage of anomalous points detected by SincVAE and VAE in the preictal, ictal and postictal phases.
Both models detect seizures during the ictal phase with a high rate, and demonstrate effectiveness in recognizing the interictal phase with a low rate of false positives. It is interesting to notice that both of the models are able to recognize anomalies during the preictal and postictal phases.
Specifically, SincVAE identifies \SI{18.04}{\percent} of anomalies in the preictal phase compared to \SI{7.49}{\percent} by VAE, and it detects \SI{50}{\percent} of anomalies in the postictal phase, against the \SI{27.81}{\percent} by VAE. This enhanced capability to detect anomalies during the preictal and postictal phases suggests that SincVAE could be particularly advantageous for applications that require an early warning system for seizures and consistent monitoring during the recovery phase following a seizure.
Table \ref{table:chbmit_results_preictal} extends the analysis shown in Figure \ref{fig:bonn_subj9_19_barplot} covering all subjects but specifically focusing on the preictal and postictal phases. 
The anomalies detected during these phases are averaged across the seizure tracks for each subject, as their numerosity varies. 
The findings from Figure \ref{fig:bonn_subj9_19_barplot} are consistent across the majority of subjects, indicating that SincVAE detects a higher rate of anomalies during the preictal and postictal phases.

\section{Conclusions}
\label{sec:conclusions}
This work proposed the SincVAE architecture, integrating SincNet within the VAE framework to perform semi-supervised anomaly detection on time series data. In particular, the framework was explored through the seizure detection problem on EEG data.
From the experimental session, SincVAE has demonstrated considerable potential in improving the reliability and accuracy of seizure detection in EEG data compared to a standard VAE. 

This method not only simplifies the preprocessing steps by effectively utilizing the bandpass filters of SincNet but also enhances the overall detection process by focusing on significant EEG frequency bands detected by the neural network during the training stage. The experiments conducted on various datasets validated the effectiveness of SincVAE, showcasing its superiority in various scenarios and making it a valuable approach for real-world applications in seizure detection.

Also, the capability of SincVAE to discern subtle anomalies in EEG data indicates SincVAE as a tool to detect early signs of epilepsy in the preictal stage and to monitor the patients' status during the postictal stage. This aspect can profoundly affect patient monitoring.

Furthermore, the semi-supervised nature of SincVAE, requiring only non-seizure data for training, addresses challenges associated with the scarcity of labeled anomaly data in medical datasets, making it an efficient solution in real-world applications where anomalies are rare but critical to detect. \textcolor{black}{However, it is important to note that, like any DL-based method, SincVAE is data-driven and its performance heavily depends on the availability of extensive, high-quality data. A comprehensive dataset that adequately represents the patient's normal state is essential for ensuring the model generalizes effectively, enabling it to reliably differentiate between normal and anomalous conditions.}

Future work could explore the application of SincVAE in other types of time-series anomaly detection tasks, potentially broadening its utility in healthcare. 
Another promising direction involves integrating attention mechanisms \cite{niu2021review} within the SincVAE framework, in particular on the SincNet layer, to refine the selection of frequency bands identified by the SincNet layer. \textcolor{black}{Such enhancement could enable more accurate filtering by prioritizing certain frequency bands more than others, potentially leading to an improvement of the model's overall performance.}

Moreover, a key direction for future work will be enhancing the interpretability of the SincNet layer to ensure greater transparency and provide deeper insights, particularly within clinical contexts. 
\textcolor{black}{As illustrated in Figures \ref{fig:sincvae_filters} and \ref{fig:sincvae_decompos}, the learned filters and corresponding filtered signals become observable and interpretable after training.}
Attention coefficients may offer valuable insights into the relative importance of different filters, thereby providing a clearer understanding of which frequency bands are most influential for the task at hand. Also, we intend to explore the use of explainable artificial intelligence (XAI) \cite{minh2022explainable,apicella2023strategies} techniques in this direction. These methods could provide additional interpretative power, allowing for a deeper analysis of how the model processes and prioritizes different frequency bands. Finally, we also plan to conduct ablation studies \cite{li2024optimal,teng2022survey} to systematically evaluate the contribution of specific frequency bands to the model's performance. These studies will be crucial in identifying the most reliable direction to improve interpretability of the filtering stage of the SincNet layer, in order to let the analysts understand which frequency bands have the most significant impact on the solution with respect to the input data, thereby assisting in interpreting the outcomes from a clinical perspective.

\section*{Acknowledgement}
This work was supported by the Italian Ministry of University and Research, PNRR project PE00000013 Future Artificial Intelligence Research (FAIR), E63C22002150007 and by the Italian Ministry of University and Research, PRIN research project ``BRIO -- BIAS, RISK, OPACITY in AI: design, verification and development of Trustworthy AI.'', Project no. 2020SSKZ7R. The authors express their gratitude to Dr. Angela Natalizio and Dr. Antonio Esposito for their support in enhancing the understanding of the biomedical aspects of the analyzed signals.


\begin{thebibliography}{100}

\bibitem{sagiroglu2013big}
Seref Sagiroglu and Duygu Sinanc.
\newblock Big data: A review.
\newblock In {\em 2013 international conference on collaboration technologies and systems (CTS)}, pages 42--47. IEEE, 2013.

\bibitem{aggarwal2017introduction}
Charu~C Aggarwal and Charu~C Aggarwal.
\newblock {\em An introduction to outlier analysis}.
\newblock Springer, 2017.

\bibitem{chalapathy2019deep}
Raghavendra Chalapathy and Sanjay Chawla.
\newblock Deep learning for anomaly detection: A survey.
\newblock {\em arXiv preprint arXiv:1901.03407}, 2019.

\bibitem{aleskerov1997cardwatch}
Emin Aleskerov, Bernd Freisleben, and Bharat Rao.
\newblock Cardwatch: A neural network based database mining system for credit card fraud detection.
\newblock In {\em Proceedings of the IEEE/IAFE 1997 computational intelligence for financial engineering (CIFEr)}, pages 220--226. IEEE, 1997.

\bibitem{kumar2005parallel}
Vipin Kumar.
\newblock Parallel and distributed computing for cybersecurity.
\newblock {\em IEEE Distributed Systems Online}, 6(10), 2005.

\bibitem{pollastro2023semi}
Andrea Pollastro, Giusiana Testa, Antonio Bilotta, and Roberto Prevete.
\newblock Semi-supervised detection of structural damage using variational autoencoder and a one-class support vector machine.
\newblock {\em IEEE Access}, 2023.

\bibitem{de2024dynamic}
Alessandra De~Angelis, Antonio Bilotta, Maria~Rosaria Pecce, Andrea Pollastro, and Roberto Prevete.
\newblock Dynamic identification methods and artificial intelligence algorithms for damage detection of masonry infills.
\newblock {\em Journal of Civil Structural Health Monitoring}, pages 1--20, 2024.

\bibitem{schlegl2017unsupervised}
Thomas Schlegl, Philipp Seeb{\"o}ck, Sebastian~M Waldstein, Ursula Schmidt-Erfurth, and Georg Langs.
\newblock Unsupervised anomaly detection with generative adversarial networks to guide marker discovery.
\newblock In {\em International conference on information processing in medical imaging}, pages 146--157. Springer, 2017.

\bibitem{salem2014online}
Osman Salem, Yaning Liu, Ahmed Mehaoua, and Raouf Boutaba.
\newblock Online anomaly detection in wireless body area networks for reliable healthcare monitoring.
\newblock {\em IEEE journal of biomedical and health informatics}, 18(5):1541--1551, 2014.

\bibitem{kavitha2021machine}
M~Kavitha, PVVS Srinivas, PS~Latha Kalyampudi, Singaraju Srinivasulu, et~al.
\newblock Machine learning techniques for anomaly detection in smart healthcare.
\newblock In {\em 2021 Third International Conference on Inventive Research in Computing Applications (ICIRCA)}, pages 1350--1356. IEEE, 2021.

\bibitem{naidoo2020unsupervised}
Krishnan Naidoo and Vukosi Marivate.
\newblock Unsupervised anomaly detection of healthcare providers using generative adversarial networks.
\newblock In {\em Responsible Design, Implementation and Use of Information and Communication Technology: 19th IFIP WG 6.11 Conference on e-Business, e-Services, and e-Society, I3E 2020, Skukuza, South Africa, April 6--8, 2020, Proceedings, Part I 19}, pages 419--430. Springer, 2020.

\bibitem{you2022semi}
Sungmin You, Baek~Hwan Cho, Young-Min Shon, Dae-Won Seo, and In~Young Kim.
\newblock Semi-supervised automatic seizure detection using personalized anomaly detecting variational autoencoder with behind-the-ear eeg.
\newblock {\em Computer Methods and Programs in Biomedicine}, 213:106542, 2022.

\bibitem{vsabic2021healthcare}
Edin {\v{S}}abi{\'c}, David Keeley, Bailey Henderson, and Sara Nannemann.
\newblock Healthcare and anomaly detection: using machine learning to predict anomalies in heart rate data.
\newblock {\em AI \& SOCIETY}, 36(1):149--158, 2021.

\bibitem{ukil2016iot}
Arijit Ukil, Soma Bandyoapdhyay, Chetanya Puri, and Arpan Pal.
\newblock Iot healthcare analytics: The importance of anomaly detection.
\newblock In {\em 2016 IEEE 30th international conference on advanced information networking and applications (AINA)}, pages 994--997. IEEE, 2016.

\bibitem{shih2012brain}
Jerry~J Shih, Dean~J Krusienski, and Jonathan~R Wolpaw.
\newblock Brain-computer interfaces in medicine.
\newblock In {\em Mayo clinic proceedings}, volume~87, pages 268--279. Elsevier, 2012.

\bibitem{angrisani2022instrumentation}
L~Angrisani, A~Apicella, P~Arpaia, A~Cataldo, AD~Calce, A~Fullin, L~Gargiulo, L~Maffei, N~Moccaldi, A~Pollastro, et~al.
\newblock Instrumentation for eeg-based monitoring of the executive functions in a dual-task framework.
\newblock In {\em 25th IMEKO TC-4 International Symposium on Measurement of Electrical Quantities, IMEKO TC-4 2022 and 23rd International Workshop on ADC and DAC Modelling and Testing, IWADC 2022}, pages 161--165. International Measurement Confederation (IMEKO), 2022.

\bibitem{hosseini2017optimized}
Mohammad-Parsa Hosseini, Dario Pompili, Kost Elisevich, and Hamid Soltanian-Zadeh.
\newblock Optimized deep learning for eeg big data and seizure prediction bci via internet of things.
\newblock {\em IEEE Transactions on Big Data}, 3(4):392--404, 2017.

\bibitem{liang2010closed}
Sheng-Fu Liang, Fu-Zen Shaw, Chung-Ping Young, Da-Wei Chang, and Yi-Cheng Liao.
\newblock A closed-loop brain computer interface for real-time seizure detection and control.
\newblock In {\em 2010 Annual International Conference of the IEEE Engineering in Medicine and Biology}, pages 4950--4953. IEEE, 2010.

\bibitem{vaid2015eeg}
Swati Vaid, Preeti Singh, and Chamandeep Kaur.
\newblock Eeg signal analysis for bci interface: A review.
\newblock In {\em 2015 fifth international conference on advanced computing \& communication technologies}, pages 143--147. IEEE, 2015.

\bibitem{shahbazi2018generalizable}
Mohamad Shahbazi and Hamid Aghajan.
\newblock A generalizable model for seizure prediction based on deep learning using cnn-lstm architecture.
\newblock In {\em 2018 IEEE Global Conference on Signal and Information Processing (GlobalSIP)}, pages 469--473. IEEE, 2018.

\bibitem{litt2002prediction}
Brian Litt and Javier Echauz.
\newblock Prediction of epileptic seizures.
\newblock {\em The Lancet Neurology}, 1(1):22--30, 2002.

\bibitem{shoeb2009application}
Ali~Hossam Shoeb.
\newblock {\em Application of machine learning to epileptic seizure onset detection and treatment}.
\newblock PhD thesis, Massachusetts Institute of Technology, 2009.

\bibitem{shoeb2010application}
Ali~H Shoeb and John~V Guttag.
\newblock Application of machine learning to epileptic seizure detection.
\newblock In {\em Proceedings of the 27th international conference on machine learning (ICML-10)}, pages 975--982, 2010.

\bibitem{georgis2023supervised}
Zakary Georgis-Yap, Milos~R Popovic, and Shehroz~S Khan.
\newblock Supervised and unsupervised deep learning approaches for eeg seizure prediction.
\newblock {\em arXiv preprint arXiv:2304.14922}, 2023.

\bibitem{ahmed2016survey}
Mohiuddin Ahmed, Abdun~Naser Mahmood, and Jiankun Hu.
\newblock A survey of network anomaly detection techniques.
\newblock {\em Journal of Network and Computer Applications}, 60:19--31, 2016.

\bibitem{bhuyan2013network}
Monowar~H Bhuyan, Dhruba~Kumar Bhattacharyya, and Jugal~K Kalita.
\newblock Network anomaly detection: methods, systems and tools.
\newblock {\em Ieee communications surveys \& tutorials}, 16(1):303--336, 2013.

\bibitem{bishop2006pattern}
Christopher~M Bishop and Nasser~M Nasrabadi.
\newblock {\em Pattern recognition and machine learning}, volume~4.
\newblock Springer, 2006.

\bibitem{goodfellow2016deep}
Ian Goodfellow, Yoshua Bengio, and Aaron Courville.
\newblock {\em Deep learning}.
\newblock MIT press, 2016.

\bibitem{omar2013machine}
Salima Omar, Asri Ngadi, and Hamid~H Jebur.
\newblock Machine learning techniques for anomaly detection: an overview.
\newblock {\em International Journal of Computer Applications}, 79(2), 2013.

\bibitem{pang2021deep}
Guansong Pang, Chunhua Shen, Longbing Cao, and Anton Van~Den Hengel.
\newblock Deep learning for anomaly detection: A review.
\newblock {\em ACM computing surveys (CSUR)}, 54(2):1--38, 2021.

\bibitem{wang2020deep}
Ruoying Wang, Kexin Nie, Tie Wang, Yang Yang, and Bo~Long.
\newblock Deep learning for anomaly detection.
\newblock In {\em Proceedings of the 13th international conference on web search and data mining}, pages 894--896, 2020.

\bibitem{li2019video}
Nanjun Li and Faliang Chang.
\newblock Video anomaly detection and localization via multivariate gaussian fully convolution adversarial autoencoder.
\newblock {\em Neurocomputing}, 369:92--105, 2019.

\bibitem{fan2020robust}
Jinan Fan, Qianru Zhang, Jialei Zhu, Meng Zhang, Zhou Yang, and Hanxiang Cao.
\newblock Robust deep auto-encoding gaussian process regression for unsupervised anomaly detection.
\newblock {\em Neurocomputing}, 376:180--190, 2020.

\bibitem{zhou2017anomaly}
Chong Zhou and Randy~C Paffenroth.
\newblock Anomaly detection with robust deep autoencoders.
\newblock In {\em Proceedings of the 23rd ACM SIGKDD international conference on knowledge discovery and data mining}, pages 665--674, 2017.

\bibitem{kingma2013auto}
Diederik~P Kingma and Max Welling.
\newblock Auto-encoding variational bayes.
\newblock {\em arXiv preprint arXiv:1312.6114}, 2013.

\bibitem{challu2022deep}
Cristian~I Challu, Peihong Jiang, Ying~Nian Wu, and Laurent Callot.
\newblock Deep generative model with hierarchical latent factors for time series anomaly detection.
\newblock In {\em International Conference on Artificial Intelligence and Statistics}, pages 1643--1654. PMLR, 2022.

\bibitem{sheynin2021hierarchical}
Shelly Sheynin, Sagie Benaim, and Lior Wolf.
\newblock A hierarchical transformation-discriminating generative model for few shot anomaly detection.
\newblock In {\em Proceedings of the IEEE/CVF International Conference on Computer Vision}, pages 8495--8504, 2021.

\bibitem{choi2018generative}
Hyunsun Choi and Eric Jang.
\newblock Generative ensembles for robust anomaly detection, 2019.

\bibitem{an2015variational}
Jinwon An and Sungzoon Cho.
\newblock Variational autoencoder based anomaly detection using reconstruction probability.
\newblock {\em Special lecture on IE}, 2(1):1--18, 2015.

\bibitem{luchnikov2019variational}
Ilia~A Luchnikov, Alexander Ryzhov, Pieter-Jan Stas, Sergey~N Filippov, and Henni Ouerdane.
\newblock Variational autoencoder reconstruction of complex many-body physics.
\newblock {\em Entropy}, 21(11):1091, 2019.

\bibitem{zimmerer2018context}
David Zimmerer, Simon~AA Kohl, Jens Petersen, Fabian Isensee, and Klaus~H Maier-Hein.
\newblock Context-encoding variational autoencoder for unsupervised anomaly detection.
\newblock {\em arXiv preprint arXiv:1812.05941}, 2018.

\bibitem{ren2020data}
Yifu Ren, Jinhai Liu, Jianan Zhang, Lin Jiang, and Yanhong Luo.
\newblock A data reconstruction method based on adversarial conditional variational autoencoder.
\newblock In {\em 2020 IEEE 9th Data Driven Control and Learning Systems Conference (DDCLS)}, pages 622--626. IEEE, 2020.

\bibitem{noriega2005multilayer}
Leonardo Noriega.
\newblock Multilayer perceptron tutorial.
\newblock {\em School of Computing. Staffordshire University}, 4(5):444, 2005.

\bibitem{medsker2001recurrent}
Larry~R Medsker and LC~Jain.
\newblock Recurrent neural networks.
\newblock {\em Design and Applications}, 5(64-67):2, 2001.

\bibitem{zhao2017convolutional}
Bendong Zhao, Huanzhang Lu, Shangfeng Chen, Junliang Liu, and Dongya Wu.
\newblock Convolutional neural networks for time series classification.
\newblock {\em Journal of Systems Engineering and Electronics}, 28(1):162--169, 2017.

\bibitem{cui2016multi}
Zhicheng Cui, Wenlin Chen, and Yixin Chen.
\newblock Multi-scale convolutional neural networks for time series classification.
\newblock {\em arXiv preprint arXiv:1603.06995}, 2016.

\bibitem{borovykh2017conditional}
Anastasia Borovykh, Sander Bohte, and Cornelis~W Oosterlee.
\newblock Conditional time series forecasting with convolutional neural networks.
\newblock {\em arXiv preprint arXiv:1703.04691}, 2017.

\bibitem{koprinska2018convolutional}
Irena Koprinska, Dengsong Wu, and Zheng Wang.
\newblock Convolutional neural networks for energy time series forecasting.
\newblock In {\em 2018 international joint conference on neural networks (IJCNN)}, pages 1--8. IEEE, 2018.

\bibitem{zheng2014time}
Yi~Zheng, Qi~Liu, Enhong Chen, Yong Ge, and J~Leon Zhao.
\newblock Time series classification using multi-channels deep convolutional neural networks.
\newblock In {\em International conference on web-age information management}, pages 298--310. Springer, 2014.

\bibitem{lindsay2021convolutional}
Grace~W Lindsay.
\newblock Convolutional neural networks as a model of the visual system: Past, present, and future.
\newblock {\em Journal of cognitive neuroscience}, 33(10):2017--2031, 2021.

\bibitem{ravanelli2018speaker}
Mirco Ravanelli and Yoshua Bengio.
\newblock Speaker recognition from raw waveform with sincnet.
\newblock In {\em 2018 IEEE spoken language technology workshop (SLT)}, pages 1021--1028. IEEE, 2018.

\bibitem{humairani2022wavelet}
Annisa Humairani, Achmad Rizal, Inung Wijayanto, Sugondo Hadiyoso, and Yunendah~Nur Fuadah.
\newblock Wavelet-based entropy analysis on eeg signal for detecting seizures.
\newblock In {\em 2022 10th International Conference on Information and Communication Technology (ICoICT)}, pages 93--98. IEEE, 2022.

\bibitem{ahmad2017mallat}
Muhammad~Zubair Ahmad, Awais~Mehmood Kamboh, Sajid Saleem, and Amir~Ali Khan.
\newblock Mallat’s scattering transform based anomaly sensing for detection of seizures in scalp eeg.
\newblock {\em IEEE Access}, 5:16919--16929, 2017.

\bibitem{boonyakitanont2020review}
Poomipat Boonyakitanont, Apiwat Lek-Uthai, Krisnachai Chomtho, and Jitkomut Songsiri.
\newblock A review of feature extraction and performance evaluation in epileptic seizure detection using eeg.
\newblock {\em Biomedical Signal Processing and Control}, 57:101702, 2020.

\bibitem{ang2012filter}
Kai~Keng Ang, Zheng~Yang Chin, Chuanchu Wang, Cuntai Guan, and Haihong Zhang.
\newblock Filter bank common spatial pattern algorithm on bci competition iv datasets 2a and 2b.
\newblock {\em Frontiers in neuroscience}, 6:39, 2012.

\bibitem{shajil2020multiclass}
Nijisha Shajil, Sasikala Mohan, Poonguzhali Srinivasan, Janani Arivudaiyanambi, and Arunnagiri Arasappan~Murrugesan.
\newblock Multiclass classification of spatially filtered motor imagery eeg signals using convolutional neural network for bci based applications.
\newblock {\em Journal of Medical and Biological Engineering}, 40:663--672, 2020.

\bibitem{apicella2022eeg}
Andrea Apicella, Pasquale Arpaia, Mirco Frosolone, Giovanni Improta, Nicola Moccaldi, and Andrea Pollastro.
\newblock Eeg-based measurement system for monitoring student engagement in learning 4.0.
\newblock {\em Scientific Reports}, 12(1):5857, 2022.

\bibitem{arpaia2023sinc}
Pasquale Arpaia, Elisa Bertone, Antonio Esposito, Angela Natalizio, Marco Parvis, Alessandra Laura~Giulia Pedrocchi, and Andrea Pollastro.
\newblock Sinc-eegnet for improving performance while reducing calibration of a motor imagery-based bci.
\newblock In {\em 2023 IEEE International Conference on Metrology for eXtended Reality, Artificial Intelligence and Neural Engineering (MetroXRAINE)}, pages 1063--1068. IEEE, 2023.

\bibitem{li2022interpretable}
Xuhong Li, Haoyi Xiong, Xingjian Li, Xuanyu Wu, Xiao Zhang, Ji~Liu, Jiang Bian, and Dejing Dou.
\newblock Interpretable deep learning: Interpretation, interpretability, trustworthiness, and beyond.
\newblock {\em Knowledge and Information Systems}, 64(12):3197--3234, 2022.

\bibitem{chakraborty2017interpretability}
Supriyo Chakraborty, Richard Tomsett, Ramya Raghavendra, Daniel Harborne, Moustafa Alzantot, Federico Cerutti, Mani Srivastava, Alun Preece, Simon Julier, Raghuveer~M Rao, et~al.
\newblock Interpretability of deep learning models: A survey of results.
\newblock In {\em 2017 IEEE smartworld, ubiquitous intelligence \& computing, advanced \& trusted computed, scalable computing \& communications, cloud \& big data computing, Internet of people and smart city innovation (smartworld/SCALCOM/UIC/ATC/CBDcom/IOP/SCI)}, pages 1--6. IEEE, 2017.

\bibitem{tan2025seda}
Weilong Tan, Hongyi Zhang, Yingbei Wang, Weimin Wen, Liang Chen, Han Li, Xingen Gao, and Nianyin Zeng.
\newblock Seda-eeg: A semi-supervised emotion recognition network with domain adaptation for cross-subject eeg analysis.
\newblock {\em Neurocomputing}, 622:129315, 2025.

\bibitem{hu2023}
Dayu Hu, Ke~Liang, Sihang Zhou, Wenxuan Tu, Meng Liu, and Xinwang Liu.
\newblock scdfc: A deep fusion clustering method for single-cell rna-seq data.
\newblock {\em Briefings in Bioinformatics}, 24(4):bbad216, 06 2023.

\bibitem{hu2024}
Dayu Hu, Zhibin Dong, Ke~Liang, Hao Yu, Siwei Wang, and Xinwang Liu.
\newblock High-order topology for deep single-cell multiview fuzzy clustering.
\newblock {\em IEEE Transactions on Fuzzy Systems}, 32(8):4448--4459, 2024.

\bibitem{ruff2019deep}
Lukas Ruff, Robert~A Vandermeulen, Nico G{\"o}rnitz, Alexander Binder, Emmanuel M{\"u}ller, Klaus-Robert M{\"u}ller, and Marius Kloft.
\newblock Deep semi-supervised anomaly detection.
\newblock {\em arXiv preprint arXiv:1906.02694}, 2019.

\bibitem{akcay2019ganomaly}
Samet Akcay, Amir Atapour-Abarghouei, and Toby~P Breckon.
\newblock Ganomaly: Semi-supervised anomaly detection via adversarial training.
\newblock In {\em Computer Vision--ACCV 2018: 14th Asian Conference on Computer Vision, Perth, Australia, December 2--6, 2018, Revised Selected Papers, Part III 14}, pages 622--637. Springer, 2019.

\bibitem{zhao2024digan}
Puyang Zhao, Xinhui Liu, Zhiyi Yue, Qianyu Zhao, Xinzhi Liu, Yuhui Deng, and Jingjin Wu.
\newblock Digan breakthrough: Advancing diabetic data analysis with innovative gan-based imbalance correction techniques.
\newblock {\em Computer Methods and Programs in Biomedicine Update}, 5:100152, 2024.

\bibitem{yu2014improved}
Hualong Yu and Jun Ni.
\newblock An improved ensemble learning method for classifying high-dimensional and imbalanced biomedicine data.
\newblock {\em IEEE/ACM transactions on computational biology and bioinformatics}, 11(4):657--666, 2014.

\bibitem{roy2024learning}
Debashis Roy, Anandarup Roy, and Utpal Roy.
\newblock Learning from imbalanced data in healthcare: State-of-the-art and research challenges.
\newblock {\em Computational Intelligence in Healthcare Informatics}, pages 19--32, 2024.

\bibitem{khan2023shallow}
Gul~Hameed Khan, Nadeem~Ahmad Khan, Muhammad Awais~Bin Altaf, and Qammer Abbasi.
\newblock A shallow autoencoder framework for epileptic seizure detection in eeg signals.
\newblock {\em Sensors}, 23(8):4112, 2023.

\bibitem{hastie2009elements}
Trevor Hastie, Robert Tibshirani, Jerome~H Friedman, and Jerome~H Friedman.
\newblock {\em The elements of statistical learning: data mining, inference, and prediction}, volume~2.
\newblock Springer, 2009.

\bibitem{yuan2018multi}
Ye~Yuan, Guangxu Xun, Kebin Jia, and Aidong Zhang.
\newblock A multi-view deep learning framework for eeg seizure detection.
\newblock {\em IEEE journal of biomedical and health informatics}, 23(1):83--94, 2018.

\bibitem{abdelhameed2019semi}
Ahmed~M Abdelhameed and Magdy Bayoumi.
\newblock Semi-supervised eeg signals classification system for epileptic seizure detection.
\newblock {\em IEEE Signal Processing Letters}, 26(12):1922--1926, 2019.

\bibitem{abdelhameed2021deep}
Ahmed Abdelhameed and Magdy Bayoumi.
\newblock A deep learning approach for automatic seizure detection in children with epilepsy.
\newblock {\em Frontiers in Computational Neuroscience}, 15:650050, 2021.

\bibitem{abdelhameed2021efficient}
Ahmed~M Abdelhameed and Magdy Bayoumi.
\newblock An efficient deep learning system for epileptic seizure prediction.
\newblock In {\em 2021 IEEE International Symposium on Circuits and Systems (ISCAS)}, pages 1--5. IEEE, 2021.

\bibitem{daoud2019deep}
Hisham Daoud and Magdy Bayoumi.
\newblock Deep learning approach for epileptic focus localization.
\newblock {\em IEEE transactions on biomedical circuits and systems}, 14(2):209--220, 2019.

\bibitem{wang2022epileptic}
Ruyan Wang, Linhai Wang, Peng He, Yaping Cui, and Dapeng Wu.
\newblock Epileptic seizures prediction based on unsupervised learning for feature extraction.
\newblock In {\em ICC 2022-IEEE International Conference on Communications}, pages 4643--4648. IEEE, 2022.

\bibitem{he2023unsupervised}
Peng He, Linhai Wang, Yaping Cui, Ruyan Wang, and Dapeng Wu.
\newblock Unsupervised feature learning based on autoencoder for epileptic seizures prediction.
\newblock {\em Applied Intelligence}, 53(18):20766--20784, 2023.

\bibitem{wen2018deep}
Tingxi Wen and Zhongnan Zhang.
\newblock Deep convolution neural network and autoencoders-based unsupervised feature learning of eeg signals.
\newblock {\em IEEE Access}, 6:25399--25410, 2018.

\bibitem{shoeibi2022detection}
Afshin Shoeibi, Navid Ghassemi, Marjane Khodatars, Parisa Moridian, Roohallah Alizadehsani, Assef Zare, Abbas Khosravi, Abdulhamit Subasi, U~Rajendra Acharya, and Juan~M Gorriz.
\newblock Detection of epileptic seizures on eeg signals using anfis classifier, autoencoders and fuzzy entropies.
\newblock {\em Biomedical Signal Processing and Control}, 73:103417, 2022.

\bibitem{huang2022novel}
Xiaojie Huang, Xiangtao Sun, Lijun Zhang, Tong Zhu, Hao Yang, Qingsong Xiong, and Lijie Feng.
\newblock A novel epilepsy detection method based on feature extraction by deep autoencoder on eeg signal.
\newblock {\em International Journal of Environmental Research and Public Health}, 19(22):15110, 2022.

\bibitem{lundberg2017unified}
Scott~M Lundberg and Su-In Lee.
\newblock A unified approach to interpreting model predictions.
\newblock {\em Advances in neural information processing systems}, 30, 2017.

\bibitem{yildiz2022unsupervised}
{\.I}lkay Y{\i}ld{\i}z, Rachael Garner, Matthew Lai, and Dominique Duncan.
\newblock Unsupervised seizure identification on eeg.
\newblock {\em Computer methods and programs in biomedicine}, 215:106604, 2022.

\bibitem{de2024detection}
Ana Maria~Amaro de~Sousa, Michel~JAM van Putten, St{\'e}phanie van~den Berg, and Maryam~Amir Haeri.
\newblock Detection of interictal epileptiform discharges with semi-supervised deep learning.
\newblock {\em Biomedical Signal Processing and Control}, 88:105610, 2024.

\bibitem{potter2022unsupervised}
Ilkay~Y{\i}ld{\i}z Potter, George Zerveas, Carsten Eickhoff, and Dominique Duncan.
\newblock Unsupervised multivariate time-series transformers for seizure identification on eeg.
\newblock In {\em 2022 21st IEEE International Conference on Machine Learning and Applications (ICMLA)}, pages 1304--1311. IEEE, 2022.

\bibitem{grenander1959nyquist}
Ulf Grenander.
\newblock The nyquist frequency is that frequency whose period is two sampling intervals.
\newblock {\em Probability and Statistics: The Harald Cram{\'e}r Volume}, 434, 1959.

\bibitem{rabiner2010theory}
Lawrence Rabiner and Ronald Schafer.
\newblock {\em Theory and applications of digital speech processing}.
\newblock Prentice Hall Press, 2010.

\bibitem{mitra2001digital}
Sanjit~K Mitra.
\newblock {\em Digital signal processing: a computer-based approach}.
\newblock McGraw-Hill Higher Education, 2001.

\bibitem{burgess2018understanding}
Christopher~P Burgess, Irina Higgins, Arka Pal, Loic Matthey, Nick Watters, Guillaume Desjardins, and Alexander Lerchner.
\newblock Understanding disentangling in $\beta$-vae.
\newblock {\em arXiv preprint arXiv:1804.03599}, 2018.

\bibitem{sadouk2019cnn}
Lamyaa Sadouk.
\newblock Cnn approaches for time series classification.
\newblock {\em Time series analysis-data, methods, and applications}, 5:57--78, 2019.

\bibitem{choi2022multivariate}
Taesung Choi, Dongkun Lee, Yuchae Jung, and Ho-Jin Choi.
\newblock Multivariate time-series anomaly detection using seqvae-cnn hybrid model.
\newblock In {\em 2022 International Conference on Information Networking (ICOIN)}, pages 250--253. IEEE, 2022.

\bibitem{brigham1988fast}
EO~Brigham.
\newblock The fast fourier transform and its applications, 1988.

\bibitem{ravi2019user}
Aravind Ravi, Nargess Heydari, and Ning Jiang.
\newblock User-independent ssvep bci using complex fft features and cnn classification.
\newblock In {\em 2019 IEEE International Conference on Systems, Man and Cybernetics (SMC)}, pages 4175--4180. IEEE, 2019.

\bibitem{nakayama2007brain}
Kenji Nakayama, Yasuaki Kaneda, and Akihiro Hirano.
\newblock A brain computer interface based on fft and multilayer neural network-feature extraction and generalization.
\newblock In {\em 2007 International Symposium on Intelligent Signal Processing and Communication Systems}, pages 826--829. IEEE, 2007.

\bibitem{andrzejak2001indications}
Ralph~G Andrzejak, Klaus Lehnertz, Florian Mormann, Christoph Rieke, Peter David, and Christian~E Elger.
\newblock Indications of nonlinear deterministic and finite-dimensional structures in time series of brain electrical activity: Dependence on recording region and brain state.
\newblock {\em Physical Review E}, 64(6):061907, 2001.

\bibitem{physiobank2000physionet}
PhysioToolkit PhysioBank.
\newblock Physionet: components of a new research resource for complex physiologic signals.
\newblock {\em Circulation}, 101(23):e215--e220, 2000.

\bibitem{prasanna2021automated}
J~Prasanna, MSP Subathra, Mazin~Abed Mohammed, Robertas Dama{\v{s}}evi{\v{c}}ius, Nanjappan~Jothiraj Sairamya, and S~Thomas George.
\newblock Automated epileptic seizure detection in pediatric subjects of chb-mit eeg database—a survey.
\newblock {\em Journal of Personalized Medicine}, 11(10):1028, 2021.

\bibitem{choi2019novel}
Gwangho Choi, Chulkyun Park, Junkyung Kim, Kyoungin Cho, Tae-Joon Kim, HwangSik Bae, Kyeongyuk Min, Ki-Young Jung, and Jongwha Chong.
\newblock A novel multi-scale 3d cnn with deep neural network for epileptic seizure detection.
\newblock In {\em 2019 IEEE International Conference on Consumer Electronics (ICCE)}, pages 1--2. IEEE, 2019.

\bibitem{zhou2018epileptic}
Mengni Zhou, Cheng Tian, Rui Cao, Bin Wang, Yan Niu, Ting Hu, Hao Guo, and Jie Xiang.
\newblock Epileptic seizure detection based on eeg signals and cnn.
\newblock {\em Frontiers in neuroinformatics}, 12:95, 2018.

\bibitem{tasci2023epilepsy}
Irem Tasci, Burak Tasci, Prabal~D Barua, Sengul Dogan, Turker Tuncer, Elizabeth~Emma Palmer, Hamido Fujita, and U~Rajendra Acharya.
\newblock Epilepsy detection in 121 patient populations using hypercube pattern from eeg signals.
\newblock {\em Information Fusion}, 96:252--268, 2023.

\bibitem{park2018epileptic}
Chulkyun Park, Gwangho Choi, Junkyung Kim, Sangdeok Kim, Tae-Joon Kim, Kyeongyuk Min, Ki-Young Jung, and Jongwha Chong.
\newblock Epileptic seizure detection for multi-channel eeg with deep convolutional neural network.
\newblock In {\em 2018 International Conference on Electronics, Information, and Communication (ICEIC)}, pages 1--5. IEEE, 2018.

\bibitem{kaziha2020convolutional}
Omar Kaziha and Talal Bonny.
\newblock A convolutional neural network for seizure detection.
\newblock In {\em 2020 Advances in Science and Engineering Technology International Conferences (ASET)}, pages 1--5. IEEE, 2020.

\bibitem{hassan2022epileptic}
Fatima Hassan, Syed~Fawad Hussain, and Saeed~Mian Qaisar.
\newblock Epileptic seizure detection using a hybrid 1d cnn-machine learning approach from eeg data.
\newblock {\em Journal of Healthcare Engineering}, 2022, 2022.

\bibitem{qiu2018denoising}
Yang Qiu, Weidong Zhou, Nana Yu, and Peidong Du.
\newblock Denoising sparse autoencoder-based ictal eeg classification.
\newblock {\em IEEE Transactions on Neural Systems and Rehabilitation Engineering}, 26(9):1717--1726, 2018.

\bibitem{liashchynskyi2019grid}
Petro Liashchynskyi and Pavlo Liashchynskyi.
\newblock Grid search, random search, genetic algorithm: a big comparison for nas.
\newblock {\em arXiv preprint arXiv:1912.06059}, 2019.

\bibitem{yao2007early}
Yuan Yao, Lorenzo Rosasco, and Andrea Caponnetto.
\newblock On early stopping in gradient descent learning.
\newblock {\em Constructive Approximation}, 26:289--315, 2007.

\bibitem{apicella2023effects}
Andrea Apicella, Francesco Isgr{\`o}, Andrea Pollastro, and Roberto Prevete.
\newblock On the effects of data normalization for domain adaptation on eeg data.
\newblock {\em Engineering Applications of Artificial Intelligence}, 123:106205, 2023.

\bibitem{shapiro1965analysis}
Samuel~Sanford Shapiro and Martin~B Wilk.
\newblock An analysis of variance test for normality (complete samples).
\newblock {\em Biometrika}, 52(3-4):591--611, 1965.

\bibitem{ostertagova2014methodology}
Eva Ostertagova, Oskar Ostertag, and Jozef Kov{\'a}{\v{c}}.
\newblock Methodology and application of the kruskal-wallis test.
\newblock {\em Applied mechanics and materials}, 611:115--120, 2014.

\bibitem{mcknight2010mann}
Patrick~E McKnight and Julius Najab.
\newblock Mann-whitney u test.
\newblock {\em The Corsini encyclopedia of psychology}, pages 1--1, 2010.

\bibitem{kim2017understanding}
Tae~Kyun Kim.
\newblock Understanding one-way anova using conceptual figures.
\newblock {\em Korean journal of anesthesiology}, 70(1):22, 2017.

\bibitem{givnan2022anomaly}
Sean Givnan, Carl Chalmers, Paul Fergus, Sandra Ortega-Martorell, and Tom Whalley.
\newblock Anomaly detection using autoencoder reconstruction upon industrial motors.
\newblock {\em Sensors}, 22(9):3166, 2022.

\bibitem{tun2020network}
May~Thet Tun, Dim~En Nyaung, and Myat~Pwint Phyu.
\newblock Network anomaly detection using threshold-based sparse.
\newblock In {\em Proceedings of the 11th International Conference on Advances in Information Technology}, pages 1--8, 2020.

\bibitem{elsayed2020detecting}
Mahmoud~Said Elsayed, Nhien-An Le-Khac, Soumyabrata Dev, and Anca~Delia Jurcut.
\newblock Detecting abnormal traffic in large-scale networks.
\newblock In {\em 2020 International Symposium on Networks, Computers and Communications (ISNCC)}, pages 1--7. IEEE, 2020.

\bibitem{liu2008isolation}
Fei~Tony Liu, Kai~Ming Ting, and Zhi-Hua Zhou.
\newblock Isolation forest.
\newblock In {\em 2008 eighth ieee international conference on data mining}, pages 413--422. IEEE, 2008.

\bibitem{alghushairy2020review}
Omar Alghushairy, Raed Alsini, Terence Soule, and Xiaogang Ma.
\newblock A review of local outlier factor algorithms for outlier detection in big data streams.
\newblock {\em Big Data and Cognitive Computing}, 5(1):1, 2020.

\bibitem{moore2009introduction}
David~S Moore, George~P McCabe, and Bruce~A Craig.
\newblock {\em Introduction to the Practice of Statistics}, volume~4.
\newblock WH Freeman New York, 2009.

\bibitem{niu2021review}
Zhaoyang Niu, Guoqiang Zhong, and Hui Yu.
\newblock A review on the attention mechanism of deep learning.
\newblock {\em Neurocomputing}, 452:48--62, 2021.

\bibitem{minh2022explainable}
Dang Minh, H~Xiang Wang, Y~Fen Li, and Tan~N Nguyen.
\newblock Explainable artificial intelligence: a comprehensive review.
\newblock {\em Artificial Intelligence Review}, pages 1--66, 2022.

\bibitem{apicella2023strategies}
Andrea Apicella, Luca Di~Lorenzo, Francesco Isgrò, Andrea Pollastro, and Roberto Prevete.
\newblock Strategies to exploit xai to improve classification systems.
\newblock {\em Communications in Computer and Information Science}, 1901 CCIS:147 – 159, 2023.
\newblock Cited by: 5.

\bibitem{li2024optimal}
Maximilian Li and Lucas Janson.
\newblock Optimal ablation for interpretability.
\newblock {\em Advances in Neural Information Processing Systems}, 37:109233--109282, 2024.

\bibitem{teng2022survey}
Qiaoying Teng, Zhe Liu, Yuqing Song, Kai Han, and Yang Lu.
\newblock A survey on the interpretability of deep learning in medical diagnosis.
\newblock {\em Multimedia Systems}, 28(6):2335--2355, 2022.

\end{thebibliography}
\end{document}